\documentclass[review, table]{elsarticle}

\usepackage{multirow}
\usepackage{mhchem}
\usepackage{lineno,hyperref}
\usepackage{amsmath}
\usepackage{bm}
\usepackage{graphicx}
\usepackage{subcaption}
\usepackage{caption}
\usepackage{verbatim}
\usepackage{textcomp}
\usepackage[T1]{fontenc}
\usepackage[utf8]{inputenc}
\usepackage{latexsym}
\usepackage{setspace}
\usepackage[titletoc]{appendix}
\usepackage{amssymb}
\usepackage{dirtytalk}
\usepackage{adjustbox}
\usepackage[flushleft]{threeparttable}
\usepackage{array,booktabs,makecell}
\usepackage[font=small]{caption}
\usepackage{stackengine}
\usepackage[bottom]{footmisc}
\usepackage{amsmath}
\usepackage{graphicx} 
\usepackage{gensymb}
\usepackage{float}
\usepackage{natbib}
\usepackage{amssymb}
\usepackage{amsthm}
\usepackage{relsize}
\usepackage{csquotes}
\usepackage{dirtytalk}
\usepackage{hyperref}
\usepackage{comment}
\modulolinenumbers[5]
\usepackage{graphicx}
\usepackage{collcell}
\usepackage{booktabs}
\usepackage{siunitx}
\usepackage[table]{xcolor}
\usepackage{tikz}
\usepackage{etoolbox}
\usepackage{pgf} % for calculating the values for gradientNeuro
\usepackage{multirow}
\usepackage{mhchem}
\usepackage{xcolor}
\usepackage{lineno,hyperref}
\usepackage{amsmath}
\usepackage{bm}
\usepackage{graphicx}
\usepackage{subcaption}
\usepackage{caption}
\usepackage{verbatim}
\usepackage{textcomp}
\usepackage{latexsym}
\usepackage{setspace}
\usepackage[titletoc]{appendix}
\usepackage{amssymb}
\usepackage{dirtytalk}
\usepackage{adjustbox}
\usepackage[flushleft]{threeparttable}
\usepackage{array,booktabs,makecell}
\usepackage[font=small]{caption}
\usepackage{stackengine}
\usepackage[bottom]{footmisc}
\usepackage{amsmath}
\usepackage{graphicx} 
\usepackage{gensymb}
\usepackage{float}
\usepackage{natbib}
\usepackage{amssymb}
\usepackage{amsthm}
\usepackage{relsize}
\usepackage{csquotes}
\usepackage{dirtytalk}
\usepackage{hyperref}

\modulolinenumbers[5]
\journal{NeuroImage}

\biboptions{authoryear}

\definecolor{highNeuro}{HTML}{F8F7F5}  % the color for the highNeuroest number in your data set
\definecolor{lowNeuro}{HTML}{7B3A98}  % the color for the lowNeuroest number in your data set
\newcommand*{\opacity}{100}% here you can change the opacity of the background color!
\newcommand*{\minvalNeuro}{-1.0}% define the minimum value on your data set
\newcommand*{\maxvalNeuro}{0.5}% define the maximum value in your data set!
\newcommand{\gradientNeuro}[1]{
    \ifdimcomp{#1pt}{>}{\maxvalNeuro pt}{#1}{
        \ifdimcomp{#1pt}{<}{\minvalNeuro pt}{#1}{
            \pgfmathparse{int(round(100*(#1/(\maxvalNeuro-\minvalNeuro))-(\minvalNeuro*(100/(\maxvalNeuro-\minvalNeuro)))))}
            \xdef\tempa{\pgfmathresult}
            \cellcolor{highNeuro!\tempa!lowNeuro!\opacity} #1
    }}
}

\definecolor{highICC}{HTML}{7B3A98}  % the color for the highICCest number in your data set
\definecolor{lowICC}{HTML}{F8F7F5}  % the color for the lowICCest number in your data set
\newcommand*{\minvalICC}{-0.5}% define the minimum value on your data set
\newcommand*{\maxvalICC}{1.0}% define the maximum value in your data set!
\newcommand{\gradientICC}[1]{
    \ifdimcomp{#1pt}{>}{\maxvalICC pt}{#1}{
        \ifdimcomp{#1pt}{<}{\minvalICC pt}{#1}{
            \pgfmathparse{int(round(100*(#1/(\maxvalICC-\minvalICC))-(\minvalICC*(100/(\maxvalICC-\minvalICC)))))}
            \xdef\tempa{\pgfmathresult}
            \cellcolor{highICC!\tempa!lowICC!\opacity} #1
    }}
}

\definecolor{highWMH}{HTML}{F8F7F5}  % the color for the highWMHest number in your data set
\definecolor{lowWMH}{HTML}{7B3A98}  % the color for the lowWMHest number in your data set
\newcommand*{\minvalWMH}{-1.0}% define the minimum value on your data set
\newcommand*{\maxvalWMH}{0.5}% define the maximum value in your data set!
\newcommand{\gradientWMH}[1]{
    \ifdimcomp{#1pt}{>}{\maxvalWMH pt}{#1}{
        \ifdimcomp{#1pt}{<}{\minvalWMH pt}{#1}{
            \pgfmathparse{int(round(100*(#1/(\maxvalWMH-\minvalWMH))-(\minvalWMH*(100/(\maxvalWMH-\minvalWMH)))))}
            \xdef\tempa{\pgfmathresult}
            \cellcolor{highWMH!\tempa!lowWMH!\opacity} #1
    }}
}

\begin{document}
\sloppy
\begin{frontmatter}

\title{Automated deep learning segmentation of high-resolution 7 T postmortem MRI for quantitative analysis of structure-pathology correlations in\\
neurodegenerative diseases}

\author[1,2]{Pulkit Khandelwal\corref{cor2}}
\author[1]{Michael Tran Duong}
\author[3]{Shokufeh Sadaghiani}
\author[2,4]{Sydney Lim}
\author[2,4]{Amanda Denning}
\author[2,4]{Eunice Chung}
\author[2,4]{Sadhana Ravikumar}
\author[3]{Sanaz Arezoumandan}
\author[3]{Claire Peterson}
\author[2,4]{Madigan Bedard}
\author[3]{Noah Capp}
\author[2,4]{Ranjit Ittyerah}
\author[3]{Elyse Migdal}
\author[3]{Grace Choi}
\author[3]{Emily Kopp}
\author[3]{Bridget Loja}
\author[3]{Eusha Hasan}
\author[3]{Jiacheng Li}
\author[3]{Alejandra Bahena}
\author[4]{Karthik Prabhakaran}
\author[4]{Gabor Mizsei}
\author[3]{Marianna Gabrielyan}
\author[5]{Theresa Schuck}
\author[3]{Winifred Trotman}
\author[5]{John Robinson}
\author[3]{Daniel T. Ohm}
\author[5]{Edward B. Lee}
\author[5]{John Q. Trojanowski\corref{cor1}}
\author[3]{Corey McMillan}
\author[3]{Murray Grossman\corref{cor1}}
\author[3]{David J. Irwin}
\author[3]{John A. Detre}
\author[4]{M. Dylan Tisdall}
\author[2,3]{Sandhitsu R. Das}
\author[6]{Laura E.M. Wisse}
\author[3]{David A. Wolk}
\author[2,4]{Paul A. Yushkevich}

\address[1]{Department of Bioengineering, University of Pennsylvania, Philadelphia, USA}
\address[2]{Penn Image Computing and Science Laboratory, University of Pennsylvania, Philadelphia, USA}
\address[3]{Department of Neurology, University of Pennsylvania, Philadelphia, USA}
\address[4]{Department of Radiology, University of Pennsylvania, Philadelphia, USA}
\address[5]{Department of Pathology and Laboratory Medicine, University of Pennsylvania, Philadelphia, Pennsylvania, USA}
\address[6]{Department of Diagnostic Radiology, Lund University, Lund, Sweden}
\cortext[cor1]{Author deceased.}
\cortext[cor2]{Corresponding author: Pulkit Khandelwal\\
\hskip 2cm Richards Medical Research Laboratories\\
\hskip 2cm 6th Floor 605, 3700 Hamilton Walk, Philadelphia, PA 19104\\
\hskip 2cm Email address: pulks@seas.upenn.edu}

\begin{abstract} \textcolor{blue}{\textit{\textbf{Postmortem}} MRI allows brain anatomy to be examined at high resolution and to link pathology measures with morphometric measurements.} However, automated segmentation methods for brain mapping in postmortem MRI are not well developed, primarily due to limited availability of labeled datasets, and heterogeneity in scanner hardware and acquisition protocols. In this work, we present a high resolution of \textcolor{blue}{135} postmortem human brain tissue specimens imaged at 0.3 mm$^{3}$ isotropic using a T2w sequence on a 7T whole-body MRI scanner. We developed a deep learning pipeline to segment the cortical mantle by benchmarking the performance of nine deep neural architectures, \textcolor{blue}{followed by post-hoc topological correction.} We then segment four subcortical structures (caudate, putamen, globus pallidus, and thalamus), white matter hyperintensities, and the normal appearing white matter. \textcolor{blue}{We show generalizing capabilities across whole brain hemispheres in different specimens, and also on unseen images acquired at 0.28 mm$^{3}$ and 0.16 mm$^{3}$ isotropic T2*w FLASH sequence at 7T.} We then compute localized cortical thickness and volumetric measurements across key regions, and link them with semi-quantitative neuropathological ratings. Our code, Jupyter notebooks, and the containerized executables are publicly available at the \href{https://pulkit-khandelwal.github.io/exvivo-brain-upenn/}{\textbf{project webpage}}.
\end{abstract}


\end{frontmatter}

\section{Introduction}
\label{section:introduction}
Neurodegenerative diseases are increasingly understood to be heterogeneous, with multiple distinct neuropathological processes jointly contributing to neurodegeneration in most patients, called \emph{mixed pathology} \cite{schneider2007mixed}. For example, many patients diagnosed at autopsy with Alzheimer's disease (AD) also have brain lesions associated with vascular disease, TDP-43 proteinopathy, and $\alpha$-synuclein pathology \cite{robinson2018neurodegenerative, matej2019alzheimer}. Currently, some of these pathological processes (particularly TDP-43 and $\alpha$-synuclein pathologies) cannot be reliably detected with antemortem biomarkers, which makes it difficult for clinicians to determine to what extent cognitive decline in individual patients is driven by AD vs. other factors. The recent modest successes of AD treatments in clinical trials \cite{van2022lecanemab} make it ever more important to derive antemortem biomarkers that can detect and quantify mixed pathology, so that treatments can be prioritized for those most likely to benefit from them.

Importantly, the understanding and utility of antemortem biomarkers are augmented by the coupling of imaging and postmortem pathology. Following autopsy, histological examination of the donor's brain tissue provides a semi-quantitative assessment of the presence and severity of various pathological drivers of neurodegeneration. Associations between these pathology measures and regional measures of neurodegeneration, such as cortical thickness, can identify patterns of neurodegeneration probabilistically linked to specific pathological drivers \cite{de2020contribution, frigerio2021amyloid, wisse2021downstream, wisse2020high}. Such association studies can use either antemortem or postmortem imaging \cite{wisse2017comparison}. Both approaches have their limitations. When the time between antemortem imaging and death is susbtantial, the postmortem pathology may not accurately match the state of pathology at the time of imaging. Postmortem MRI generally requires dedicated imaging facilities and image analysis algorithms, and whatever neurodegeneration/pathology association patterns are discovered have to be ``translated" into the antemortem imaging domain for use as antemortem biomarkers. However, postmortem MRI allows imaging at much greater resolution than antemortem, allowing structure/pathology associations to be examined with greater granularity than antemortem antemortem imaging.

Furthermore, postmortem MRI of the brain can provide an advantage over antemortem MRI for visualizing detailed and intricate neuroanatomy and linking macroscopic morphometric measures such as cortical thickness to underlying cytoarchitecture and pathology \cite{mancini2020multimodal, beaujoin2018post, pallebage2018dissecting, garcia2020protocol, augustinack2010direct, vega2021deep, alkemade2022unified, iglesias2015computational, adler2014histology}. Recent inquiries comparing postmortem imaging with histopathology have demonstrated relationships between atrophy measures and neurodegenerative pathology \cite{wisse2021downstream, ravikumar2021ex, yushkevich2021three, makkinejad2019associations}. Such associations corroborate patterns of neurodegeneration by specifically linking them with the underlying contributing pathology such as TAR DNA-binding protein 43 (TDP-43), phosphorylated tau (p-tau) and $\alpha$-synuclein in Alzheimer's Disease (AD). \textcolor{blue}{In particular, \cite{ravikumar2021ex} found significant correlations between tau pathology and thickness in the entorhinal cortex (ERC) and stratum radiatum lacunosum moleculare (SRLM) consistent with early Braak stages. Separately, \cite{wisse2021downstream} found significant associations of TDP-43 with thickness in the hippocampal subregions.} Postmortem imaging also helps in characterizing underlying anatomy at the scale of subcortical layers \cite{augustinack2013medial, kenkhuis20197t}, such as hippocampal subfields in the medial temporal lobe (MTL) \cite{yushkevich2021three, ravikumar2020building}. Several studies have also explored pathology/MRI associations in other neurodegenerative diseases, such as frontotemporal lobar degeneration (FTLD) and amyotrophic lateral sclerosis (ALS) \cite{gordon2016advances, irwin2015frontotemporal, mackenzie2011harmonized}. Previous work has identified correlations between high resolution postmortem MRI and histopathology, to map myelin and iron deposits in cortical laminae \cite{bulk2020quantitative}, due to oligodendrocytes and pathologic iron inclusions in astrocytes and microglia \cite{tisdall2021joint}, which are a major source of iron because of myelination demands. Therefore, postmortem MRI would be helpful for validating and refining pathophysiological correlates derived from antemortem studies. \textcolor{blue}{Additionally, the volume of the WMH burden is an indirect marker of cerebrovascular pathology and the associations between cortical thickness and WMH \cite{rizvi2018effect, du2005white, dadar2022white} are complementary to the associations between cortical thickness and tau, TDP-43, amyloid-$\beta$ and $\alpha$-synuclein pathology. Also, \cite{van2023subcortical} suggests that the subcortical brain structures are highly involved in dementia risk. Smaller volumes and thickness measurements of thalamus, amygdala, and hippocampus were associated with incident dementia.}

Compared to antemortem MRI, postmortem MRI is not affected by head or respiratory motion artifacts and has much less stringent time. \textcolor{blue}{Compared to histology, postmortem MRI is less affected by distortion or tearing, and provides a continuous 3D representation of brain anatomy. However, both histology and postmortem MRI are affected by changes occurring in the agonal state, during brain removal, and during tissue fixation and handling.} Indeed, postmortem MRI is often used to provide a 3D reference space onto which 2D histological images are mapped. Combined analysis of postmortem MRI and histology makes it possible to link morphological changes in the brain to underlying pathology as well as to generate anatomically correct parcellations of the brain based on cytoarchitecture \cite{schiffer2021convolutional, amunts2020julich}, and pathoarchitecture \cite{augustinack2013medial}. Postmortem MRI could also act as a reference space to generate quantitative 3D maps of neurodegenerative proteinopathies from serial histology imaging \cite{ushizima2022deep, yushkevich2021three}.

Given the rising use of high resolution postmortem MRI in neurodegenerative disease research, automated techniques are imperative to effectively analyze such growing datasets. Particularly, in the case of structure-pathology association studies, scaling them beyond a few dozen datasets requires reliable morphometry measurements from postmortem MRI via accurate 3D segmentation and reconstruction of the structures of interest. There has been substantial work in brain MRI parcellation such as  \emph{FreeSurfer} \cite{fischl2012freesurfer} and recent efforts based on deep learning \cite{henschel2020fastsurfer, chen2018voxresnet}. However, these approaches focus on antemortem MRI, and there is limited work on developing automated segmentation methods for postmortem MRI segmentation. Postmortem segmentation methods have been region specific. Recent developments include automated deep learning methods for high resolution cytoarchitectonic mapping of the occipital lobe in 2D histological sections [\cite{schiffer2021convolutional, spitzer2018improving, kiwitz2020deep, schiffer2021contrastive, schiffer20212d, eckermann2021three}]. The work by \cite{iglesias2015computational, iglesias2018probabilistic} has developed an atlas to segment the MTL and the thalamus using manual segmentations in postmortem images. Yet, a postmortem segmentation method applicable to a variety of brain regions has yet to be described. This is attributable to several factors. Some groups have developed robust whole brain postmortem image analysis tools \cite{jonkman2019normal, mancini2020multimodal, edlow20197}, though overall there is limited availability of postmortem specimens, scans, segmentation algorithms and labeled reference standard segmentations. Compared to antemortem structural MRI, postmortem MRI currently exhibits substantial heterogeneity in scanning protocols, larger image dimensions, greater textual complexity, and more profound artifacts. These issues can be addressed with new datasets and automated segmentation tools open to the public.

In this study, we expand upon our pilot study \cite{khandelwal2021gray, https://doi.org/10.1002/alz.062628, https://doi.org/10.1002/alz.065737} and develop a methodological framework to segment cortical gray matter; subcortical structures (caudate, putamen, globus pallidus, thalamus), white matter (WM) and WMH in  high resolution (0.3 x 0.3 x 0.3 mm$^{3}$) 7~T T2w postmortem MRI scans of whole brain hemispheres. We train and evaluate our approach using \textcolor{blue}{135} brain hemisphere scans from the Center for Neurodegenerative Disease Research of the University of Pennsylvania. We measure cortical thickness at several key locations in the cortex based on our automatic segmentation of the cortex, and correlate these measures with thickness measurements obtained using a user-guided semi-automated protocol. High consistency between these two sets of measures supports the use of deep learning-based automated thickness measures for postmortem brain segmentation and morphometry. We then report regional patterns of association between cortical thickness at a set of anatomical locations and neuropathology ratings (regional measures of p-tau, neuronal loss; as well as global  amyloid-$\beta$, Braak staging, and CERAD ratings) obtained from histology data and WMH burden. Additionally, we show that networks trained on T2-weighted spin echo images acquired at 7~T generalize to postmortem images obtained with T2*w gradient echo fast low angle shot (FLASH) 7T MRI acquired at \textcolor{blue}{resolution of 0.28 x 0.28 x 0.28 mm$^{3}$ and 0.16 x 0.16 x 0.16 mm$^{3}$.}

\section{Materials}
\label{section:materials}

\subsection{Donor Cohort}
\label{section:demographics}
\begin{figure}[H]
\centering
\includegraphics[width=\textwidth,height=\textheight,keepaspectratio]{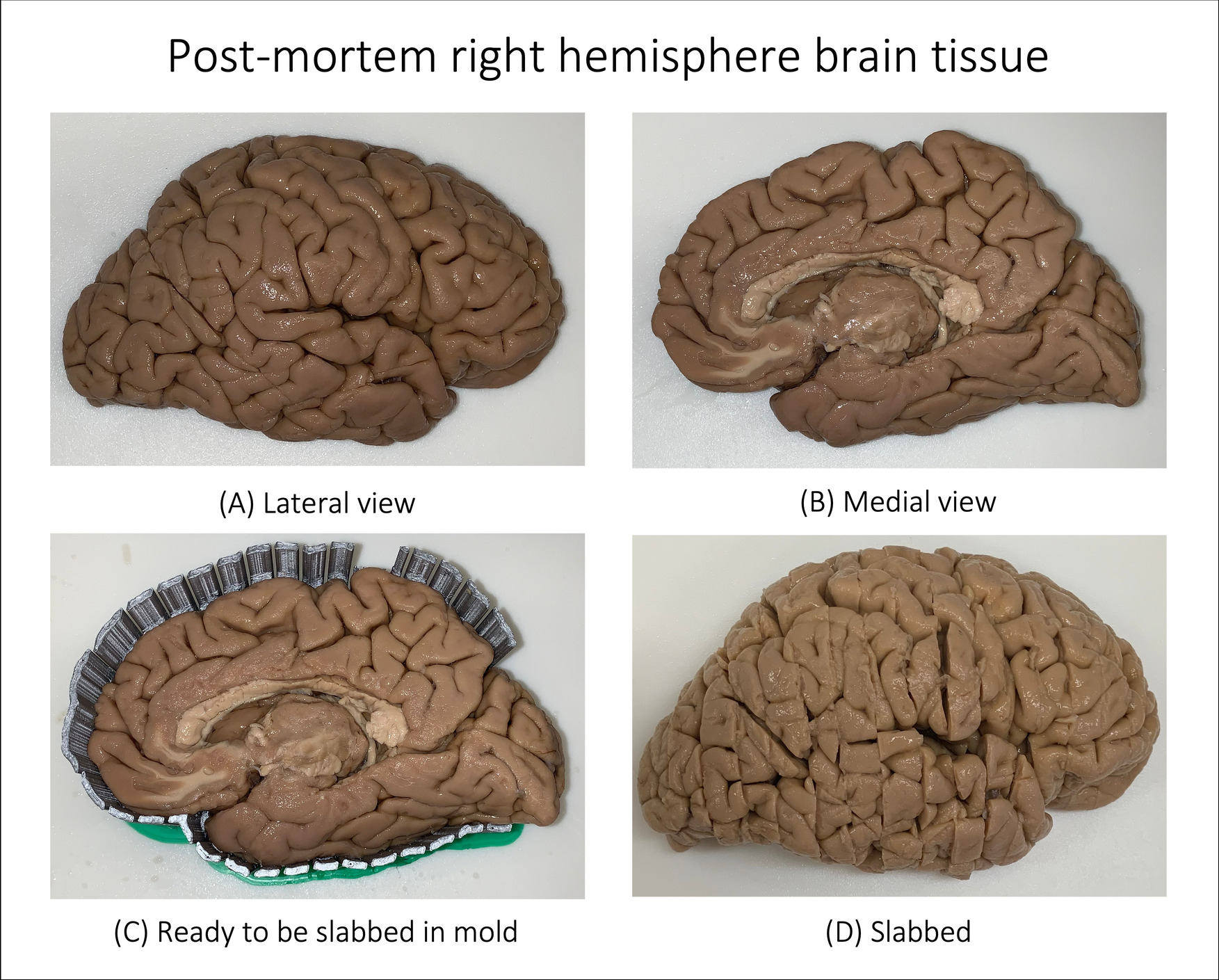}
        \caption{Post-mortem tissue blockface photograph of a donor with diagnosis of Parkinson's disease (not demented) and Lewy body disease (deceased at the age of 79). Shown are the lateral and medial views of the right hemisphere. The tissue is then placed in a mold \cite{jadeMold} and is subsequently slabbed as shown in the bottom right panel. See Supplementary Figure 1 for the slabbed slices of the given brain tissue.}
        \label{figure_brain_tissue}
\end{figure}

We analyze a dataset of \textcolor{blue}{135} postmortem whole-hemisphere MRI scans selected from Penn Integrated Neurodegenerative Disease Database (INDD) \cite{toledo2014platform}. Patients were evaluated at the Penn Frontotemporal Degeneration Center (FTDC) or Alzheimer’s Disease Research Center (ADRC) and followed to autopsy at the Penn Center for Neurodegenerative Disease Research (CNDR) as part of ongoing and previous clinical research programs \cite{tisdall2021joint, irwin2016progressive, arezoumandan2022regional}. The cohort included \textcolor{blue}{62} female (sex assigned at birth) donors (Age: \textcolor{blue}{75.37} $\pm$ \textcolor{blue}{10.02} years, Age range: \textcolor{blue}{53-97}) and \textcolor{blue}{73} male donors (Age: \textcolor{blue}{73.95} $\pm$ \textcolor{blue}{11.59} years, Age range: \textcolor{blue}{44-101}) with Alzheimer's Disease or related dementias (ADRD), such as Lewy body disease (LBD), FTLD-TDP43, progressive supranuclear palsy (PSP), or cognitively normal adults. Human brain specimens were obtained in accordance with local laws and regulations, and includes informed consent from next of kin at time of death. The patients were evaluated at FTDC and ADRC as per standard diagnostic criteria \cite{toledo2014platform}, and imaged by the teams at the ADRC and the Penn Image Computing and Science Laboratory (PICSL) and the FTDC. Autopsy was performed at the CNDR. Figure \ref{figure_brain_tissue} shows an example of a brain specimen with Parkinson's and LBD ready for autopsy, with the slabbed slices shown in Supplementary Figure 1. Separately, the post-mortem tissue photograph of a specimen with progressive non-fluent aphasia (PFNA) and Globular glial tauopathy (GGT) disease is shown in Supplementary Figure 2. Table \ref{table:demographics_summary} details the primary neuropathological diagnostic groups in the cohort with complete details tabulated in the \textcolor{blue}{Supplementary spreadsheet}.

\begin{table}[]
\caption{Demographics of the Alzheimer's Disease and Related Dementia (ADRD) patient cohort in the current study. Abbreviations: AD: Alzheimer's Disease, ALS: Amyotrophic lateral sclerosis, CVD: Cerebrovascular Disease, LATE: Limbic-predominant Age-related TDP-43 Encephalopathy, LBD: Lewy Body Disease, CTE:, FTLD-TDP: Frontotemporal lobar degeneration with TDP inclusions, GGT: Globular glial tauopathy, CVD: Cerebrovascular Disease, CBD: Corticobasal degeneration, PART: Primary age-related tauopathy, PSP: Progressive supranuclear palsy,  FUS: Fused-in-Sarcoma, tau-Misc: tauopathy unclassifiable.}
\label{table:demographics_summary}
\begin{tabular}{|ccccccccccc|}
\hline
\multicolumn{11}{|c|}{\textbf{\textcolor{blue}{Brain donor cohort}}} \\ \hline
\multicolumn{1}{|c|}{N} &
  \multicolumn{10}{c|}{135 (Female: 62 and Male: 73)} \\ \hline
\multicolumn{1}{|c|}{Age (years)} &
  \multicolumn{10}{c|}{74.60 $\pm$ 10.88 (range 44-101)} \\ \hline
\multicolumn{1}{|c|}{Race} &
  \multicolumn{10}{c|}{White: 125 Black: 8 More than one race: 1 Unknown: 1} \\ \hline
\multicolumn{1}{|c|}{Hemisphere imaged} &
  \multicolumn{10}{c|}{Right: 66 Left: 69} \\ \hline
\multicolumn{1}{|c|}{Post-mortem interval (hours)} &
  \multicolumn{10}{c|}{20.22 $\pm$ 15.05 (range: 3-105)} \\ \hline
\multicolumn{1}{|c|}{Fixation time (days)} &
  \multicolumn{10}{c|}{298.67 $\pm$ 338.60 (range: 29-1559)} \\ \hline  
\multicolumn{11}{|c|}{\textbf{Neuropathological diagnosis}} \\ \hline
\multicolumn{1}{|c|}{} &
  \multicolumn{5}{c|}{\textbf{Primary}} &
  \multicolumn{5}{c|}{\textbf{Secondary/ tertiary}} \\ \hline
  
\multicolumn{1}{|c|}{Alzheimer's disease} &
  \multicolumn{5}{c|}{53} &
  \multicolumn{5}{c|}{40} \\ \hline
  
\multicolumn{1}{|c|}{Lewy body disease} &
  \multicolumn{5}{c|}{25} &
  \multicolumn{5}{c|}{26} \\ \hline

\multicolumn{1}{|c|}{FTLD-TDP} &
  \multicolumn{5}{c|}{17} &
  \multicolumn{5}{c|}{-} \\ \hline

\multicolumn{1}{|c|}{Pick's disease} &
  \multicolumn{5}{c|}{9} &
  \multicolumn{5}{c|}{-} \\ \hline

\multicolumn{1}{|c|}{Progressive supranuclear palsy} &
  \multicolumn{5}{c|}{6} &
  \multicolumn{5}{c|}{1} \\ \hline

\multicolumn{1}{|c|}{Cerebrovascular disease} &
  \multicolumn{5}{c|}{5} &
  \multicolumn{5}{c|}{5} \\ \hline

\multicolumn{1}{|c|}{Corticobasal degeneration} &
  \multicolumn{5}{c|}{4} &
  \multicolumn{5}{c|}{1} \\ \hline

\multicolumn{1}{|c|}{Globular glial tauopathy} &
\multicolumn{5}{c|}{4} &
\multicolumn{5}{c|}{-} \\ \hline

\multicolumn{1}{|c|}{PART} &
  \multicolumn{5}{c|}{2} &
  \multicolumn{5}{c|}{10} \\ \hline

\multicolumn{1}{|c|}{LATE} &
  \multicolumn{5}{c|}{2} &
  \multicolumn{5}{c|}{15} \\ \hline

\multicolumn{1}{|c|}{Amyotrophic lateral sclerosis} &
  \multicolumn{5}{c|}{1} &
  \multicolumn{5}{c|}{1} \\ \hline
  
  \multicolumn{1}{|c|}{FTLD-FUS} &
  \multicolumn{5}{c|}{1} &
  \multicolumn{5}{c|}{-} \\ \hline
  
\multicolumn{1}{|c|}{\begin{tabular}[c]{@{}c@{}}FTD with Parkinsonism\\ (chromosome 17)\end{tabular}} &
  \multicolumn{5}{c|}{1} &
  \multicolumn{5}{c|}{-} \\ \hline 
  
\multicolumn{1}{|c|}{Argyrophilic grain disease} &
\multicolumn{5}{c|}{-} &
\multicolumn{5}{c|}{4} \\ \hline

\multicolumn{1}{|c|}{Hippocampal Sclerosis} &
  \multicolumn{5}{c|}{-} &
  \multicolumn{5}{c|}{5} \\ \hline

\multicolumn{11}{|c|}{\textbf{Global neuropathological staging}} \\ \hline
\multicolumn{1}{|c|}{\textbf{Ratings}} &
  \multicolumn{2}{c|}{\textbf{0}} &
  \multicolumn{3}{c|}{\textbf{1}} &
  \multicolumn{3}{c|}{\textbf{2}} &
  \multicolumn{2}{c|}{\textbf{3}} \\ \hline
\multicolumn{1}{|c|}{Amyloid-$\beta$ Thal staging (A score)} &
  \multicolumn{2}{c|}{26} &
  \multicolumn{3}{c|}{21} &
  \multicolumn{3}{c|}{32} &
  \multicolumn{2}{c|}{56} \\ \hline
\multicolumn{1}{|c|}{Braak three stage scheme (B score)} &
  \multicolumn{2}{c|}{14} &
  \multicolumn{3}{c|}{34} &
  \multicolumn{3}{c|}{27} &
  \multicolumn{2}{c|}{48} \\ \hline
\multicolumn{1}{|c|}{CERAD (C score)} &
  \multicolumn{2}{c|}{49} &
  \multicolumn{3}{c|}{20} &
  \multicolumn{3}{c|}{16} &
  \multicolumn{2}{c|}{50} \\ \hline
\end{tabular}
\end{table}

\begin{comment}
\multicolumn{1}{|c|}{\multirow{2}{*}{Six-stage Braak scheme}} &
  \multicolumn{1}{c|}{\textbf{0}} &
  \multicolumn{2}{c|}{\textbf{1}} &
  \multicolumn{1}{c|}{\textbf{2}} &
  \multicolumn{2}{c|}{\textbf{3}} &
  \multicolumn{1}{c|}{\textbf{4}} &
  \multicolumn{2}{c|}{\textbf{5}} &
  \textbf{6} \\ \cline{2-11} 
\multicolumn{1}{|c|}{} &
  \multicolumn{1}{c|}{13} &
  \multicolumn{2}{c|}{8} &
  \multicolumn{1}{c|}{26} &
  \multicolumn{2}{c|}{13} &
  \multicolumn{1}{c|}{14} &
  \multicolumn{2}{c|}{7} &
  41 \\ \hline
\end{comment}

\subsection{MRI Acquisition}
\label{section:MRI_Acquisition}
\begin{figure}[H]
\centering
\includegraphics[width=\textwidth,height=0.85\textheight,keepaspectratio]{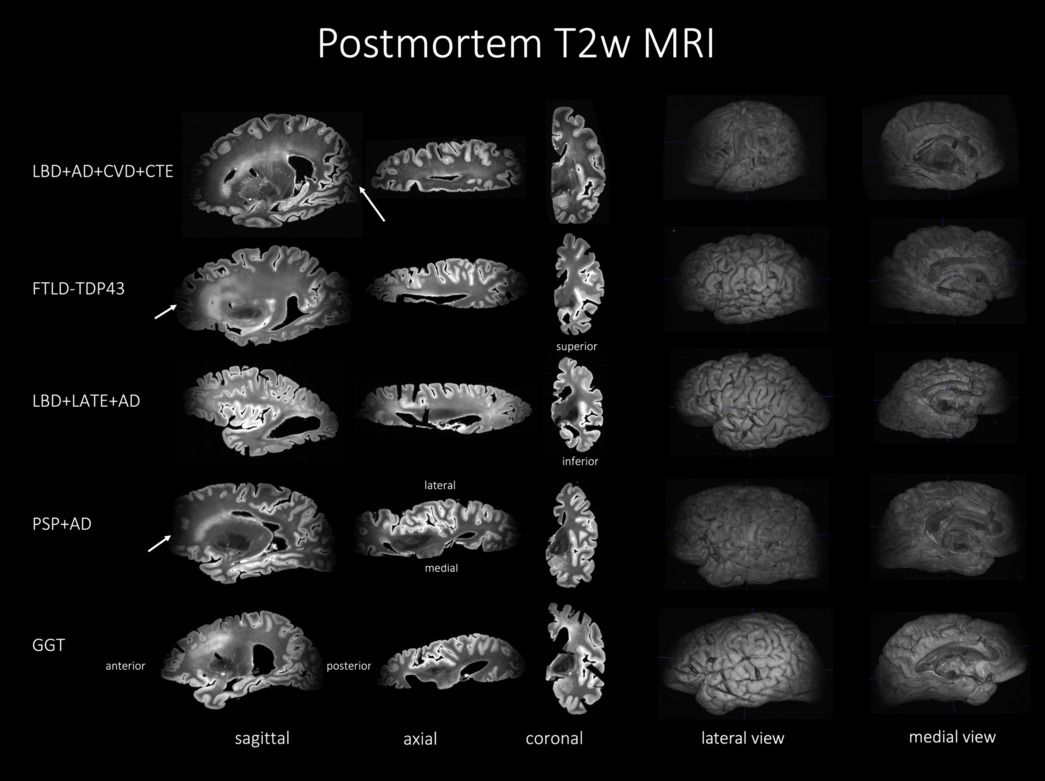}
        \caption{MRI of the T2w sequence representative of ADRD spectrum with mixed pathology and diagnoses of five subjects. The heterogeneity amongst the subjects can be appreciated through the three different viewing planes. Notice that the MRI signal drops off at the anterior and posterior ends of the hemisphere, a drawback of the current acquisition protocol. Also shown are the three-dimensional MRI renderings of the left and right brain hemispheres of two different subjects. Abbreviations: AD: Alzheimer's disease, ALS: amyotrophic lateral sclerosis, CVD: cerebrovascular disease, LATE: limbic-predominant age-related TDP-43 Encephalopathy, LBD: Lewy body disease, CTE: chronic traumatic encephalopathy, FTLD-TDP: frontotemporal lobar degeneration with TDP inclusions, GGT: globular glial tauopathy, CVD: cerebrovascular disease, CBD: corticobasal degeneration, PART: primary age-related tauopathy, PSP: progressive supranuclear palsy, FUS: fused-in-Sarcoma, tau-Misc: tauopathy unclassifiable.}
        \label{figure_MRI_2D_slices}
\end{figure}

During the autopsy of a specimen, one hemisphere was immersed in 10$\%$ neutral buffered formalin for at least 4 weeks prior to imaging. After the fixation time, samples were placed in Fomblin (California Vacuum Technology; Freemont, CA), enclosed in custom-built plastic bag holders. Samples were left to rest for at least 2 days to allow the air bubbles to escape from the tissue. The samples were scanned using either a custom-built small solenoid coil or a custom-modified quadrature birdcage (Varian, Palo Alto, CA, USA) coil  \cite{tisdall2021joint, edlow20197}. The samples were then placed into a whole-body 7~T scanner (MAGNETOM Terra, Siemens Healthineers, Erlangen, Germany). T2-weighted images were acquired using a 3D-encoded T2 SPACE sequence with 0.3 x 0.3 x 0.3 mm$^{3}$ isotropic resolution, 3 s repetition time (TR), echo time (TE) 383 ms, turbo factor 188, echo train duration 951 ms, bandwidth 348 Hz/px in approximately 2-3 hours per measurement. Image reconstruction was done using the vendor’s on-scanner reconstruction software which also corrected the global frequency drift, combined the signal averages in k-space, and produced magnitude images for each echo. A total of 4 repeated measurements were acquired for each sample and subsequently averaged to generate the final image. Sample MRI slices are shown in Figure \ref{figure_MRI_2D_slices} for a range of specimens with different diseases. The image acquisition suffers from geometric distortions due to the non-linearity of the magnetic gradient field that increases towards the ends (both anterior and posterior) of the sample and B1-transmit inhomogeneity, which results in decreased image quality as shown in Figure \ref{figure_MRI_2D_slices}.

\textcolor{blue}{We also acquire images at a much higher resolution using two separate T2*w FLASH MRI sequences, one at 280 microns and the other at 160 microns. We use these sequences for the generalization experiments as described in Section \ref{results:generalization}. Here we briefly describe their acquisition parameters. For the 280 microns isotropic resolution: MRI data were reacquired with a 3D-encoded, 8-echo gradient-recalled echo (GRE) sequence with non-selective RF pulses. To maintain readout polarity and minimize distortions due to field inhomogeneity, each readout was followed by a flyback rephrasing gradient. The final echo was followed by an additional completely rephrased readout to measure frequency drifts. Each line of k-space was acquired with multiple averages sequentially before advancing to the next phase-encode step. Common parameters for the sequence were: 280 microns isotropic resolution, 25$\degree$ flip angle, 60 ms repetition time (TR), minimum echo time (TE) 3.48 ms, echo spacing 6.62 ms, bandwidth 400 Hz/px. The field of view was adapted to each sample,and subsequently TRs and TEs were slightly modified based on the necessary readout duration. Total scan times were 8–10 hours for each sample.}

\textcolor{blue}{For the 160 microns isotropic resolution: MRI data was acquired with a 3D-encoded, 3-echo GRE sequence with non-selective RF pulses with the sequence: 25$\degree$ flip angle, 60 ms repetition time (TR), minimum echo time (TE), 9.37 ms, echo spacing 11.33 ms, bandwidth 90 Hz/px similar to methods described in \cite{tisdall2022ex}.}

\subsection{Neuropathological ratings}
\label{section:Neuropathological_Ratings}
\begin{figure}[H]
\centering
\includegraphics[width=1.25\textwidth,height=1.25\textheight,keepaspectratio]{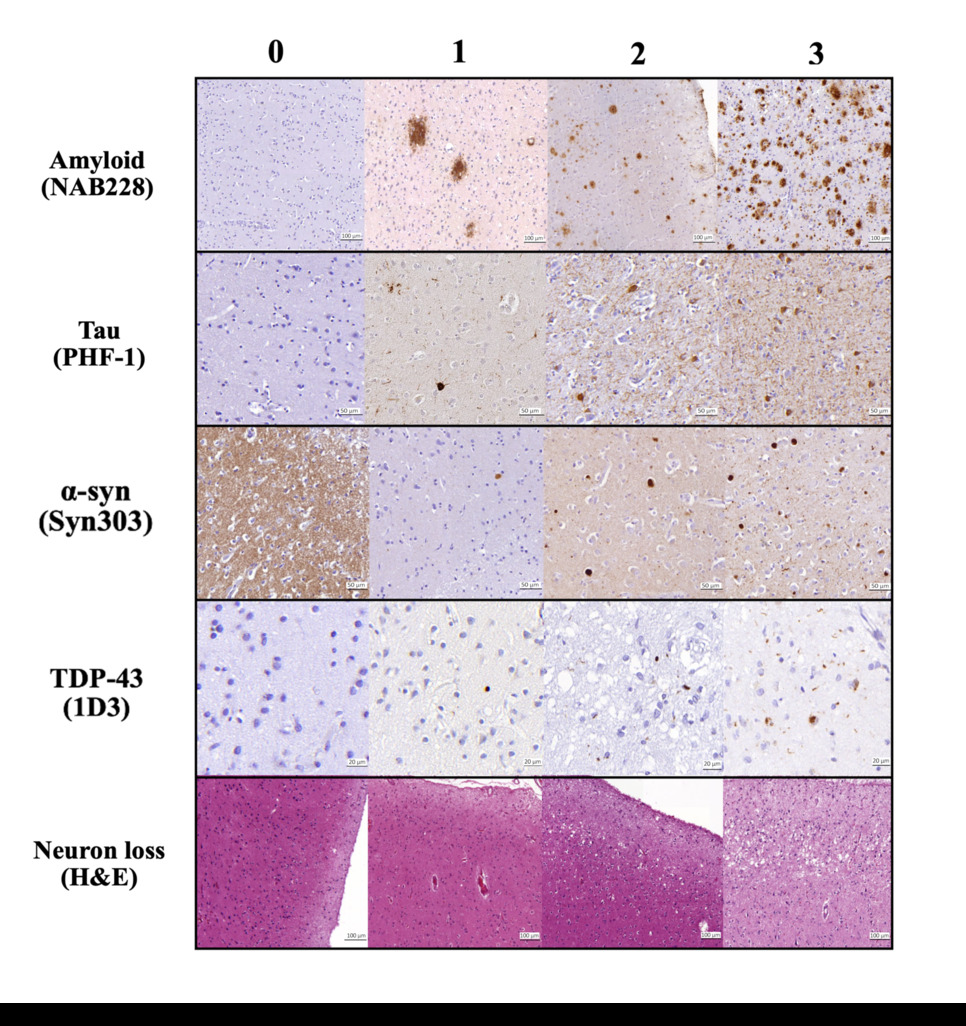} %March2023_figure_3_font
    \caption{Pathological assessment for each regional pathology derived from histology as per the protocol discussed in Section \ref{section:Neuropathological_Ratings}. The columns are severity ratings (left to right): 0-3. The shown pathologies are: amyloid-$\beta$ plaques (NAB228 antibody), p-tau pathology (PHF-1 antibody), $\alpha$-synuclein (Syn303 antibody), TDP-43 inclusions (pS409/410 antibody), and neuron loss (Hematoxylin and Eosin staining) in different cortical and medial temporal lobe regions included in the current work.}
\label{figure_neuropathologies_ratings_gradings}
\end{figure}

The non-imaged hemisphere, i.e., the contralateral tissue of each specimen systematically underwent histological processing for neuropathological examination at the CNDR \cite{toledo2014platform}. Roughly 1.5 × 1.5 × 0.5 cm$^{3}$ tissue blocks were extracted from the contralateral hemisphere. Paraffin-embedded blocks were then sectioned into 6 $\mu$m for immunohistochemistry using phosphorylated tau (AT8; Fisher, Waltham, MA; Catalogue No. ENMN1020) to detect phosphorylated p-tau deposits and p409/410 (mAb, 1:500, a gift from Dr. Manuela Neumann and Dr. E. Kremmer) to detect phosphorylated TDP-43 deposits. Immunohistochemistry evaluation was performed on the
hemisphere contralateral to the hemisphere that was scanned using
previously validated antibodies and established methods: NAB228
(monoclonal antibody [mAb], 1:8000, generated in the CNDR), phosphorylated tau PHF-1
(mAb, 1:1000, a gift from Dr. Peter Davies), TAR5P-1D3 (mAb, 1:500,
a gift from Dr. Manuela Neumann and Dr. E. Kremmer), and Syn303
(mAb, 1:16,000, generated in the CNDR) to detect amyloid $\beta$ (A$\beta$)
deposits, phosphorylated tau (p-tau) deposits, phosphorylated TDP-43
deposits, and the presence of pathological conformation of $\alpha$-synuclein, respectively. In 16 cortical regions, semi-quantitative ratings of p-tau, TDP-43, $\beta$-amyloid, and $\alpha$-synuclein pathology, as well as neuronal loss, were visually assigned by an expert neuropathologist (E.L. and J.Q.T.) on a scale of 0–3 i.e. “0: None”, “0.5: Rare”, “1: Mild”, “2: Moderate” or “3: Severe" \cite{toledo2014platform}. These ratings are illustrated in Figure \ref{figure_neuropathologies_ratings_gradings}. Global ratings of neurodegenerative disease progression were also derived, including A, B, C scores \cite{hyman2012national}. Supplementary Table 2 details the locations from where the neuropathology ratings were obtained from, either the exact (main regions) or the closest (exploratory regions), to the cortical brain regions.

\subsection{Localized thickness measurement pipeline at key cortical locations}
\label{section:dots_thickness_pipeline}
\begin{figure}[H]
\centering
\includegraphics[width=\textwidth,height=\textheight,keepaspectratio]{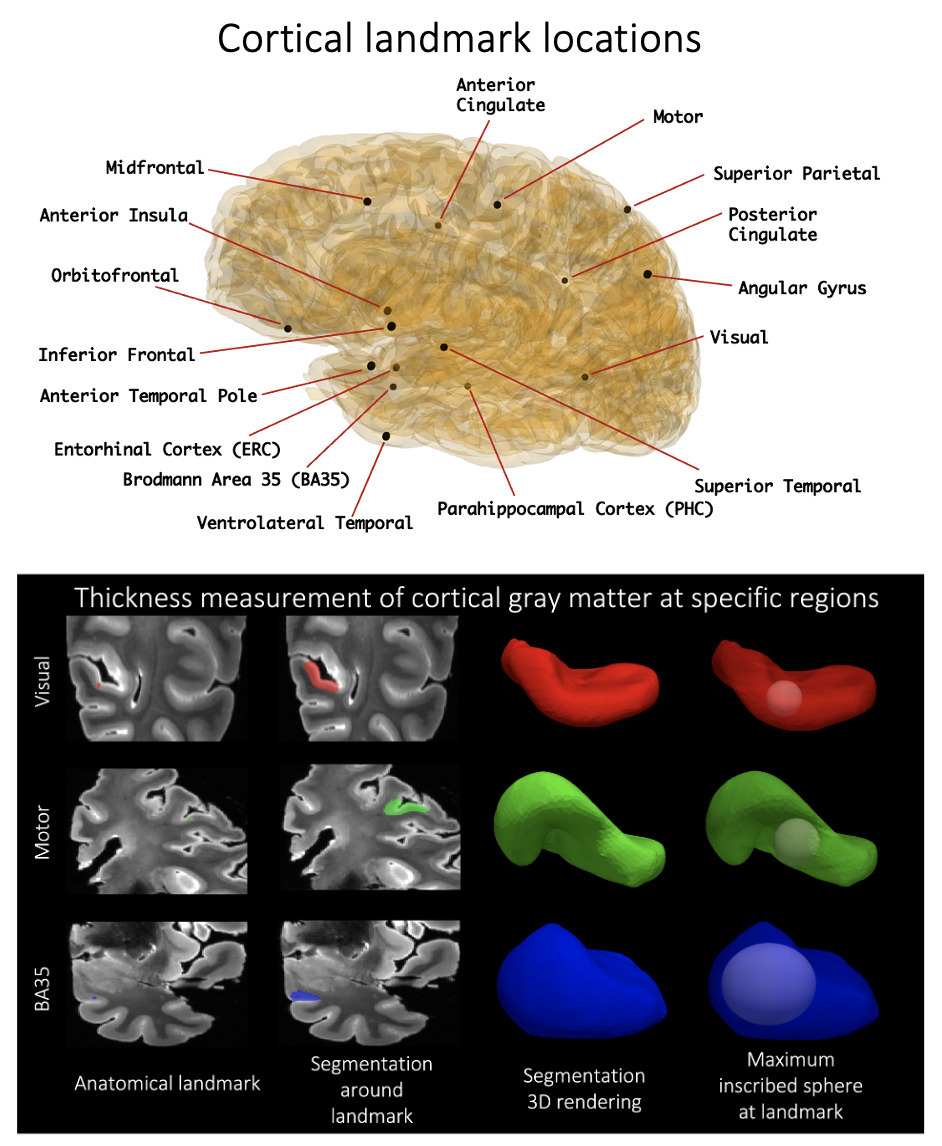}
        \caption{Cortical thickness is measured at the 16 landmarks. A dot (shown here: motor, visual, BA35) is first placed to define an anatomical landmark, around which a semi-automatic level set segmentation of the surrounding cortical ribbon is provided. A maximally inscribed sphere is then computed using Voronoi skelentonization \cite{ogniewicz1992voronoi}, and the diameter of the sphere gives thickness at that landmark.}
        \label{figure_cortical_dots_locations}
\end{figure}

Our center has adopted an expert-supervised semi-automatic protocol to obtain localized quantitative measures of cortical thickness in all 7~T postmortem MRI scans, as described in the work by \cite{sadaghiani2022associations} Section 2.3 and Supplementary material S1, \cite{wisse2021downstream} Section: MTL thickness measurements and Supplementary material: Thickness measurement. In the current study, we use these measures as the reference standard for evaluating automated cortical segmentation. In each hemisphere, 16 cortical landmarks are identified and labeled on the MRI scan as shown in Figure \ref{figure_cortical_dots_locations} A. To measure cortical thickness at these locations, a semi-automatic level set segmentation of the surrounding cortical ribbon is performed and the maximal sphere inscribed into the cortical segmentation and containing the landmark is found; the diameter of this sphere gives thickness at that landmark (Figure \ref{figure_cortical_dots_locations} B).

\section{Methods}
\label{ch:methodology}
In Section \ref{section:ManualSegmentationProtocols}, we describe the manual segmentation protocols developed for cortical gray matter, the four subcortical structures, WMH, and normal-appearing white matter. Then, in sections \ref{sec:methods_segmentation_cortical} and \ref{sec:methods_segmentation_other}, we describe the deep learning-based pipeline and performance evaluation criteria developed for the automated segmentation of these structures. In Section \ref{sec:methods_segmentation_topology}, we describe the post-hoc topology correction step employed to provide geometrically accurate segmentations. Finally, in section \ref{section:neuropath_morph}, we describe the statistical models used to link neuropathological ratings of p-tau, amyloid-$\beta$ neuronal loss, CERAD scores, Braak Staging with morphometry based on volume and regional thickness measurements obtained from the automated segmentations.

\subsection{Manual segmentation protocols}
\label{section:ManualSegmentationProtocols}
We developed protocols to manually segment structures in postmortem MRI: cortical gray matter, subcortical structures (caudate, putamen, globus pallidus, thalamus), WM, and WMH. We used these manual segmentations along with the corresponding MR images to train the neural networks. All manual segmentations were done by raters in ITK-SNAP \cite{yushkevich2019user}. Supplementary Table 1 shows which subjects were manually segmented to obtain the training data and the reference standard for the different labels, and the subsequent cross-validation studies as described in Sections \ref{sec:methods_segmentation_cortical} and \ref{sec:methods_segmentation_other}.

\subsubsection{Cortical Gray Matter}
\label{section:manual_cortical_gray_matter}
\begin{figure}[H]
\centering
\includegraphics[width=\textwidth,height=\textheight,keepaspectratio]{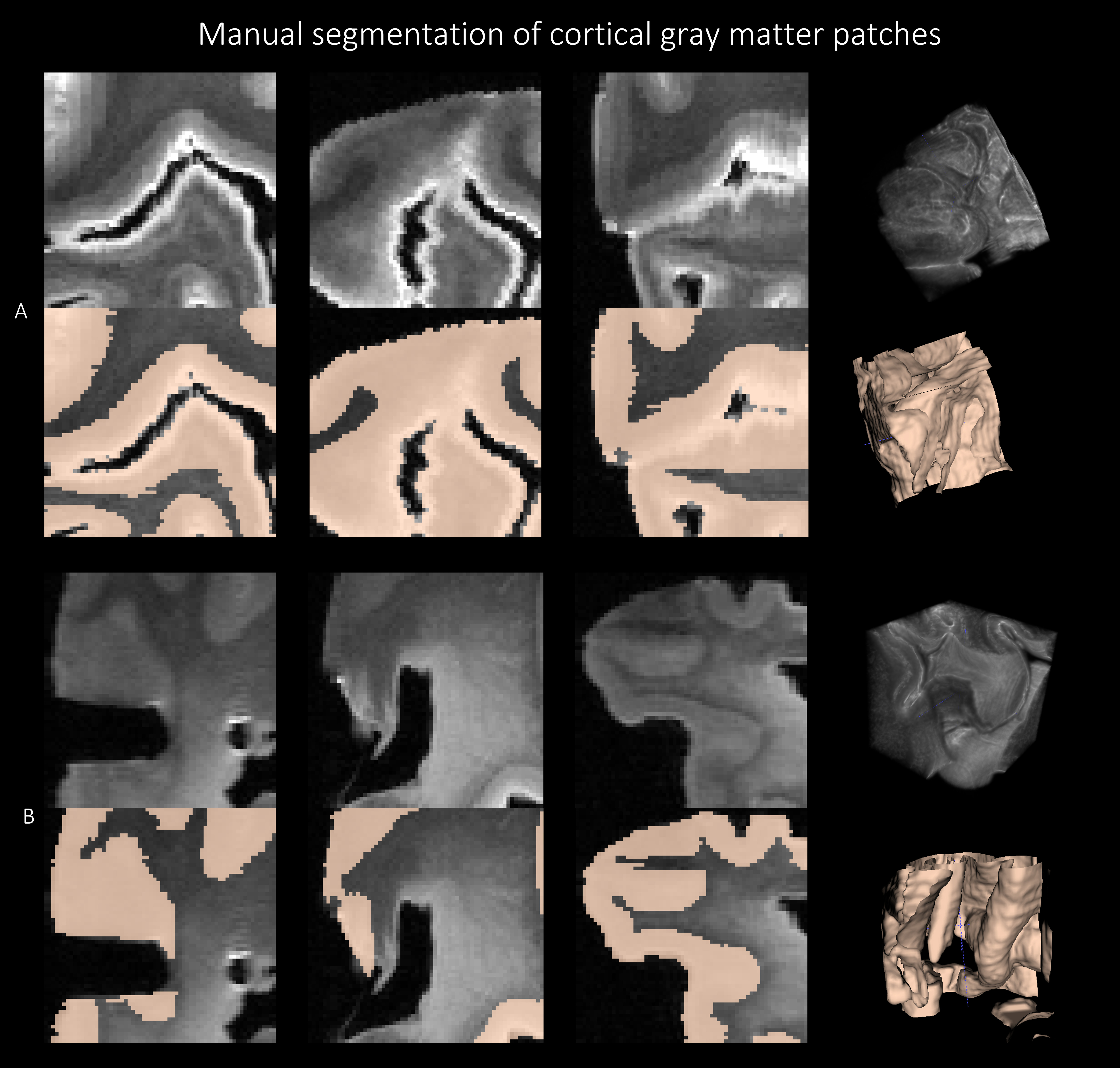}
        \caption{Manual segmentation of the cortical gray matter patches of size  64 x 64 x 64 using the protocol explained in \ref{section:manual_cortical_gray_matter} for two subjects with FTLD-TDP (top) and GGT (bottom) in three viewing planes with the corresponding 3D renderings.}
        \label{figure_manual_cortex}
\end{figure}

We sampled five 3D image patches of size 64 x 64 x 64 voxel$^{3}$, as shown in Figure \ref{figure_manual_cortex} around the orbitofrontal cortex (OFC), anterior temporal cortex (ATC), inferior prefrontal cortex (IPFC), primary motor cortex (PMC), and primary somatosensory cortex (PSC) from 6 brain hemispheres, resulting in a total of 30 patches. These regions were selected as representative regions with variable levels of pathology in ADRD cases and control regions (PMC, PSC) \cite{tisdall2021joint}. For example, OFC, ATC and IPFC have high pathology in PSP, whereas, the PSC, which is generally less-affected in most neurodegenerative diseases, was sampled as a negative control. In each patch, gray matter was segmented manually in ITK-SNAP. Five manual raters, divided into groups of two (E.K. and G.C.) and three (E.H, B.L.P. and E.M.), labeled gray matter as the foreground, and rest of the image as the background using a combination of manual tracing and the semi-automated segmentation tool. The manual segmentation of cortical gray matter was supervised by author P.K. Figure \ref{figure_manual_cortex} shows sample patch images and the corresponding reference standard labels with 3D renderings. We followed some guiding principles for manual segmentation: (1) we note that the white layer enveloping the cortex is not an imaging artifact, but is the outermost layer of the cortex, and thus is labeled as the gray matter, (2) adjoining gyri in deep sulci region are correctly labeled as gray matter and demarcates the deep sulci as the background. (3) in several regions, gray matter have similar intensities with the nearby WMH which were resolved by visual inspection of texture in the surrounding region, (4) the 64x64x64 patches provided little context when segmenting the GM, therefore, the corresponding whole hemisphere image was displayed on a separate ITK-SNAP window for the user to examine the structures surrounding the given image patch. Inter-rater reliability scores were computed for these manual segmentations in terms of Dice Coefficient (DSC): Raters 1\&2: 95.26 $\pm$ 1.37 \%, Raters 1\&3: 94.64 $\pm$ 1.64 \%, Raters 2\&3: 94.54 $\pm$ 1.20 \%, Raters 4\&5: 92.04 $\pm$ 4.26 \%.

\subsubsection{Subcortical structures}
\label{section:manual_segm_protocol_subcort}
Four subcortical gray matter structures (caudate, putamen, globus pallidus, thalamus) were manually segmented on 7 specimens, selected to span varying pathological diagnoses (including Alzheimer’s spectrum, p-tauopathies, TDP-43 encephalopathies, LBD, cerebrovascular disease and mixed disease), the range of postmortem age and levels of cerebral atrophy, tissue quality and tearing, mineral deposition and calcification, blood vessel size, vascular pathology on imaging and autopsy. Figure \ref{figure_manual_segm_subcort_wmh} shows an example manual segmentation of the four subcortical structures. First, structures were segmented across the axial plane based on several anatomical considerations. (1) The borders of the caudate (head, body and tail) were defined by the frontal horns of the lateral ventricles (cerebrospinal fluid, anterior border) and the internal capsule (WM, boundaries of the body). Periventricular WM disease adjacent to the head of the caudate was excluded from the caudate segmentation. (2) The borders of the putamen were determined by the globus pallidus (medial) and the claustrum (lateral). (3) The globus pallidus was annotated to include the pallidum interna and externa, bounded by WM surrounding the thalamus and subthalamic nucleus. Mineral deposition (such as calcium) was noted by areas of heterogeneous T2 hypointensities and predominantly localized to the lentiform nucleus; where needed, the contours of some segmentations were adjusted to include these regions of heterogeneous signal within the lentiform structures. (4) The thalamus was segmented as the medial gray matter bounded by the subthalamic nucleus, midbrain, mamillary bodies, corpus callosum, lateral ventricles and caudate. After segmentation in the axial plane, volumes were edited in the coronal and sagittal planes to ensure smoothness of the segmentation across all three planes. Boundaries between structures (such as striations between caudate and putamen) were agreed upon among authors. Manual segmentation of the four subcortical structures was performed by author, E.C. and supervised by and edited by M.T.D. Subcortical segmentations were discussed in consensus meetings with P.K., P.A.Y, S.R.D and D.A.W.

\subsubsection{White Matter Hyperintensities}
\label{section:manual_segm_protocol_wmh}
Nine specimens were chosen to segment WMHs across a gamut of vascular pathology with differing levels of WMH appearance, including focal small-vessel ischemia to intermediate periventricular and juxtacortical patterns to large, diffuse cerebrovascular disease. General principles for segmentation were applied as follows. (1) segmented lesions should be generally larger than 1 cm$^{3}$, (2) segmentations should appear for at least 4-6 slices to be a 1 cm$^{3}$ region, (3) segmented WMH should include both periventricular (anterior frontal and posterior temporal/occipital horns of the lateral ventricles) and juxtacortical lesions. This distinguished WMH from insular cortex, claustrum, basal ganglia and other gray matter structures embedded in WM. (4) WMH segmentations generally exhibited signal intensity above a threshold of $\geq$450-550 intensity, (but this was influenced by field inhomogeneity artifacts, either between images or within the same image, often at the anterior and posterior cortical poles), given the image intensity range was normalized between 0-1000. (5) WMH segmentations included T2 hyperintense perivascular spaces and cortical venules but must also include surrounding T2 hyperintense white matter regions that occupy a region larger than 1 cm$^{3}$. (6) WMH was segmented primarily in axial plane and then assessed for contiguity and smoothness in sagittal and axial planes as well as 3D renderings. Manual segmentation was performed by the author A.D. and supervised and edited by M.T.D. WMH segmentations were discussed in consensus meetings with P.K., P.A.Y, S.R.D and D.A.W.
Figure \ref{figure_manual_segm_subcort_wmh} shows an example manual segmentation of the WMH.

\begin{figure}[H]
\centering
    \includegraphics[width=\textwidth,height=\textheight,keepaspectratio]{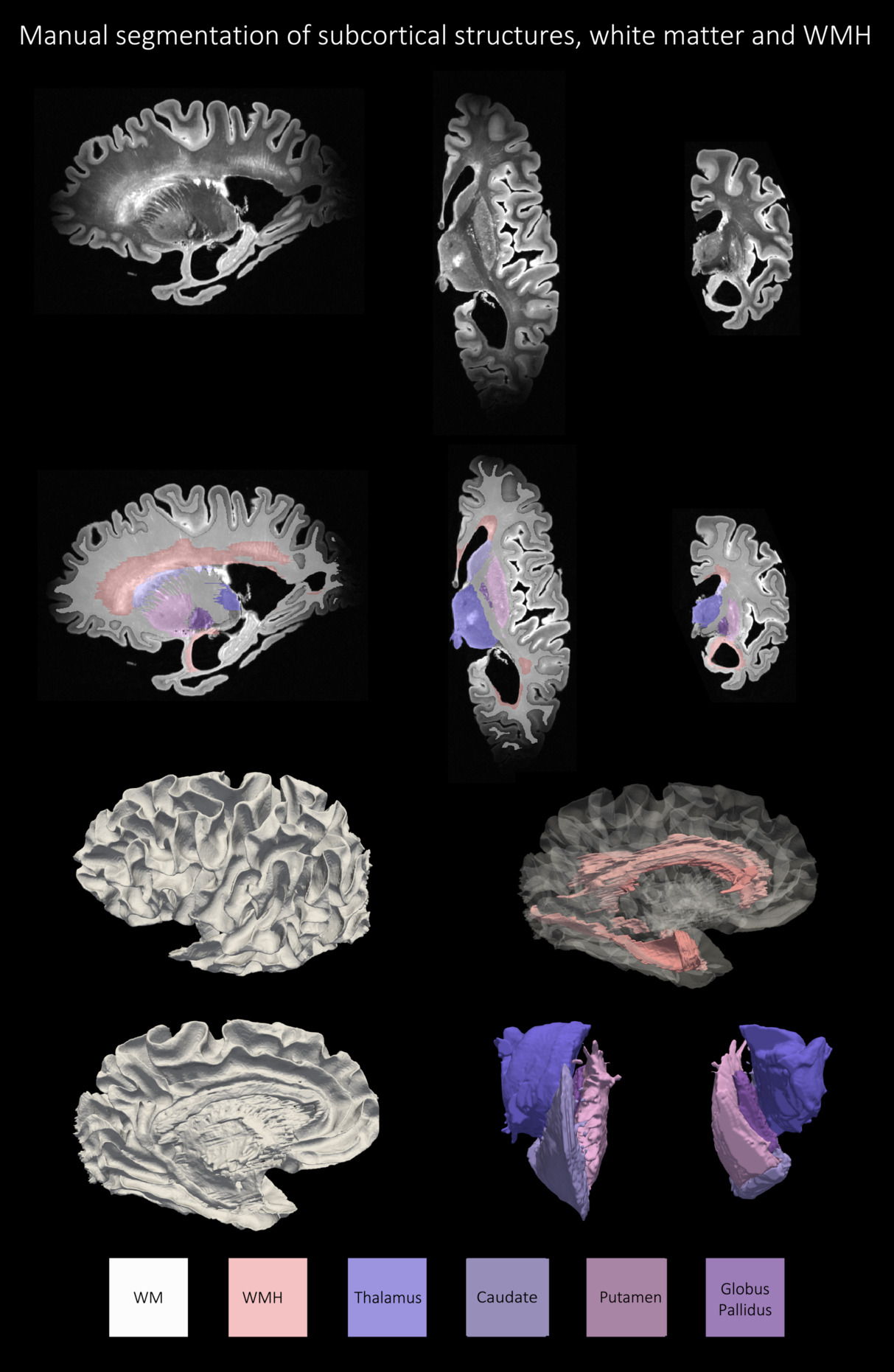} %March2023_Figure5.jpg
        %dec_manual_segm_resized again_manual_subcort_wmh.jpg
        \caption{An example of manual segmentation of the thalamus, putamen, caudate, globus pallidus and white matter hyperintensities as per the protocol described in Sections and \ref{section:manual_segm_protocol_subcort} and \ref{section:manual_segm_protocol_wmh} for a subject with Alzheimer's Disease and Lewy body disease (87 years old).}
        \label{figure_manual_segm_subcort_wmh}
\end{figure}

\subsubsection{White Matter}
Two specimens with manually labelled subcortical structures and WMH, but automated segmentations for cortical mantle were thresholded to obtain the WM label, which was then manually corrected by the author A.D., supervised by P.K., to remove incorrectly labeled spurious voxels to fill-in holes and thereby define a clear GM/WM boundary.

\subsection{Automated deep learning-based segmentation for cortical gray matter}
\label{sec:methods_segmentation_cortical}
We first developed an approach for labeling cortical gray matter in postmortem 7~T MRI scans using the 30 patch-level cortical gray matter segmentations from six subjects (described in Section \ref{section:manual_cortical_gray_matter}) for training and cross-validation. Given the exceptional performance of convolutional neural network (CNN) models for antemortem medical image segmentation \cite{schiffer2021convolutional}, our effort focused on benchmarking leading existing CNN models, rather than developing another model for the postmortem cortical segmentation task. We benchmarked the following variants of popular biomedical image segmentation deep learning models: (1) nnU-Net \cite{isensee2021nnu}; four variants of AnatomyNet \cite{zhu2019anatomynet} based on squeeze-and-excitation blocks \cite{roy2018recalibrating}: (2)  Channel Squeeze and Spatial Excitation AnatomyNet (SE), (3)  Spatial Squeeze and Channel Excitation AnatomyNet (CE), (4) AnatomyNet (Vanilla), (5)  Spatial and Channel Squeeze and Excitation AnatomyNet (SE + CE); (6) 3D Unet-like network \cite{khandelwal2020domain}; (7) VoxResNet \cite{chen2018voxresnet}; (8) VNet \cite{milletari2016v} ; and (9) Attention Unet \cite{oktay2018attention}. Supplementary material provides architectural details of the nine deep neural networks. We use PyTorch 1.5.1 and Nvidia Quadro RTX 5000 GPUs to train the models using user-annotated images described in Sections \ref{section:manual_cortical_gray_matter} - \ref{section:manual_segm_protocol_wmh}. To train the deep neural networks, the input images were standardized, and then normalized between 0 and 1.

To ensure a fair and systematic evaluation of the nine networks, we trained and evaluated the nine network architectures within the nnU-Net framework under matched conditions (i.e., same split of data into training/validation/testing subsets; same data augmentation strategy, same hyper-parameter tuning strategy). We first evaluated the accuracy of cortical segmentation by six-fold cross-validation on the patch-level manual segmentations. We report average DSC and average Hausdorff Distance 95th percentile (HD95) across the 30 segmented patches in the six-fold cross-validation experiments. 

\textcolor{blue}{To choose the best performing model for subsequent analyses, the nine trained models were then used to segment the whole hemisphere cortical mantle across a subset of the cohort (N=36 for which the user-supervised semi-automated segmentations are available). We used these segmentations to compute the thickness around the 16 landmarks in these 36 subjects at the cortical landmarks. We then compared the thickness measurements to corresponding measures obtained via user supervised semi-automated segmentations (reference standard) using our protocol (Section \ref{section:dots_thickness_pipeline}).} For each of the 16 landmarks, we report Spearman's correlation coefficient between the automated and the reference manual segmentation-based cortical thickness measurements, and the average fixed raters intra-class correlation coefficient (ICC). Based on the combination of cross-validation DSC accuracy, agreement between the automatic and manual segmentation-based cortical thickness measures, and overall visual inspection of the segmentations, a single model was selected for subsequent experiments. As reported in Section \ref{results:segmentation}, this best-performing method was the \textbf{nnU-Net} model.

\subsection{Automated deep learning-based segmentation for other structures}
\label{sec:methods_segmentation_other}
We trained another vanilla nnU-Net model to obtain the automated segmentation of subcortical structures, WMH and normal-appearing WM. \textcolor{blue}{In particular, for the purpose of five-fold cross-validation studies to report DSC and HD95 scores, we trained a nnU-Net model based on the manually-labeled training data for the segmentation of subcortical structures in 7 subjects and WMH in 9 subjects; and the manually labelled WM and the automated cortical mantle segmentations obtained from the nnU-Net model trained in the previous Section \ref{sec:methods_segmentation_cortical} as additional labels. Supplementary Table 1 lists the subjects used for training and evaluation of all the seven labels.}

\subsection{Post-hoc automated topology correction of segmentations}
\label{sec:methods_segmentation_topology}
\textcolor{blue}{As shown in Figure \ref{figure_MRI_2D_slices}, the image acquisition suffers from geometric distortions due to the non-linearity of the magnetic gradient field and might also suffer from partial volume averaging which generally introduces holes and handles in the segmentation and makes the WM surface to lack a spherical topology. Furthermore, the opposing banks of sulcus often appear to be fused, which makes it harder for the segmentation algorithm to correctly segment the cortex and thus might produce erroneous cortical thickness values. Hence, it is important to correct such mis-segmentations by enforcing a topology correction method. Therefore, we employ the methods developed in CRUISE: Cortical reconstruction using implicit surface evolution \cite{han2004cruise} made available as part of the `\emph{nighres}' package, with default settings, \cite{huntenburg2018nighres} for post-hoc topology correction of WM surface and constraining the GM segmentations. In particular, the method uses a fast marching-based method \cite{bazin2007topology}, a graph-based topology correction algorithm (GTCA) \cite{han2002topology} and a topology preserving geometric deformable model (TGDM) \cite{han2003topology} to rectify holes and handles, thus producing a WM surface with a spherical topology. Separately, the Anatomically Consistent Enhancement (ACE) method is used to provide a GM representation that has evidence of sulci where it might not otherwise exist due to the partial volume effect. Thus we obtain geometrically accurate and topologically correct segmentations for the cortical GM and WM. In the current study, we apply this topology correction step on the automated GM/WM segmentation labels obtained from the nnU-Net model trained in the previous Section \ref{sec:methods_segmentation_other}. We called the post-hoc topology corrected model as ``nnU-Net-CRUISE".}

\subsection{Linking neuropathology ratings with morphometry}
\label{section:neuropath_morph}
We computed Spearman's correlation between thickness measurements obtained from automated segmentations at the 16 anatomical landmarks described above with semi-quantitative pathology ratings from approximately corresponding anatomical locations in the contralateral hemisphere (regional p-tau score, regional neuronal loss) and global pathology ratings (CERAD stage, Braak stage, $\beta$-amyloid). We repeat this analysis with thickness measurements obtained from user-annotated manual reference segmentations, and thereby test the hypothesis that similar trends would be seen between pathology correlations with automated and manual thickness measures, which, in turn, would imply that automated segmentations are viable for morphometry-based studies (Section \ref{results:morphometry_neuropath}). In particular, we follow the experimental design from our recent work \cite{sadaghiani2022associations} for a subset of the cohort \textcolor{blue}{within the AD spectrum having AD as their primary diagnosis and also diagnosis of either: LBD, PART, LATE, CBD (N=82) out of 135. The criteria for AD continuum are based on excluding cases with diagnoses of FLTD or non-AD tauopathy (whether primary or secondary). Finally, we normalized the WMH volumes by the corresponding WM volumes, and then computed one-sided Spearman correlation between the normalized WMH volume with regional cortical thickness and subcortical volumes for the subjects within the AD spectrum (N=82). We include nuisance covariates of age, sex and postmortem interval (PMI) in all of our analyses.}

\section{Results}

\subsection{Cortical Gray Matter Segmentation}
\label{results:segmentation}
\subsubsection{Dice coefficient volume overlap and qualitative analysis}

\begin{table}[H]
    \caption{Six-fold cross validation Dice Coefficient (DSC) and Hausdorff Distance 95th percentile (HD95) scores between reference standard and automated patch-level cortical segmentations.}
    \label{table:results}
    \centering
    %\resizebox{2.0in}{2.0in}
        {\begin{tabular}{c|c|c}
        \makecell{\textbf{Deep learning method}} & \makecell{\textbf{DSC} \textbf{(\%)}} & \makecell{\textbf{HD95} \textbf{(mm)}} \\
    \hline
    nnU-Net &
    93.98 $\pm$ 5.25 & 
    0.49 $\pm$ 0.45 \\
    \hline
    AnatomyNet (SE) & 
    94.84 $\pm$ 3.84 & 
    0.45 $\pm$ 0.42 \\
    \hline
    AnatomyNet (CE) & 
    94.91 $\pm$ 3.27 & 
    0.45 $\pm$ 0.42 \\
    \hline
    AnatomyNet (Vanilla) & 
    94.86 $\pm$ 3.83 & 
    0.46 $\pm$ 0.44 \\
    \hline
    AnatomyNet (CE + SE) & 
    94.66 $\pm$ 3.79 & 
    0.47 $\pm$ 0.44 \\
    \hline
    3D Unet & 
    93.57 $\pm$ 5.22 & 
    0.58 $\pm$ 0.51 \\
    \hline
    VoxResNet & 
    94.84 $\pm$ 4.00 & 
    0.45 $\pm$ 0.42 \\    
    \hline
    VNet & 
    90.84 $\pm$ 5.93 & 
    0.99 $\pm$ 0.56 \\
    \hline
    Attention Unet & 
    93.65 $\pm$ 4.91 & 
    0.62 $\pm$ 0.66
  \end{tabular}}
\end{table}

Table \ref{table:results} tabulates the \textit{patch}-level cortical gray matter segmentation performance of the nine different networks across six-fold cross validation. AnatomyNet and its variants attain the highest patch-level DSC, closely followed by VoxResNet. The nnU-Net model has slightly lower DSC than the best AnatomyNet model, but the difference is less that $1\%$. However, since the patches used to train the segmentation networks were only sampled from select regions of the hemispheres, cross-validation accuracy on these patches is not necessarily indicative of the networks' ability to generalize to other brain regions. Figure \ref{figure_2D_nnunet_two} illustrates, that consistently across our specimens, AnatomyNet and its variants are able to distinguish gray matter from white matter in high-contrast regions, but fail to segment the anterior and posterior regions where contrast is lower due to limitations of the MRI coil. There is also some systematic under-segmentation (see white arrows) of the cortex even in higher-contrast regions (see the white circled regions in Figure \ref{figure_2D_nnunet_two}). By contrast, nnU-Net clearly demarcates GM/WM boundary even in low-contrast regions, which is remarkable considering that these regions were not captured by the training patches.

\begin{figure}[H]
\centering
\includegraphics[width=\textwidth,height=\textheight,keepaspectratio]{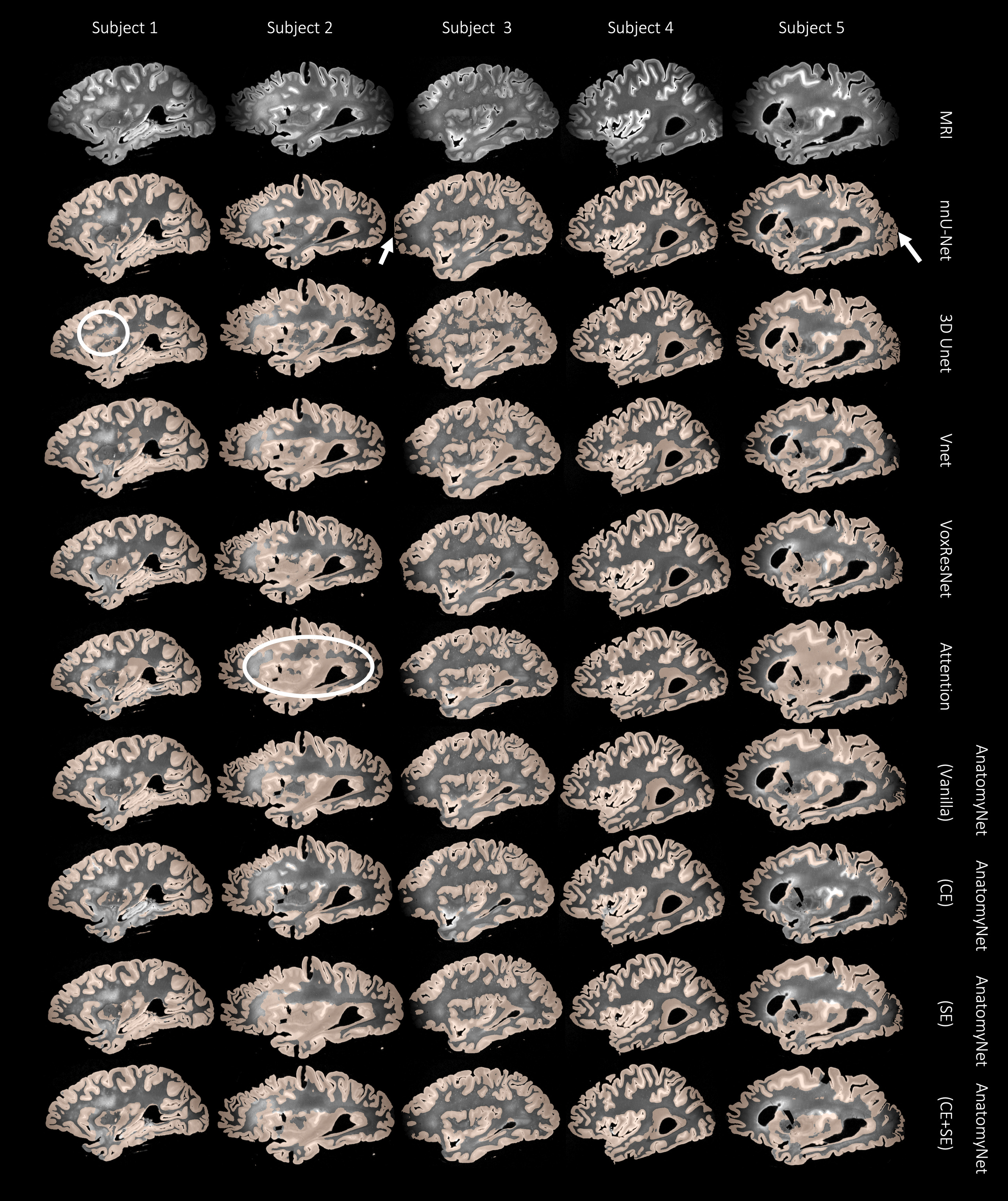} %dec_nine_archs_1_edit_reszied again_updated_2D_nnuet_all_archs.jpg
        \caption{For a subset of 10 subjects, each row shows an example sagittal view of the automated cortical mantle segmentation predicted across the whole brain hemisphere using the different neural network architectures. We notice that all architectures except nnU-Net shows either under- or over-segmentation of the cortical gray matter, which, together with the results reported in Table 3 (ICC results) prompted us to select nnU-Net as the preferred model for cortical gray matter segmentation. For example, notice that how networks such as 3D Unet and Attention Unet incorrectly segments large chunks of WM as cortical GM (white circles). Whereas, nnU-Net performs the best in difficult to segment areas such as the anterior and posterior regions of the brain with poor MRI signal.}
        \addtocounter{figure}{-1} 
\end{figure}

\begin{figure}[H]
\centering
\includegraphics[width=\textwidth,height=\textheight,keepaspectratio]{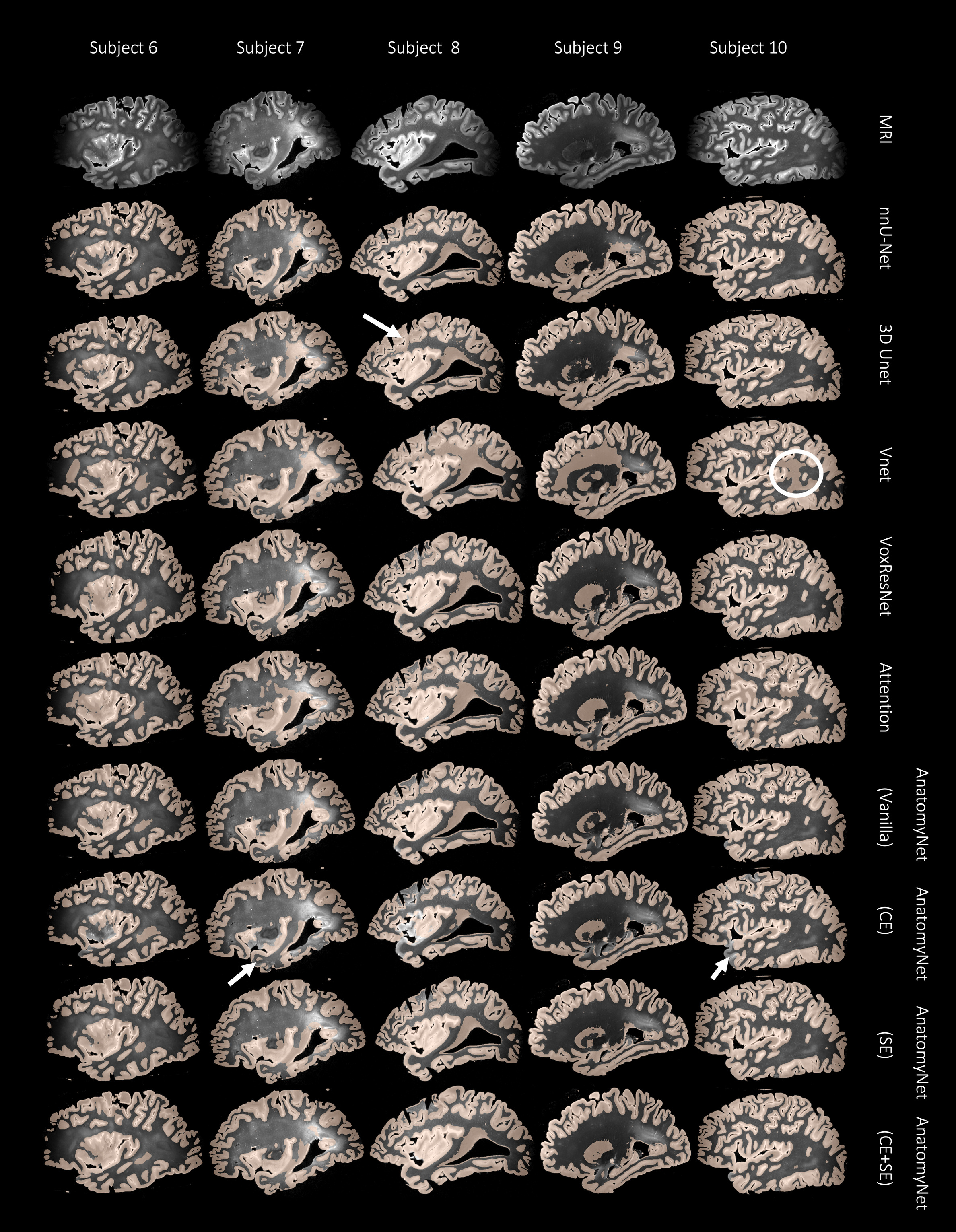} 
%dec_nine_archs_2_edit_reszied again_updated_2D_nnuet_all_archs.jpg
        \caption{\textit{\textbf{Continued.}} Again, notice that AnatomyNet (CE) is not able to segment certain areas in the cortical mantle (white arrows). The 3D Unet and Vnet networks oversegments the cortical mantle into the WM. However, nnU-Net overcomes all of these limitations.}
        \label{figure_2D_nnunet_two}
\end{figure}

\subsubsection{Intra-class correlation coefficient}
\label{results:corr_bland_atman_icc}
We compare cortical thickness (mm) at 16 cortical landmarks between automated gray matter segmentations obtained by the nine networks and the user-supervised semi-automated segmentation method, which serves as the reference standard.

\textcolor{blue}{Table 3 tabulates the Average fixed raters ICC scores for all the nine networks trained as described in \ref{sec:methods_segmentation_cortical}. We observe that nnU-Net (mean ICC=0.72) is clearly the best amongst the nine patch-based models. The final nnU-Net-CRUISE model which has been topologically corrected as described in Sections \ref{sec:methods_segmentation_other} and \ref{sec:methods_segmentation_topology} slightly exceeds the patch-based vanilla nnU-Net (mean ICC of 0.73) on average.} The variants of AnatomyNet have mean ICC values of, Vanilla: 0.40 , CE: 0.47, CE+SE: 0.40 and SE: 0.34. We observe that the variants of AnatomyNet were the top performing models when evaluated using DSC scores at patch-level, but did not generalize to robustly segment the entire cortical mantle; and thereby failed to show good correlation of regional cortical thickness when compared with the reference standard reference cortical thickness. The other four models also had very low ICC values: VoxResNet: 0.45, VNet: 0.28, 3D Unet: 0.47, and Attention Unet: 0.35.

The Bland-Altman plots in Figure \ref{fig:Bland_Altman} shows strong agreement between reference standard and \textcolor{blue}{post-hoc topologically corrected} automated nnU-Net segmentation-based thickness measurements for the 16 cortical landmarks.  \textcolor{blue}{Furthermore Supplementary Figure 3 shows that 13 out of the 16 regions have correlation coefficient (r) greater than 0.6. We also observe high ICC scores with 12 regions having ICC greater than 0.7}. These results confirm that automated segmentations are accurate to give desirable cortical thickness measurements.

Therefore, based on quantitative evaluation in terms of DSC and HD95 scores, ICC values, and the qualitative visual inspection of the segmentations for the different neural network architectures, we conclude that nnU-Net-CRUISE is the best performing model. \textcolor{blue}{Figure \ref{figure_topo} shows the post-hoc topology corrected automated nnU-Net segmentations of cortical gray matter and white matter shown in sagittal view for five randomly chosen subjects. The first columns shows the MRI slice, the automated nnU-Net segmentation before and after topology correction step are shown in columns 2 and 3. Columns 4 and 5 show the zoomed-in area with the red arrows indicating the regions where topology correction improves the segmentation. We notice that the opposite banks of sulci are no longer fused and in fact well demarcated after correcting for topology. These segmentations will therefore provide more reliable and accurate estimates of cortical thickness.}

\begin{figure}[H]
\centering
\includegraphics[width=\textwidth,keepaspectratio]{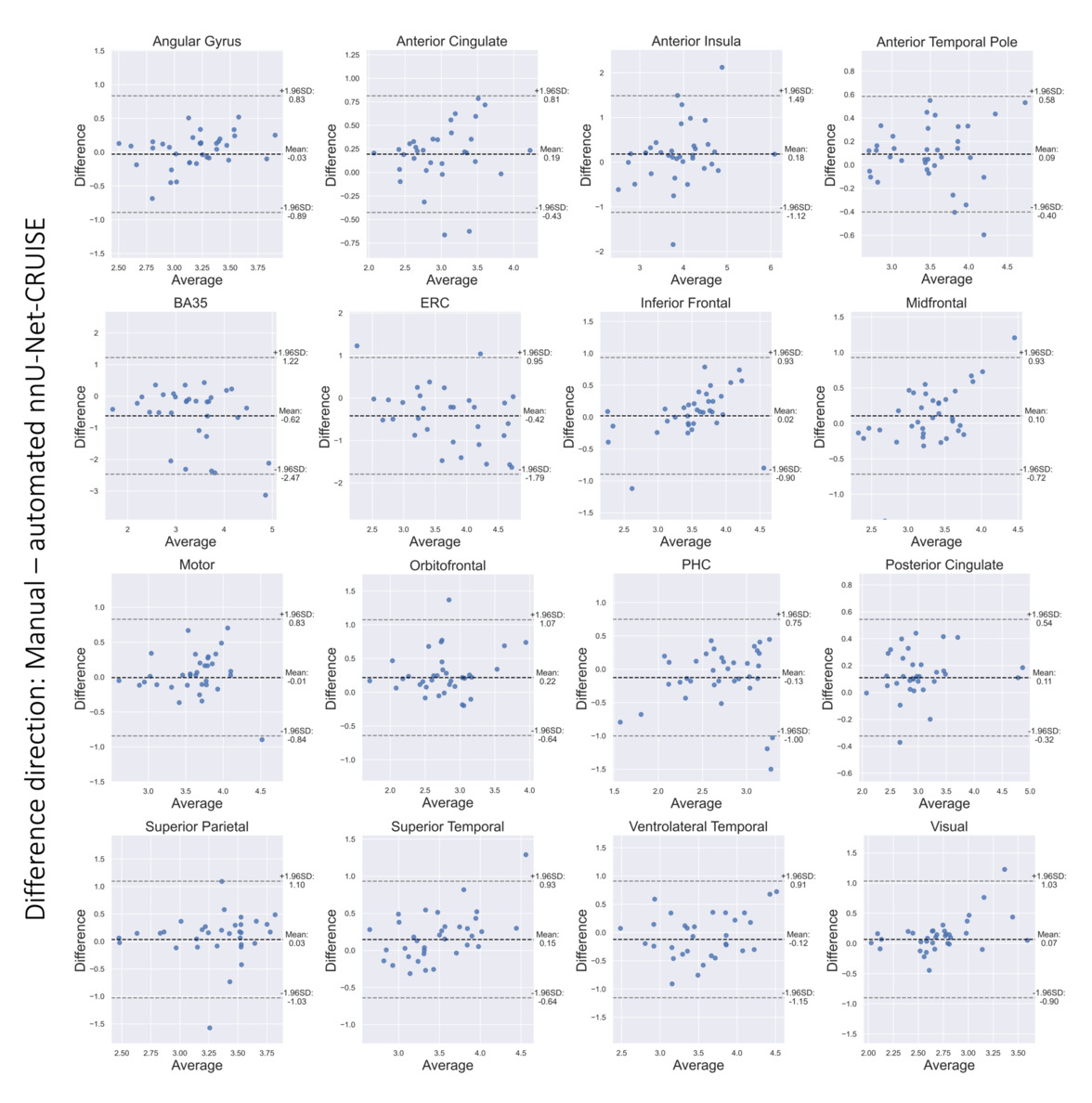}
        \caption{\textcolor{blue}{Bland-Altman plots for the cortical regions. We observe good agreement between the manual and nnU-Net-CRUISE automated segmentations-based thickness, which supports the hypothesis that topologically corrected deep learning-based automated segmentations are reliable for morphometric measurements. Note that the highlighted differences are in the direction: manual - automated segmentations. Supplementary Figure 3 shows the corresponding correlation plots.}}
        \label{fig:Bland_Altman}
\end{figure}

\subsection{Other structures: Subcortical, WMH and WM segmentation}
Based on the superior performance of nnU-Net, as explained in the previous section, we employ nnU-Net to perform the  multi-label segmentation for WMH, caudate, putamen, globus pallidus, thalamus and WM. The mean DSC scores for the four subcortical structures and WMH across all leave-one-out cross-validation experiments were: WMH: 79.70 \%, caudate: 88.18 \%, putamen: 85.20 \%, globus pallidus: 80.12 \%, and the thalamus: 87.29 \%. Note that currently, we do not have a large sample size to perform cross-validation evaluation on the WM, due to the time-consuming process of manually correcting the segmentations. \textcolor{blue}{Obtaining manual WM label is beyond the scope of the current study}. Figures \ref{figure_2D_segmentations_automated} and \ref{figure_3D_segmentations_automated} show qualitative results for all the segmented structures in 2D and 3D respectively for the the 36 subjects used for quantitative evaluation in the previous section, in sagittal view.

%%%%%
\begin{adjustbox}{angle=90, caption={Average fixed raters Intra-class correlation coefficient (ICC) scores for the regional cortical thickness measurements between automated nnU-Net and reference standard manual segmentations for all the nine neural network architectures \textcolor{blue}{, and the topologically corrected nnU-Net-CRUISE segmentations}. Across each row, each cell is color coded, with darker shades indicating higher ICC value.},float=table}
\scalebox{0.7}
{
\centering
  \label{table:ICC_all_networks}
  \begin{tabular}{|c|c|c|c|c|c|c|c|c|c|c|}
  \hline
\textbf{ROI\textbackslash Network} &
  \textbf{VoxResNet} &
  \textbf{VNet} &
  \textbf{3D Unet} &
  \textbf{\begin{tabular}[c]{@{}c@{}}Attention\\ UNet\end{tabular}} &
  \textbf{\begin{tabular}[c]{@{}c@{}}AnatomyNet\\ (SE)\end{tabular}} &
  \textbf{\begin{tabular}[c]{@{}c@{}}AnatomyNet\\ (CE+SE)\end{tabular}} &
  \textbf{\begin{tabular}[c]{@{}c@{}}AnatomyNet\\ (Vanilla)\end{tabular}} &
  \textbf{\begin{tabular}[c]{@{}c@{}}AnatomyNet\\ (CE)\end{tabular}} &
  \textbf{nnU-Net} &
  \textbf{nnU-Net-CRUISE} \\ \hline
  \textbf{Visual}                                  & \gradientICC{0.40}          & \gradientICC{0.27}          & \gradientICC{0.39}        & \gradientICC{0.40}          & \gradientICC{0.32}          & \gradientICC{0.41}          & \gradientICC{0.34}          & \gradientICC{0.53}          & \gradientICC{0.54}        & \gradientICC{0.55}   \\ \hline
  \textbf{Motor}                                   & \gradientICC{0.59}          & \gradientICC{0.03}          & \gradientICC{0.25}        & \gradientICC{0.03}          & \gradientICC{0.11}          & \gradientICC{0.04}          & \gradientICC{0.08}          & \gradientICC{0.46}          & \gradientICC{0.72}     & \gradientICC{0.75}      \\ \hline
  \textbf{Posterior cingulate}                     & \gradientICC{0.57}          & \gradientICC{0.51}          & \gradientICC{0.46}        & \gradientICC{0.39}          & \gradientICC{0.46}          & \gradientICC{0.68}          & \gradientICC{0.62}          & \gradientICC{0.88}          & \gradientICC{0.92}    & \gradientICC{0.96}       \\ \hline
  \textbf{Midfrontal}                              & \gradientICC{0.63}          & \gradientICC{0.55}          & \gradientICC{0.32}        & \gradientICC{0.54}          & \gradientICC{0.64}          & \gradientICC{0.73}          & \gradientICC{0.72}          & \gradientICC{0.61}          & \gradientICC{0.84}     & \gradientICC{0.79}      \\ \hline
  \textbf{Anterior cingulate}                      & \gradientICC{0.86}          & \gradientICC{0.67}          & \gradientICC{0.72}        & \gradientICC{-0.12}         & \gradientICC{0.34}          & \gradientICC{0.56}          & \gradientICC{0.51}          & \gradientICC{0.70}          & \gradientICC{0.94}     & \gradientICC{0.88}      \\ \hline
  \textbf{Orbitofrontal}                           & \gradientICC{0.66}          & \gradientICC{0.44}          & \gradientICC{0.48}        & \gradientICC{0.62}          & \gradientICC{0.69}          & \gradientICC{0.61}          & \gradientICC{0.64}          & \gradientICC{0.51}          & \gradientICC{0.69}     & \gradientICC{0.75}      \\ \hline
  \textbf{Superior temporal}                       & \gradientICC{0.52}          & \gradientICC{0.27}          & \gradientICC{0.44}        & \gradientICC{0.34}          & \gradientICC{0.34}          & \gradientICC{0.35}          & \gradientICC{0.38}          & \gradientICC{0.30}          & \gradientICC{0.54}    & \gradientICC{0.79}       \\ \hline
  \textbf{Inferior frontal}                        & \gradientICC{0.13}          & \gradientICC{-0.01}         & \gradientICC{0.28}        & \gradientICC{0.16}          & \gradientICC{-0.17}         & \gradientICC{-0.09}         & \gradientICC{0.34}          & \gradientICC{0.37}          & \gradientICC{0.46}    & \gradientICC{0.79}       \\ \hline
  \textbf{Anterior insula}                         & \gradientICC{-0.14}         & \gradientICC{0.44}          & \gradientICC{0.50}        & \gradientICC{0.08}          & \gradientICC{0.36}          & \gradientICC{0.22}          & \gradientICC{0.30}          & \gradientICC{-0.16}         & \gradientICC{0.67}   & \gradientICC{0.77}        \\ \hline
  \textbf{Anterior temporal}                       & \gradientICC{0.53}          & \gradientICC{0.28}          & \gradientICC{0.52}        & \gradientICC{0.58}          & \gradientICC{0.24}          & \gradientICC{0.12}          & \gradientICC{0.20}          & \gradientICC{-0.21}         & \gradientICC{0.82}   & \gradientICC{0.94}        \\ \hline
  \textbf{Ventrolaterl}                            & \gradientICC{0.27}          & \gradientICC{0.50}          & \gradientICC{0.40}        & \gradientICC{0.39}          & \gradientICC{0.08}          & \gradientICC{0.64}          & \gradientICC{0.31}          & \gradientICC{0.60}          & \gradientICC{0.87}   & \gradientICC{0.71}        \\ \hline
  \textbf{Superior pareital}                       & \gradientICC{0.78}          & \gradientICC{0.46}          & \gradientICC{0.48}        & \gradientICC{0.46}          & \gradientICC{0.81}          & \gradientICC{0.71}          & \gradientICC{0.81}          & \gradientICC{0.82}          & \gradientICC{0.70}   & \gradientICC{0.79}        \\ \hline
  \textbf{Angular gyrus}                           & \gradientICC{0.39}          & \gradientICC{0.22}          & \gradientICC{0.25}        & \gradientICC{0.25}          & \gradientICC{0.33}          & \gradientICC{0.26}          & \gradientICC{0.45}          & \gradientICC{0.44}          & \gradientICC{0.86}    & \gradientICC{0.51}       \\ \hline
  \textbf{ERC}                                     & \gradientICC{0.27}          & \gradientICC{-0.13}         & \gradientICC{0.79}        & \gradientICC{0.57}          & \gradientICC{0.12}          & \gradientICC{0.47}          & \gradientICC{0.11}          & \gradientICC{0.64}          & \gradientICC{0.62}    & \gradientICC{0.75}       \\ \hline
  \textbf{BA35}                                    & \gradientICC{0.53}          & \gradientICC{0.56}          & \gradientICC{0.79}        & \gradientICC{0.56}          & \gradientICC{0.61}          & \gradientICC{0.52}          & \gradientICC{0.54}          & \gradientICC{0.56}          & \gradientICC{0.73}    & \gradientICC{0.59}       \\ \hline
  \textbf{PHC}                                     & \gradientICC{0.15}          & \gradientICC{0.41}          & \gradientICC{0.36}        & \gradientICC{0.42}          & \gradientICC{0.21}          & \gradientICC{0.23}          & \gradientICC{0.11}          & \gradientICC{0.49}          & \gradientICC{0.59}    & \gradientICC{0.73}       \\ \hline
  \textbf{Mean}                                    & \textbf{\gradientICC{0.45}} & \textbf{\gradientICC{0.28}} & \textbf{\gradientICC{0.47}} & \textbf{\gradientICC{0.35}} & \textbf{\gradientICC{0.34}} & \textbf{\gradientICC{0.40}} & \textbf{\gradientICC{0.40}} & \textbf{\gradientICC{0.47}} & \textbf{\gradientICC{0.72}} & \textbf{\gradientICC{0.73}} \\ \hline
\end{tabular}
}
\end{adjustbox}

\begin{figure}[H]
\centering
\includegraphics[width=\textwidth,height=\textheight,keepaspectratio]{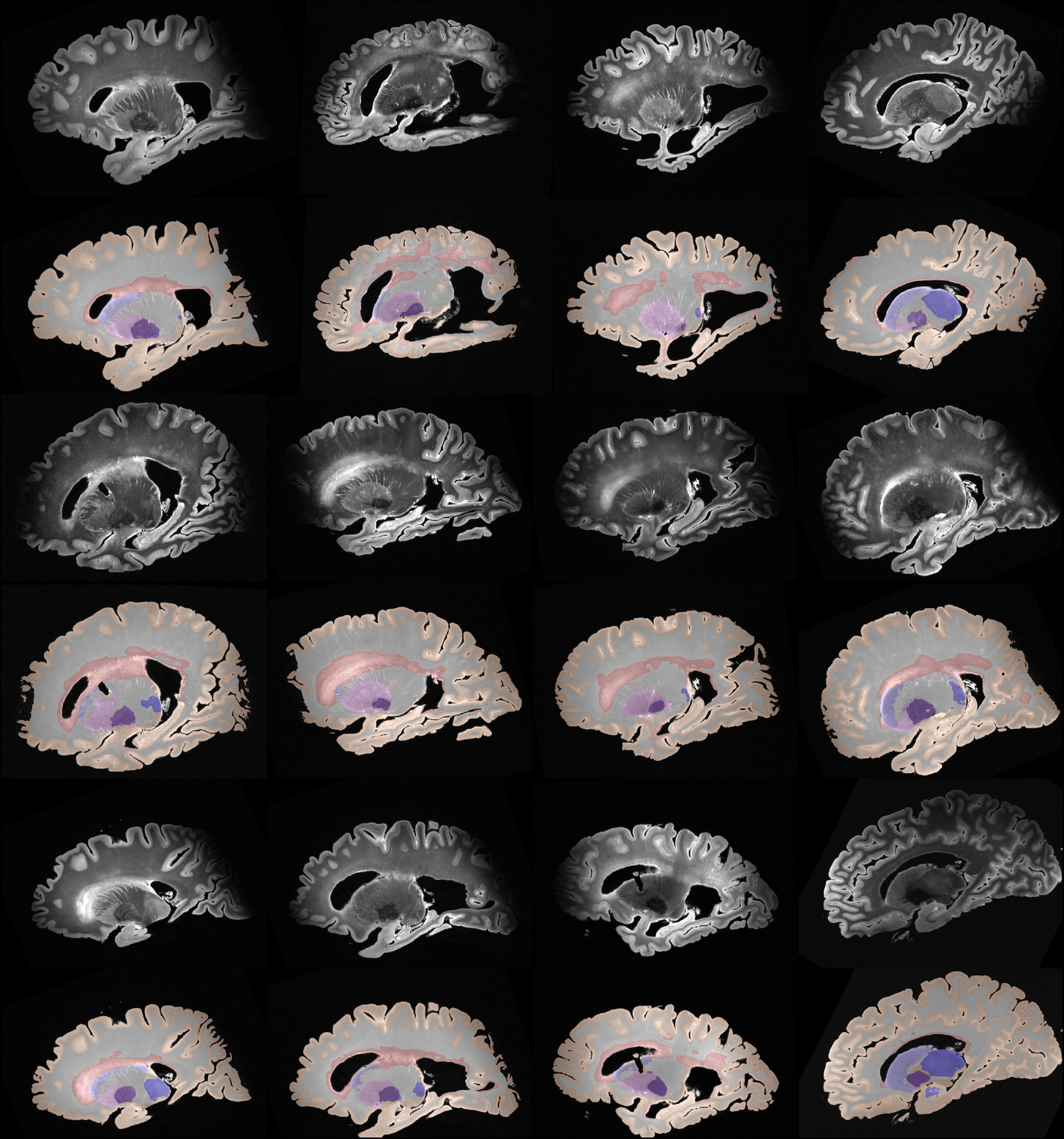} %dec_nnunet_all_1
        \caption{Automated segmentations by nnU-Net of cortical gray matter, WMH, WM and the four subcortical structures for a sample of 36 subjects in the cohort.}
        \addtocounter{figure}{-1} 
\end{figure}

\begin{figure}[H]
\centering
\includegraphics[width=\textwidth,height=\textheight,keepaspectratio]{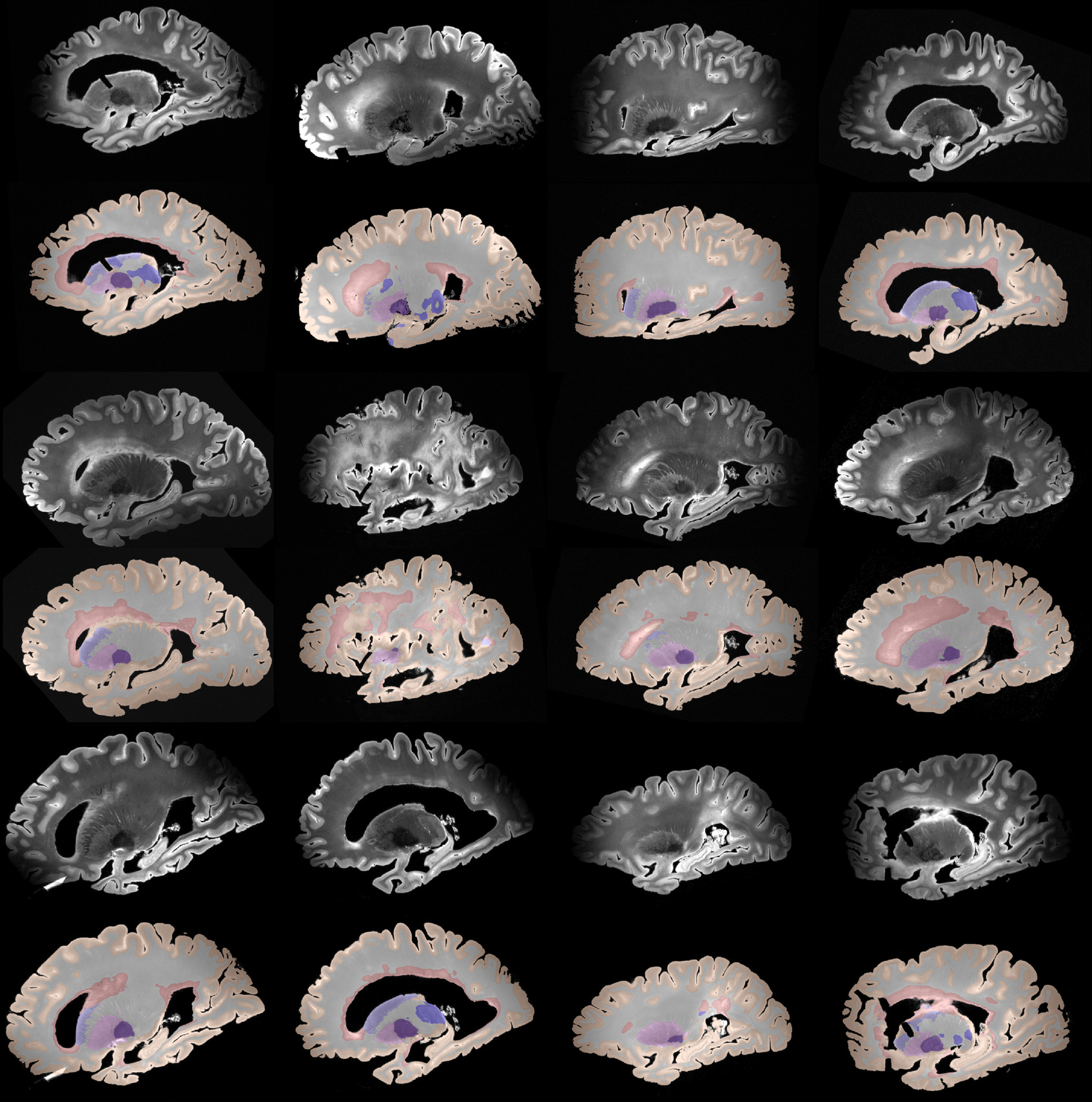} %dec_nnunet_all_2
        \caption{\textit{\textbf{Continued}}. Automated segmentations by nnU-Net of cortical gray matter, WMH, WM and the four subcortical structures for a sample of 36 subjects in the cohort.}
        \addtocounter{figure}{-1} 
\end{figure}

\begin{figure}[H]
\centering
\includegraphics[width=\textwidth,height=\textheight,keepaspectratio]{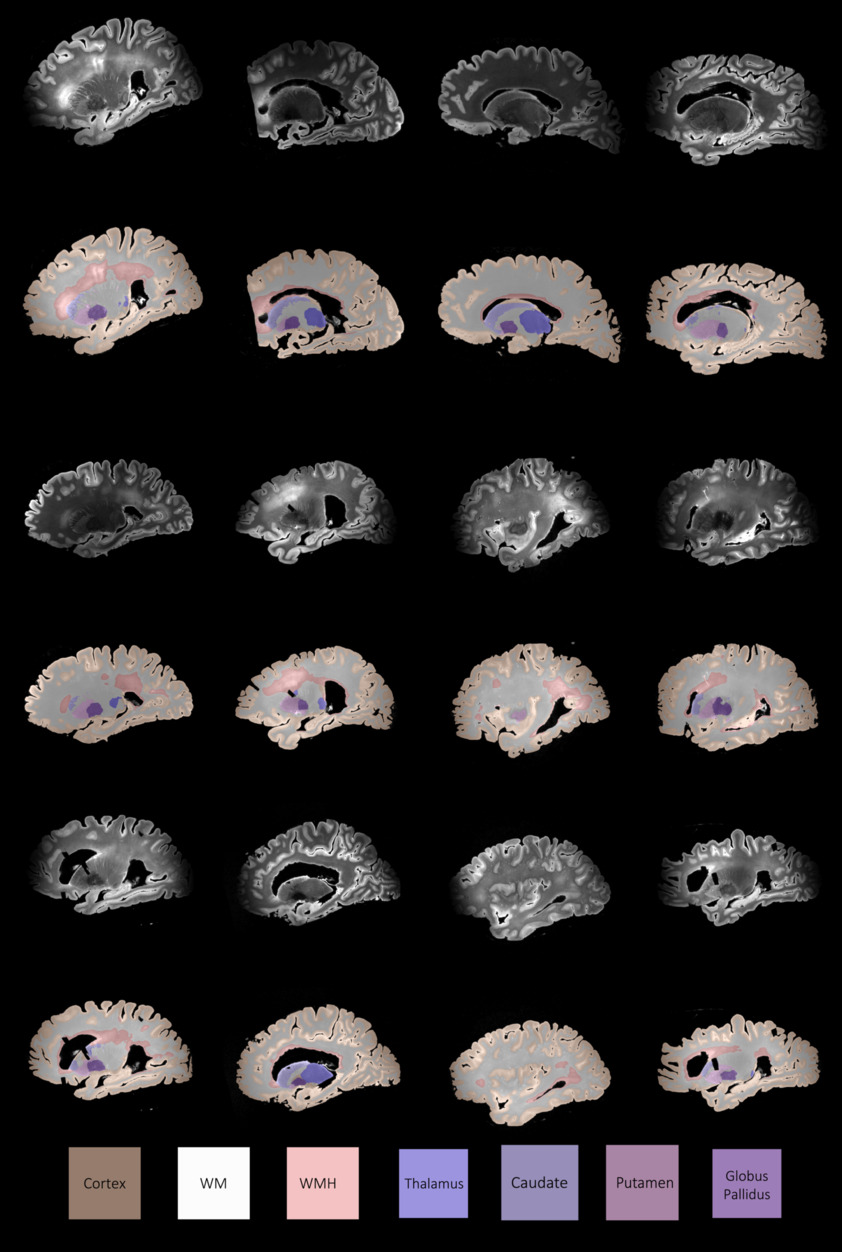} %March_2023_Figure9_part3_final_revised
        \caption{\textit{\textbf{Continued}}. Automated segmentations by nnU-Net of cortical gray matter, WMH, WM and the four subcortical structures for a sample of 36 subjects in the cohort shown in sagittal view.}
        \label{figure_2D_segmentations_automated}
\end{figure}

\begin{figure}[H]
\centering
\includegraphics[width=\textwidth,height=\textheight,keepaspectratio]{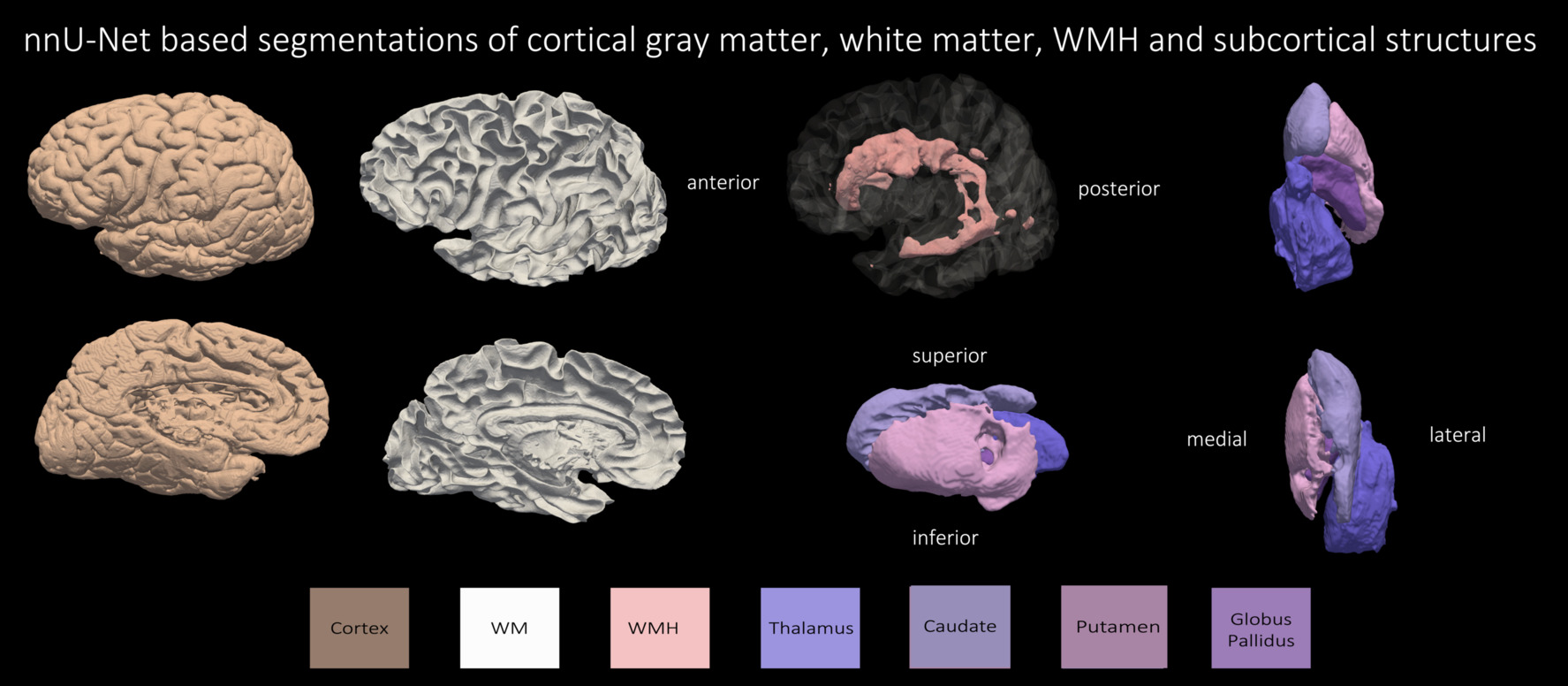} %dec_automated_3d_segm_resized.jpg
        \caption{Three-dimensional renderings of automated segmentations by nnU-Net of cortical gray matter, WM, WMH and the four subcortical structures for the subject with primary age-related tauopathy and cerebrovascular disease.} %(75 years old, black female)
        \label{figure_3D_segmentations_automated}
\end{figure}

\subsection{Generalization to other imaging sequences and protocols.}
\label{results:generalization}
\textcolor{blue}{Further highlighting the strong generalization properties of nnU-Net, Figure \ref{figure_generalization_all} illustrates that the nnU-Net model trained on 7~T 0.3 x 0.3 x 0.3 mm${^3}$ T2w images is able to generalize well to MRI sequences and resolutions unseen during training. In particular, we evaluate the trained model on 7~T T2*w GRE FLASH sequence postmortem images acquired \cite{tisdall2021joint} at 0.28 x 0.28 x 0.28 mm$^{3}$ (N=13) and 0.16 x 0.16 x 0.16 mm$^{3}$ (N=73) resolution as shown qualitatively in Figure \ref{figure_generalization_all}. We observe good generalization performance for GM, WM and WMH but observe some under-segmentations in the subcortical structures for both the 160 and 280 micron sequences. Currently, we only provide qualitative assessment on a representative sample of the FLASH images. The limitation of lies in quantitative assessment (beyond the scope of the current study) in terms of Dice score and ICV which is currently not possible due to lack of reference manual segmentation.}

\begin{figure}[]
\centering
\includegraphics[width=\textwidth,height=\textheight,keepaspectratio]{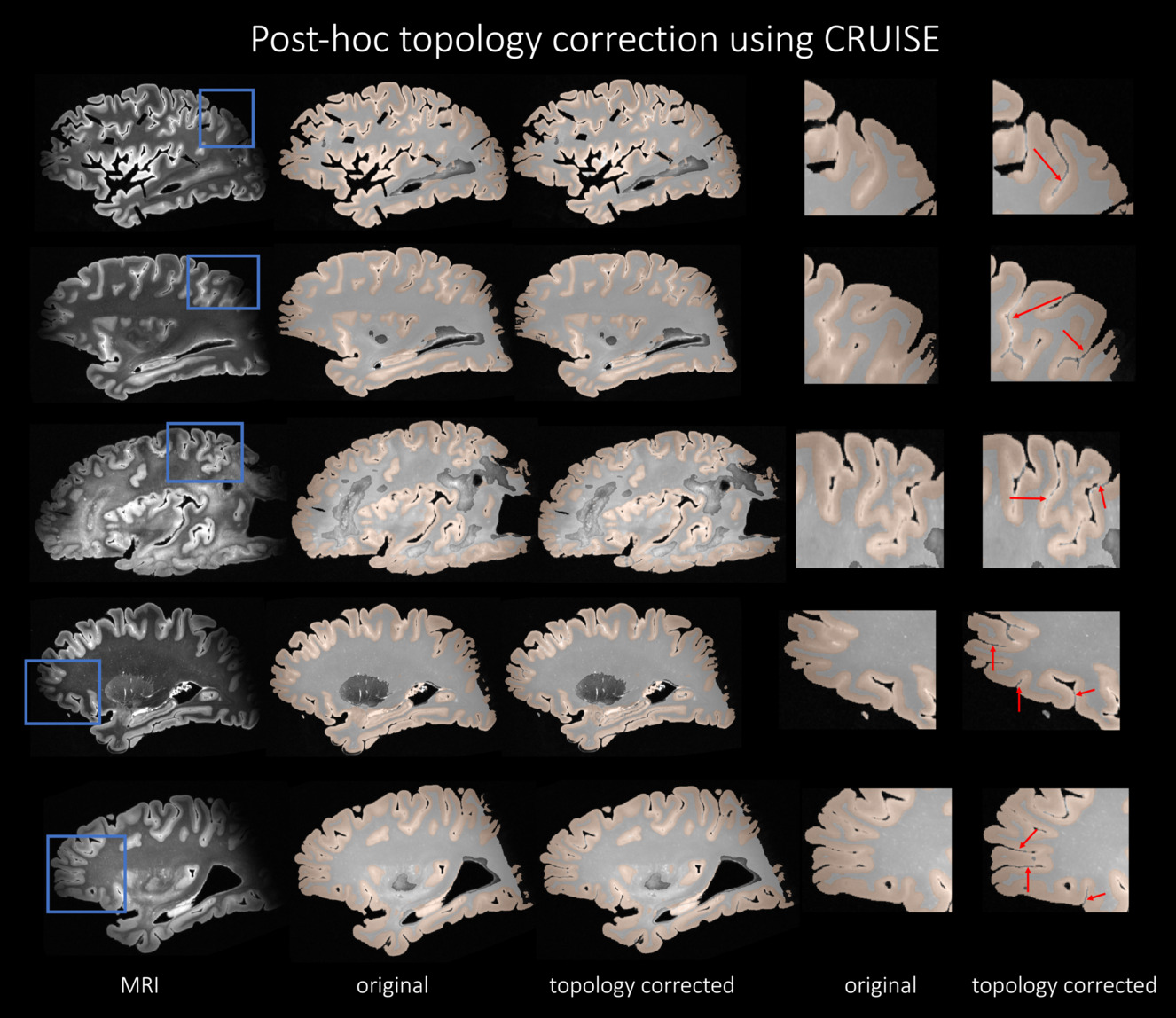}
        \caption{\textcolor{blue}{nnU-Net-CRUISE. Topologically corrected automated nnU-Net segmentations of cortical gray matter and white matter shown in sagittal view for five subjects. The first columns shows the MRI slice, the automated nnU-Net segmentation before and after topology correction step (nnU-Net-CRUISE) are shown in columns 2 and 3 respectively. Columns 4 and 5 show the zoomed-in area with the red arrows indicating the regions where topology correction improves the segmentation.}}
        \label{figure_topo}
\end{figure}

\begin{figure}[]
\centering
\includegraphics[width=\textwidth,height=\textheight,keepaspectratio]{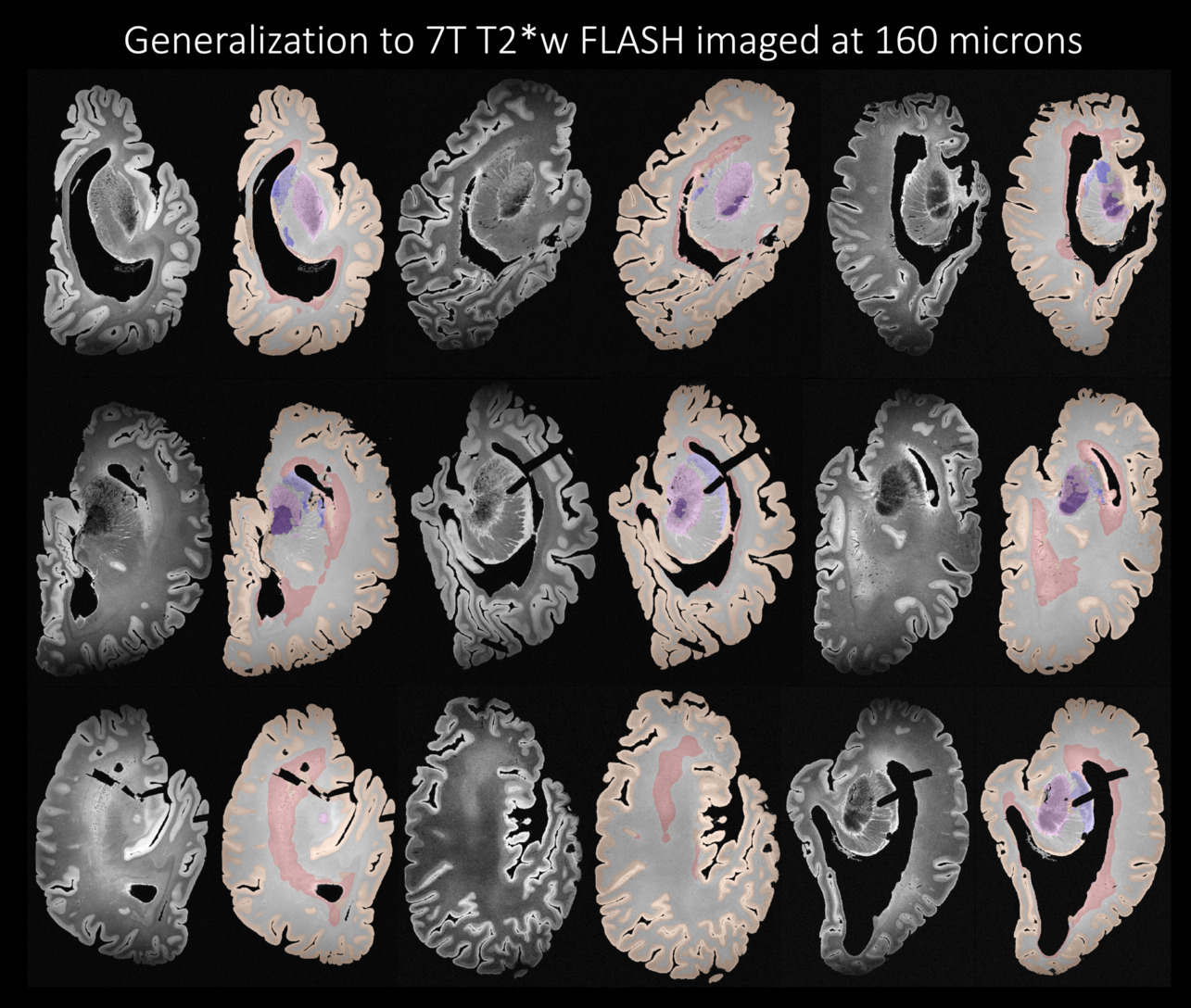}
        \caption{Generalization to other unseen imaging sequences. The nnU-Net architecture trained on 7~T T2w images at 0.3 x 0.3 x 0.3 mm$^3$ generalized well to images acquired at 7~T T2*w at a high resolution of 0.16 x 0.16 x 0.16  mm$^3$.}
        \label{figure_generalization_all}
        \addtocounter{figure}{-1} 
\end{figure}

\begin{figure}[]
\centering
\includegraphics[width=\textwidth,height=\textheight,keepaspectratio]{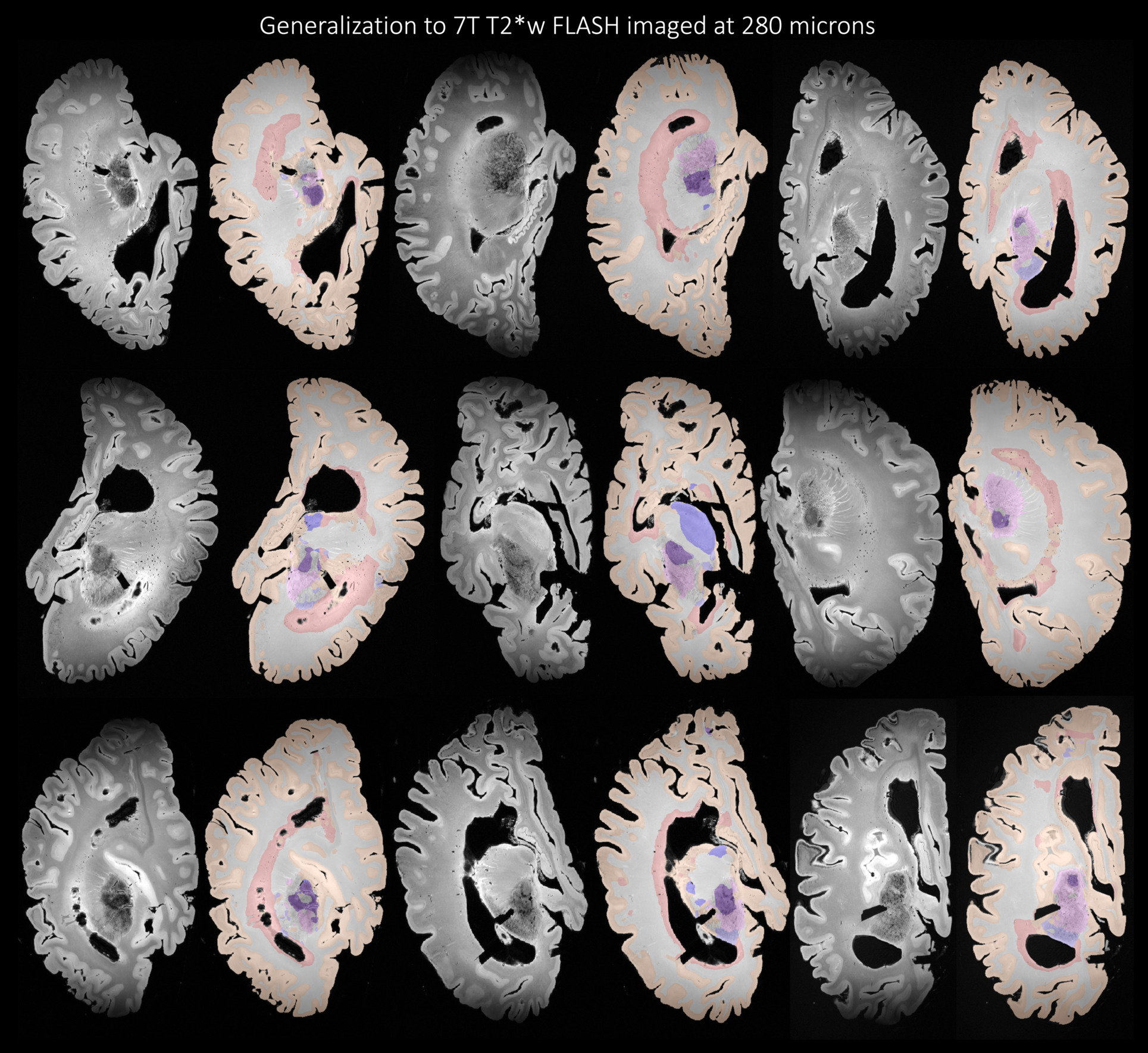}
        \caption{(\textbf{\textit{Continued.}}) The nnU-Net architecture trained on 7~T T2w images at 0.3 x 0.3 x 0.3 mm$^3$ generalized well to images acquired at 7~T T2*w at a high resolution of 0.28 x 0.28 x 0.28 mm$^3$.}
        \label{figure_generalization_all}
\end{figure}

%%%%%%%% NEW
\begin{adjustbox}{angle=90, caption={Morphometry associations with underlying neuropathology measurements. Shown is the one-sided Spearman's correlation, \textcolor{blue}{controlling for age, sex and PMI}, between regional cortical thickness measures derived from \textcolor{blue}{topologically corrected} nnU-Net-CRUISE gray matter segmentation with corresponding regional ratings of p-tau pathology, neuronal loss density, and global ratings of amyloid-$\beta$, CERAD, and Braak staging. Each cell is color coded with darker shades indicating more negative correlations. The asterisk indicates that the test survived Bonferroni multiple testing correction \cite{bonferroni1935calcolo}. CI indicates 95\% confidence interval. \underline{Legend} *:0.01 $<$ \emph{p} $\leq$ 0.05; **:0.001 $<$ \emph{p} $\leq$ 0.01; ***:0.0001 $<$ \emph{p} $\leq$ 0.001;  ****:0.00001 $<$ \emph{p} $\leq$ 0.0001.},float=table}
\label{table:path_all}
\scalebox{0.50}
{
\centering
\begin{tabular}{|c|cc|cc|cc|cc|cc|}
\hline
\textbf{Pathology ratings} &
  \multicolumn{2}{c|}{\cellcolor[HTML]{FFCCC9}\textbf{Abeta}} &
  \multicolumn{2}{c|}{\cellcolor[HTML]{FFCE93}\textbf{Braak stage}} &
  \multicolumn{2}{c|}{\cellcolor[HTML]{FFFC9E}\textbf{CERAD}} &
  \multicolumn{2}{c|}{\cellcolor[HTML]{FD6864}\textbf{p-tau}} &
  \multicolumn{2}{c|}{\cellcolor[HTML]{F8A102}\textbf{Neuronal Loss}} \\ \hline
\textbf{ROI} &
  \multicolumn{1}{c|}{\textbf{rho}} &
  \textbf{CI} &
  \multicolumn{1}{c|}{\textbf{rho}} &
  \textbf{CI} &
  \multicolumn{1}{c|}{\textbf{rho}} &
  \textbf{CI} &
  \multicolumn{1}{c|}{\textbf{rho}} &
  \textbf{CI} &
  \multicolumn{1}{c|}{\textbf{rho}} &
  \textbf{CI} \\ \hline
\textbf{Visual} &
  \multicolumn{1}{c|}{\gradientNeuro{-0.176}} &
  {[}-1.0, 0.01{]} &
  \multicolumn{1}{c|}{\gradientNeuro{-0.151}} &
  {[}-1.0, 0.04{]} &
  \multicolumn{1}{c|}{\gradientNeuro{-0.17}} &
  {[}-1.0, 0.02{]} &
  \multicolumn{1}{c|}{\gradientNeuro{-0.003}} &
  {[}-1.0, 0.19{]} &
  \multicolumn{1}{c|}{\gradientNeuro{0.1}} &
  {[}-1.0, 0.28{]} \\ \hline
\textbf{Motor} &
  \multicolumn{1}{c|}{\gradientNeuro{-0.139}} &
  {[}-1.0, 0.05{]} &
  \multicolumn{1}{c|}{\gradientNeuro{-0.135}} &
  {[}-1.0, 0.06{]} &
  \multicolumn{1}{c|}{\gradientNeuro{-0.063}} &
  {[}-1.0, 0.13{]} &
  \multicolumn{1}{c|}{NA} &
  NA &
  \multicolumn{1}{c|}{NA} &
  NA \\ \hline
\textbf{Posterior cingulate} &
  \multicolumn{1}{c|}{\gradientNeuro{-0.46} ***} &
  {[}-1.0, -0.27{]} &
  \multicolumn{1}{c|}{\gradientNeuro{-0.445} ***} &
  {[}-1.0, -0.25{]} &
  \multicolumn{1}{c|}{\gradientNeuro{-0.419} ***} &
  {[}-1.0, -0.23{]} &
  \multicolumn{1}{c|}{\gradientNeuro{-0.357} **} &
  {[}-1.0, -0.16{]} &
  \multicolumn{1}{c|}{\gradientNeuro{-0.303}} &
  {[}-1.0, -0.1{]} \\ \hline
\textbf{Midfrontal} &
  \multicolumn{1}{c|}{\gradientNeuro{-0.441} ****} &
  {[}-1.0, -0.28{]} &
  \multicolumn{1}{c|}{\gradientNeuro{-0.327} **} &
  {[}-1.0, -0.15{]} &
  \multicolumn{1}{c|}{\gradientNeuro{-0.271}} &
  {[}-1.0, -0.09{]} &
  \multicolumn{1}{c|}{\gradientNeuro{-0.383} ***} &
  {[}-1.0, -0.21{]} &
  \multicolumn{1}{c|}{\gradientNeuro{-0.256}} &
  {[}-1.0, -0.07{]} \\ \hline
\textbf{Anterior cingulate} &
  \multicolumn{1}{c|}{\gradientNeuro{-0.113}} &
  {[}-1.0, 0.09{]} &
  \multicolumn{1}{c|}{\gradientNeuro{-0.16}} &
  {[}-1.0, 0.05{]} &
  \multicolumn{1}{c|}{\gradientNeuro{-0.168}} &
  {[}-1.0, 0.04{]} &
  \multicolumn{1}{c|}{\gradientNeuro{-0.185}} &
  {[}-1.0, 0.02{]} &
  \multicolumn{1}{c|}{\gradientNeuro{-0.308}} &
  {[}-1.0, -0.11{]} \\ \hline
\textbf{Orbitofrontal} &
  \multicolumn{1}{c|}{\gradientNeuro{0.105}} &
  {[}-1.0, 0.29{]} &
  \multicolumn{1}{c|}{\gradientNeuro{0.219}} &
  {[}-1.0, 0.39{]} &
  \multicolumn{1}{c|}{\gradientNeuro{0.122}} &
  {[}-1.0, 0.3{]} &
  \multicolumn{1}{c|}{NA} &
  NA &
  \multicolumn{1}{c|}{NA} &
  NA \\ \hline
\textbf{Superior temporal} &
  \multicolumn{1}{c|}{\gradientNeuro{-0.325} **} &
  {[}-1.0, -0.14{]} &
  \multicolumn{1}{c|}{\gradientNeuro{-0.289}} &
  {[}-1.0, -0.1{]} &
  \multicolumn{1}{c|}{\gradientNeuro{-0.271}} &
  {[}-1.0, -0.09{]} &
  \multicolumn{1}{c|}{\gradientNeuro{-0.293}} &
  {[}-1.0, -0.11{]} &
  \multicolumn{1}{c|}{\gradientNeuro{-0.249}} &
  {[}-1.0, -0.06{]} \\ \hline
\textbf{Inferior frontal} &
  \multicolumn{1}{c|}{\gradientNeuro{-0.178}} &
  {[}-1.0, 0.01{]} &
  \multicolumn{1}{c|}{\gradientNeuro{-0.094}} &
  {[}-1.0, 0.1{]} &
  \multicolumn{1}{c|}{\gradientNeuro{-0.145}} &
  {[}-1.0, 0.04{]} &
  \multicolumn{1}{c|}{\gradientNeuro{-0.171}} &
  {[}-1.0, 0.02{]} &
  \multicolumn{1}{c|}{\gradientNeuro{-0.145}} &
  {[}-1.0, 0.04{]} \\ \hline
\textbf{Anterior insula} &
  \multicolumn{1}{c|}{\gradientNeuro{-0.202}} &
  {[}-1.0, -0.02{]} &
  \multicolumn{1}{c|}{\gradientNeuro{-0.17}} &
  {[}-1.0, 0.02{]} &
  \multicolumn{1}{c|}{\gradientNeuro{-0.222}} &
  {[}-1.0, -0.04{]} &
  \multicolumn{1}{c|}{\gradientNeuro{-0.201}} &
  {[}-1.0, -0.01{]} &
  \multicolumn{1}{c|}{\gradientNeuro{-0.215}} &
  {[}-1.0, -0.03{]} \\ \hline
\textbf{Anterior temporal} &
  \multicolumn{1}{c|}{\gradientNeuro{-0.229}} &
  {[}-1.0, -0.04{]} &
  \multicolumn{1}{c|}{\gradientNeuro{-0.266}} &
  {[}-1.0, -0.08{]} &
  \multicolumn{1}{c|}{\gradientNeuro{-0.136}} &
  {[}-1.0, 0.05{]} &
  \multicolumn{1}{c|}{\gradientNeuro{-0.238}} &
  {[}-1.0, -0.05{]} &
  \multicolumn{1}{c|}{\gradientNeuro{-0.194}} &
  {[}-1.0, -0.01{]} \\ \hline
\textbf{Ventrolaterl temporal cortex} &
  \multicolumn{1}{c|}{\gradientNeuro{-0.1}} &
  {[}-1.0, 0.13{]} &
  \multicolumn{1}{c|}{\gradientNeuro{-0.175}} &
  {[}-1.0, 0.06{]} &
  \multicolumn{1}{c|}{\gradientNeuro{-0.053}} &
  {[}-1.0, 0.18{]} &
  \multicolumn{1}{c|}{\gradientNeuro{-0.114}} &
  {[}-1.0, 0.12{]} &
  \multicolumn{1}{c|}{\gradientNeuro{-0.05}} &
  {[}-1.0, 0.18{]} \\ \hline
\textbf{Superior pareital} &
  \multicolumn{1}{c|}{\gradientNeuro{-0.298}} &
  {[}-1.0, -0.11{]} &
  \multicolumn{1}{c|}{\gradientNeuro{-0.225}} &
  {[}-1.0, -0.03{]} &
  \multicolumn{1}{c|}{\gradientNeuro{-0.25}} &
  {[}-1.0, -0.06{]} &
  \multicolumn{1}{c|}{\gradientNeuro{-0.295}} &
  {[}-1.0, -0.1{]} &
  \multicolumn{1}{c|}{\gradientNeuro{-0.1}} &
  {[}-1.0, 0.1{]} \\ \hline
\textbf{Angular gyrus} &
  \multicolumn{1}{c|}{\gradientNeuro{-0.166}} &
  {[}-1.0, 0.02{]} &
  \multicolumn{1}{c|}{\gradientNeuro{-0.225}} &
  {[}-1.0, -0.04{]} &
  \multicolumn{1}{c|}{\gradientNeuro{-0.109}} &
  {[}-1.0, 0.08{]} &
  \multicolumn{1}{c|}{\gradientNeuro{-0.244}} &
  {[}-1.0, -0.06{]} &
  \multicolumn{1}{c|}{\gradientNeuro{-0.139}} &
  {[}-1.0, 0.05{]} \\ \hline
\textbf{Entorhinal cortex} &
  \multicolumn{1}{c|}{\gradientNeuro{-0.429} ***} &
  {[}-1.0, -0.26{]} &
  \multicolumn{1}{c|}{\gradientNeuro{-0.522} ****} &
  {[}-1.0, -0.37{]} &
  \multicolumn{1}{c|}{\gradientNeuro{-0.335} **} &
  {[}-1.0, -0.15{]} &
  \multicolumn{1}{c|}{\gradientNeuro{-0.348} **} &
  {[}-1.0, -0.17{]} &
  \multicolumn{1}{c|}{\gradientNeuro{-0.414} ***} &
  {[}-1.0, -0.24{]} \\ \hline
\textbf{Brodmann Area 35} &
  \multicolumn{1}{c|}{\gradientNeuro{-0.158}} &
  {[}-1.0, 0.04{]} &
  \multicolumn{1}{c|}{\gradientNeuro{-0.189}} &
  {[}-1.0, 0.0{]} &
  \multicolumn{1}{c|}{\gradientNeuro{-0.265}} &
  {[}-1.0, -0.08{]} &
  \multicolumn{1}{c|}{\gradientNeuro{-0.179}} &
  {[}-1.0, 0.01{]} &
  \multicolumn{1}{c|}{\gradientNeuro{-0.285}} &
  {[}-1.0, -0.1{]} \\ \hline
\textbf{Parahippocampal cortex} &
  \multicolumn{1}{c|}{\gradientNeuro{-0.307}} &
  {[}-1.0, -0.13{]} &
  \multicolumn{1}{c|}{\gradientNeuro{-0.328} **} &
  {[}-1.0, -0.15{]} &
  \multicolumn{1}{c|}{\gradientNeuro{-0.25}} &
  {[}-1.0, -0.06{]} &
  \multicolumn{1}{c|}{\gradientNeuro{-0.133}} &
  {[}-1.0, 0.06{]} &
  \multicolumn{1}{c|}{\gradientNeuro{-0.188}} &
  {[}-1.0, 0.0{]} \\ \hline
\end{tabular}
}
\end{adjustbox}

\subsection{Morphometry associations with underlying Neuropathology}
\label{results:morphometry_neuropath}

\subsubsection{Regional cortical atrophy correlation patterns with underlying ratings}
\label{results:cort_neuropath}
We correlate regional cortical thickness measures derived from nnU-Net-CRUISE gray matter segmentation with corresponding regional ratings of p-tau pathology and neuronal loss density, amyloid-$\beta$ ratings, CERAD and Braak staging as explained in Section \ref{section:neuropath_morph}. Table \ref{table:path_all} tabulates the correlation between the regional cortical thickness and the pathology ratings for measurements based-on automated nnU-Net-CRUISE segmentations for the subjects within the \textcolor{blue}{AD continuum (N=82)}. For the automated nnU-Net segmentation-based regional cortical thickness measurements, \textcolor{blue}{we observe significant negative one-sided Spearman's correlation which survived Bonferroni multiple corrections (corrected p-value <0.05), controlling for age, sex and PMI:}

\begin{itemize}
    \item \textcolor{blue}{in posterior cingulate (\textit{r}=-0.46, \textit{p}=0.0001), mid-frontal (\textit{r}=-0.441, \textit{p}=10$^{-5}$), superior temporal (\textit{r}=-0.325, \textit{p}=0.002) and entorhinal cortex (\textit{r}=-0.429, \textit{p}=0.0001) with amyloid-$\beta$;}
    
    \item \textcolor{blue}{in posterior cingulate (\textit{r}=-0.445, \textit{p}=0.0002), midfrontal (\textit{r}=-0.327, \textit{p}=0.0018) and entorhinal cortex (\textit{r}=-0.522, \textit{p}=10$^{-5}$) and parahippocampal cortex (\textit{r}=-0.328, \textit{p}=0.0018) with Braak staging;}
    
    \item \textcolor{blue}{in posterior cingulate (\textit{r}=-0.419, \textit{p}=0.0004) and entorhinal cortex (\textit{r}=-0.335, \textit{p}=0.0016) with CERAD rating;}
    
    \item \textcolor{blue}{in posterior cingulate (\textit{r}=-0.357, \textit{p}=0.0024), midfrontal (\textit{r}=-0.383, \textit{p}=0.0003), entorhinal cortex (\textit{r}=-0.348, \textit{p}=0.0011) with regional p-tau rating;}
    
    \item \textcolor{blue}{in entorhinal cortex (\textit{r}=-0.414, \textit{p}=0.0001) with regional neuronal loss rating.}
\end{itemize}

\textcolor{blue}{Supplementary Table 3 shows the one-sided Spearman’s correlation (controlling for age, sex and PMI) between the cortical thickness measures derived from the topologically corrected nnU-Net-CRUISE gray matter segmentation with the corresponding medial temporal lobe (MTL) ratings of p-tau pathology and neuronal loss density as the MTL is a region linked to early neurodegeneration in Alzheimer’s disease. The MTL pathology ratings are computed by taking the average pathology of the entorhinal cortex, CA1 and subiculum, and the dentate gyrus. We notice significant results in entorhinal cortex (\textit{r}=-0.53, \textit{p}=10${^-5}$) with the MTL p-tau rating; and in posterior cingulate (\textit{r}=-0.396, \textit{p}=0.0008), entorhinal cortex (\textit{r}=-0.47, \textit{p}=10${^-5}$), and BA35 (\textit{r}=-0.321, \textit{p}=0.0026) with the MTL neuronal loss density.}

\textcolor{blue}{Separately, Supplementary Table 4 tabulates the morphometry associations with the underlying Neuropathology for the AD continuum (N=21) subjects from the subset of the cohort which have the manual segmentations (N=36) as described in Section \ref{sec:methods_segmentation_cortical}. We observe that the analysis based on manual segmentations show a similar trend as the nnU-Net-CRUISE automated segmentations, suggesting that automated segmentations obtained from the developed pipeline provide meaningful associations with the underlying neuropathological measurements, are thus reliable for validation of clinical ratings, and can act as surrogate for the time consuming user-supervised segmentations. Finally, Supplementary Figures 4-10 depict all the correlations as a plot along with the p-values to appreciate the data distribution.}

\subsubsection{Normalized WMH volume correlation patterns with regional cortical thickness and subcortical volumes}
\label{results:WMH_cortical_sustain_subcort}
Table \ref{tab:path_allWMH_subcort} compares the correlation between the regional cortical thickness and subcortical volumes with the normalized WMH (WMH volume divided by the total WM volume) volume for measurements based-on automated nnU-Net-CRUISE for \textcolor{blue}{the subjects in the \textcolor{blue}{AD continuum (N=82)}. We observed significant negative one-sided Spearman's correlation which survived Bonferroni multiple corrections in posterior cingulate (\textit{r}=-0.448, \textit{p}=0.0001) and midfrontal cortex (\textit{r}=-0.325, \textit{p}=0.0019) with normalized WMH volume for the automated nnU-Net segmentation-based regional thickness measurements; and in caudate (\textit{r}=-0.399, \textit{p}=0.0001) and thalamus (\textit{r}=-0.352, \textit{p}=0.0007) for the subcortical structures when correlated with the normalized WMH volume. All the tests were controlled for age, sex and PMI. Supplementary Figures 11-12 depict these correlations as a plot alongwith the p-values to observe the data distribution.}

%%%%%%%%%%%%%%% NEW
\begin{table}[]
\centering
\caption{White matter hyperintensities volume correlations. Shown is the one-sided Spearman's correlation between normalized WMH volume (obtained by dividing the WMH volume with the corresponding WM) with regional cortical thickness based-on nnU-Net-CRUISE. For the four subcortical structures, partial one-sided Spearman's correlation, with uncorrected p-value in bracket is shown. \textcolor{blue}{All the tests were controlled for age, sex and PMI.} The asterisk indicates that the test survived Bonferroni multiple tests correction. Each cell is color coded with darker shades indicating more negative correlations. CI indicates 95\% confidence interval. \underline{Legend} *:0.01 $<$ \emph{p} $\leq$ 0.05; **:0.001 $<$ \emph{p} $\leq$ 0.01; ***:0.0001 $<$ \emph{p} $\leq$ 0.001;  ****:0.00001 $<$ \emph{p} $\leq$ 0.0001.}
\label{tab:path_allWMH_subcort}
\begin{tabular}{|c|c|c|}
\hline
\textbf{ROI}             & \textbf{rho} & \textbf{CI}       \\ \hline
\textbf{Visual}             & \gradientWMH{0.243}        & {[}-1.0, 0.41{]}  \\ \hline
\textbf{Motor cortex}             & \gradientWMH{-0.256}       & {[}-1.0, -0.07{]} \\ \hline
\textbf{Posterior cingulate}            & \gradientWMH{-0.448} ***       & {[}-1.0, -0.26{]} \\ \hline
\textbf{Midfrontal}              & \gradientWMH{-0.325} **      & {[}-1.0, -0.15{]} \\ \hline
\textbf{Anterior cingulate}            & \gradientWMH{-0.262}       & {[}-1.0, -0.06{]} \\ \hline
\textbf{Orbitofrontal}             & \gradientWMH{0.355}        & {[}-1.0, 0.51{]}  \\ \hline
\textbf{Superior temporal pole}           & \gradientWMH{-0.205}       & {[}-1.0, -0.02{]} \\ \hline
\textbf{Inferior frontal}              & \gradientWMH{-0.182}      & {[}-1.0, 0.01{]}  \\ \hline
\textbf{Anterior insula}           & \gradientWMH{0.011}        & {[}-1.0, 0.2{]}   \\ \hline
\textbf{Anterior temporal pole}          & \gradientWMH{-0.050}        & {[}-1.0, 0.14{]}  \\ \hline
\textbf{Ventrolateral temporal cortex}             & \gradientWMH{-0.142}       & {[}-1.0, 0.09{]}  \\ \hline
\textbf{Superior parietal}              & \gradientWMH{-0.193}       & {[}-1.0, 0.0{]}   \\ \hline
\textbf{Angular gyrus}             & \gradientWMH{-0.243}      & {[}-1.0, -0.06{]} \\ \hline
\textbf{Entorhinal cortex}             & \gradientWMH{-0.293}       & {[}-1.0, -0.11{]} \\ \hline
\textbf{Brodmann Area 35}            & \gradientWMH{-0.156}        & {[}-1.0, 0.04{]}  \\ \hline
\textbf{Parahippocampal cortex}             & \gradientWMH{-0.148}        & {[}-1.0, 0.04{]}  \\ \hline
\textbf{Caudate}         & \gradientWMH{-0.399} ***     & {[}-1.0, -0.23{]} \\ \hline
\textbf{Globus Pallidus} & \gradientWMH{-0.193}      & {[}-1.0, -0.01{]} \\ \hline
\textbf{Putamen}         & \gradientWMH{-0.312}      & {[}-1.0, -0.13{]} \\ \hline
\textbf{Thalamus}        & \gradientWMH{-0.352} ***       & {[}-1.0, -0.18{]} \\ \hline
\end{tabular}
\end{table}

\section{Discussion}
\label{section:discussion}
\subsection*{\textbf{Segmentation pipeline}}
To our knowledge, the current study is the most comprehensive assessment of automated segmentation of 7 T postmortem human brain MRI. Our evaluation is performed in a large cohort of \textcolor{blue}{135} brain specimens with a range of neurodegenerative pathologies, and focuses on multiple tasks: cortical gray matter segmentation, subcortical gray matter structure segmentation, as well as white matter and WMH segmentation. For cortical segmentation, we evaluated nine deep learning architectures using direct metrics of segmentation accuracy (cross-validation DSC), derived morphological metrics (ICC of regional cortical thickness with the reference standard; comparison of associations with pathology), and visual assessment.

Our paper stands apart from recent work on automated segmentation of postmortem brain MRI, which has either been performed in lower-resolution 3T MRI scans \cite{mancini2020multimodal}, or in smaller high-resolution datasets \cite{jonkman2019normal}. \cite{mancini2020multimodal} used FreeSurfer \cite{fischl2012freesurfer} and a Bayesian modelling technique, SAMSEG \cite{puonti2016fast}, to map a single post-mortem specimen imaged at 3~T but did not evaluate on higher resolution 7~T. In addition to segmenting the cortical gray matter, \cite{mancini2020multimodal} parcellate the cortex into anatomical regions. A separate study \cite{kotrotsou2014ex} also segmented the entire hemisphere but applied to a smaller dataset of 7 subjects with a slice thickness of 1.5 mm and imaged at 3 T. Both these methods relied on multi-atlas based image segmentation which is dependent on registration between high resolution postmortem and low resolution antemortem atlases. Registration between inter-modality postmortem and antemortem currently remains challenging, especially for higher-resolution 7T postmortem MRI [\cite{casamitjana2021synth}, \cite{casamitjana2022robust} and \cite{daly2021convolutional}]. Other recent studies on postmortem human brain morphometry 
 [\cite{yushkevich2021three, wisse2021downstream, ravikumar2021ex, adler2014histology, adler2018characterizing, iglesias2015computational, augustinack2014mri, dekraker2018unfolding, dekraker2021surface}] focused on specific areas such as the hippocampus or the MTL, and relied on manual segmentation to guide inter-subject registration and atlas generation.

Our evaluation demonstrates that deep learning-based segmentation pipelines, particularly nnU-Net, can generate high-quality segmentations of cortical gray matter, subcortical structures, normal-appearing white matter, and white matter lesions even with very limited training data. With inference time of around 15 minutes on a CPU, our pipeline represents the first step towards fast, automated and reliable brain mapping of high resolution postmortem whole-hemisphere MRI. Our nnU-Net based pipeline generalized well to areas of low contrast unseen during training, as well as to other MRI protocols, and resolutions. \textcolor{blue}{Furthermore, our pipeline impose post-hoc constraints on the automated segmentations to produce geometrically accurate and topologically correct segmentations using  deformable surface-based methods, which have shown tremendous success in the antemortem literature. Therefore, we were able to correctly label the challenging boundary deep sulci and the cortical gray matter leading to more reliable cortical thickness estimates.} Moreover, thickness measures derived from the deep learning-based automated segmentations concur with the reference standard, and similar associations between thickness and pathology are detected using automatically-derived and reference standard thickness measurements, albeit with the latter usually having higher effect sizes. This suggests that fully automated cortical thickness analysis is feasible for postmortem MRI. Indeed, with further improvements to accuracy (e.g., a larger training set covering more of the hemisphere), automated postmortem segmentation may make the labor-intensive and subjective semi-automated approach to cortical thickness measurement unnecessary. Thus, the study suggests the feasibility of a fully automated group-wise cortical thickness analysis in postmortem MRI analogous to the way FreeSurfer is used for antemortem MRI morphometry, which forms the basis of our future research direction.

The limitations of the current pipeline include a relatively small whole hemisphere cortical gray matter segmentation training set which limited our ability to use direct metrics such as DSC to evaluate overall segmentation accuracy. Indeed, the method that performed best in terms of cross-validation DSC (AnatomyNet) performed worse in areas unseen during training than nnU-Net. To address this limitation, in future work we plan to train the method on manual segmentations of whole hemispheres; however, generating such a dataset will require a significant additional investment in time and effort. We still rely on manually placed landmarks to measure thickness in specific anatomical regions, and we do not show 3D maps of thickness as is common in antemortem morphometry studies. \textcolor{blue}{For example, the variability in the placement of the BA35 landmarks may explain our observation of the stronger thickness/pathology associations in the ERC than in BA35, even though BA35 is thought to have early tau pathology.}

Due to limited availability of reference standard segmentations, parcellating the brain into subregions, as in done in most antemortem MRI, currently remains a challenge. We intend to address this limitation by semi-manually annotating the brain into different cortical and subcortical structures, guided by anatomical priors derived from antemortem MRI studies. We could then develop techniques for groupwise normalization studies by building a template for postmortem MRI. Towards this goal, we are currently developing deep learning-based methods for registration between postmortem and antemortem MRI. Our work is limited to ADRD, but in future we plan to evaluate our methodological pipeline on a cohort of non-demented specimens obtained from a separate post-mortem postmortem dataset [\cite{jonkman2019normal, frigerio2021amyloid, boon2019can}].

\textcolor{blue}{Certain limitations of the postmortem imaging should be noted. \cite{wisse2017comparison} compared the cortical thickness of the MTL substructures between antemortem and postmortem; and also postmortem (3T) and postmortem (9.4T) MRI. They found differences in thickness on different MRI scans and attributed to the various factors such as: (1) difference in the actual size between the antemortem and postmortem tissue is due to an actual difference in size as studies have suggested that tissue changes may occur during or after death since the agonal state causes hypoxia and ischemia which results in brain swelling. (2) an increase in size could result from brain extraction, for example by a relief of intracranial pressure after autopsy. (3) formalin fixation could cause the underlying differences by shrinking the brain after several weeks.}

\textcolor{blue}{Currently, we do not have a way to quantitatively assess the accuracy of the T2*w FLASH images. In follow-up work, we plan to collect reference data by manually correcting the mis-segmentations and then re-train the deep learning model in a few-shot learning setting to achieve a boost in segmentation performance for T2*w FLASH MRI. Then, we will quantitatively assess the segmentation performance for the said generalization procedure using reference user-supervised manually corrected reference segmentation.}

\subsection*{\textbf{Neuropathology associations}}
With the help of the proposed segmentation and morphometry pipeline, we were able to conduct studies of postmortem MRI that have not been possible before: for example, we replicated some of the findings from \cite{sadaghiani2022associations}, and drew associations between WMH volume, cortical thickness and subcortical volumes. Here, we discuss some of the interesting findings. Strong postmortem image analysis frameworks allows us to better understand the distinct roles and degrees by which antemortem pathology affects neurodegeneration. Prior work shows how amyloid-$\beta$, p-tau, TDP-43, etc. have differential influences on atrophy \cite{dugger2017pathology, robinson2018neurodegenerative, matej2019alzheimer, negash2011cognition}, and an automated pipeline and dataset as shown here can bolster the quantitative interrogation of these open questions. It will allow future work to study links between macro structure and other local processes beyond pathology, including inflammatory markers, gene expression, etc. We also note that measuring WMH in postmortem imaging adds value to histological studies, as we have less clear measures of vascular burden with traditional autopsy, which will allow probing of WMH for better understanding their pathologic correlates given their non-specific nature.

The current study demonstrates the associations between regional cortical thickness measurements with the underlying semi-quantitative neuropathological ratings for the AD cohort. Negative correlations between p-tau and cortical thickness were found to be significant in angular gyrus and midfrontal regions, which is in line with previous research previous in antemortem [\cite{lapoint2017association, harrison2021distinct, xia2017association, das2019vivo, whitwell2018imaging}] and postmortem \textit{in situ} MRI \cite{frigerio2021amyloid} studies. Tau pathology is concurrent with neuronal loss in ADRD \cite{dawe2011neuropathologic, ohm2021accumulation, jack2018nia} and the loss of neurons is likely a key source of cortical atrophy. We observed that cortical thickness showed significant negative correlation with neuronal loss in BA35 and entorhinal cortex, regions where p-tau pathology have predicted the atrophy rate \cite{lapoint2017association, xie2018early, la2020prospective} in antemortem studies. Significant negative correlations were observed between Braak staging and cortical thickness in midfrontal, ERC and BA35, regions consistent with high p-tau uptake in positron emission tomography (PET) imaging with cortical thickness on MRI. Tau pathology in Braak regions play an important role in cortical atrophy and cognitive decline during the course of AD. Similar findings are reported for global cortical thickness with \textit{in situ} post-mortem MRI in \cite{frigerio2021amyloid}. The relationship between amyloid-$\beta$ and neurodegeneration is thought to be rather indirect \cite{jack2018nia, gomez2022lesions}. Nevertheless, we did find strong negative correlations in between thickness and amyloid-$\beta$ in midfrontal, inferior frontal, and ventrolateral temporal cortex, brain regions implicated in working memory capacity \cite{barbey2013dorsolateral, chiou2018anterior}. Our observation of a significant negative correlation of CERAD scores with cortical thickness in the superior parietal region is consistent with previous studies relating CERAD with cortical thickness \cite{santos2011morphological, paajanen2013cerad}.

Lastly, WMH have been implicated in age-related cognitive decline and AD, which is characterized by atrophy in the cortical mantle and the MTL [\cite{rizvi2018effect, du2005white, dadar2022white}]. In our study, we observed significant negative correlations between normalized WMH and thickness in posterior cingulate and superior temporal regions. Previous work \cite{reijmer2015decoupling} showed that the disruption of structural and functional connectivity has an impact on executive functioning and memory among individuals with high WMH volume. To this point, our study found that subcortical atrophy was significantly negatively correlated with WMH volume in caudate and thalamus, suggesting more global effects on brain volume.

In our dataset, amongst the \textcolor{blue}{135} specimens, \textcolor{blue}{82} had AD pathology with existence of co-pathologies. Future studies may apply this dataset and pipeline to help disentangle the differential contributions of unique pathologies to individual atrophy patterns. Separately, we are aware that the pathology measures and MRI segmentation-based measures were obtained from contralateral hemispheres, which could potentially weaken the observed associations. But pathology in AD is usually largely symmetrical between the hemispheres, and therefore leaves less room for biases in the observed correlations, as claimed in a recent study \cite{ravikumar2021ex} which showed that correlations between MTL thickness maps and both contralateral and ipsilateral semi-quantitative p-tau pathology scores did not detect substantially different correlation patterns. Another limitation is that our study relies on semi-quantitative measures of neuropathology, which are subjective and might not reflect a linear pathology burden. We are currently in the process of obtaining neuropathology measurements from the same hemisphere histology, and developing machine learning-based quantitative pathological ratings to further validate our work. The current study provides support for future work to use larger datasets and quantitative pathology measures to describe the contribution of multiple pathologies to brain morphology in neurodegenerative diseases. But, overall we observe a similar trend as described in our recent work \cite{sadaghiani2022associations}, which looked at regional cortical thickness with p-tau burden. These limitations could be avoided by expanding our analysis to a larger dataset, which we are actively working towards.

\textcolor{blue}{Lastly, we should mention that our postmortem imaging project was launched shortly before the COVID-19 pandemic, which greatly interfered with our ability to maintain consistency in formalin fixation. This has been addressed in our Center's more recent autopsies, where we aim for consistent 60 days fixation. Like other limitations, we hypothesize that if formalin fixation had been more consistent, that would only lead to stronger associations between structure and pathology.}

\subsection*{\textbf{Conclusion}}
While there is increased interest in using high-resolution postmortem MRI of the human brain for discovering associations between brain structure and pathology, automated tools for the analysis of such complex images have received much less attention compared to \emph{antemortem} MRI. Our study used a relatively large dataset of \textcolor{blue}{135} high resolution T2w 7~T postmortem whole-hemisphere MRI scans to evaluate multiple deep learning image segmentation architectures and to develop an automatic segmentation pipeline that labels cortical gray matter, four subcortical structures (caudate, globus pallidus, putamen, and thalamus), WMH, and normal-appearing white matter. We report good agreement between thickness measures derived from our deep learning pipeline with the reference standard of semi-automated thickness measurement. Our analysis linking morphometry measures and pathology demonstrated that automated analysis of postmortem MRI yields similar findings to a labor-intensive semi-automated approach, and more broadly, that automated segmentation of postmortem MRI can complement and inform antemortem neuroimaging in neurodegenerative diseases. We have released our pipeline as a stand-alone containerized tool that can be readily applied to other postmortem brain datasets.

\section*{Acknowledgments}
We gratefully acknowledge the tissue donors and their families. We also thank all the staff at the Center for Neurodegenerative Research (University of Pennsylvania) for performing the autopsies and making the tissue available for this project.

\section*{Ethics approval and consent to participate}
Human brain specimens were obtained in accordance with local laws and regulations, and includes informed consent from next of kin at time of death or where possible, pre-consent during life.

\section*{Funding}
This work was supported in part by the National Institute of Health Grants: P30 AG072979, R01 AG056014, RF1 AG069474, R01 AG054519, P01 AG017586, U19 AG062418, R01-NS-109260 and P01-AG-066597.

\section*{Availability of data and materials}
We have provided the code, scripts and Jupyter notebooks to reproduce the findings of this study at the \href{https://pulkit-khandelwal.github.io/exvivo-brain-upenn/}{\textbf{project webpage}}. The MRI data will be available upon request due to compliance and ethical issues.

\section*{Consent for publication}
All authors have reviewed the contents of the manuscript being submitted, approved of its contents and validated the accuracy of the data and consented to publication.

\section*{Competing interests}
D.A.W has received grant support from Merck, Biogen, and Eli
Lilly/Avid. D.A.W received consultation fees from Neuronix, Eli Lilly,
and Qynaps and is on the Data and Safety Monitoring Board for a clinical trial run by Functional Neuromodulation. J.Q.T. received revenue from the sale of Avid to Eli Lilly as co-inventor on imaging-related
patents submitted by the University of Pennsylvania. D.J.I. is member of science advisory board of Denali Therapeutics. S.R.D. received
consultation fees from Rancho Biosciences and Nia Therapeutics. The
other authors have nothing to disclose.

\section*{CRediT authorship contribution statement}
\textbf{Pulkit Khandelwal:} Conceptualization, Data curation, Formal analysis, Investigation, Methodology, Project administration, Resources, Software, Supervision, Validation, Visualization, Roles/Writing - original draft, Writing - review and editing.
\textbf{Michael Tran Duong:} Conceptualization, Data curation, Formal analysis, Investigation, Methodology, Validation, Writing - review and editing.
\textbf{Shokufeh Sadaghiani:} Conceptualization, Data curation, Resources, Methodology.
\textbf{Sydney Lim:} Data curation, Resources, Methodology.
\textbf{ Amanda Denning:} Data curation, Resources, Methodology.
\textbf{ Eunice Chung:} Data curation, Resources, Methodology.
\textbf{Sadhana Ravikumar:} Conceptualization, Validation, Writing - review and editing.
\textbf{Sanaz Arezoumandan:} Data curation, Resources.
\textbf{Claire Peterson:} Data curation, Resources.
\textbf{ Madigan Bedard:} Data curation, Resources.
\textbf{Noah Capp:} Data curation, Resources,. Software.
\textbf{Ranjit Ittyerah:} Data curation, Resources.
\textbf{Elyse Migdal:} Data curation, Resources, Methodology.
\textbf{Grace Choi:} Data curation, Resources, Methodology.
\textbf{Emily Kopp:} Data curation, Resources, Methodology.
\textbf{Bridget Loja:} Data curation, Resources, Methodology.
\textbf{Eusha Hasan:} Data curation, Resources, Methodology.
\textbf{Jiacheng Li:} Data curation, Resources, Methodology.
\textbf{Alejandra Bahena:} Data curation, Resources.
\textbf{Karthik Prabhakaran:} Data curation, Resources.
\textbf{Gabor Mizsei:} Data curation, Resources.
\textbf{Marianna Gabrielyan:} Data curation, Resources.
\textbf{Theresa Schuck:} Data curation, Resources.
\textbf{Winifred Trotman:} Data curation, Project administration, Resources.
\textbf{John Robinson:} Data curation, Resources.
\textbf{Daniel T. Ohm:} Data curation, Resources, Project administration.
\textbf{Edward B. Lee:} Data curation, Funding acquisition, Investigation, Project administration, Resources, Writing - review and editing.
\textbf{John Q. Trojanowski:} Data curation, Project administration, Resources, Funding acquisition.
\textbf{Corey McMillan:} Data curation, Resources, Funding acquisition.
\textbf{Murray Grossman:} Conceptualization, Data curation, Funding acquisition.
\textbf{David J. Irwin:} Data curation, Project administration, Resources, Funding acquisition, Validation, Writing - review and editing.
\textbf{John A. Detre:} Data curation, Project administration, Resources, Validation, Writing - review and editing.
\textbf{M. Dylan Tisdall:} Data curation, Project administration, Resources, Validation, Writing - review and editing.
\textbf{ Sandhitsu R. Das:} Conceptualization, Data curation, Formal analysis, Investigation, Methodology, Project administration, Resources, Supervision, Validation, Writing - review and editing.
\textbf{Laura E.M. Wisse:} Conceptualization, Data curation, Formal analysis, Investigation, Methodology, Project administration, Resources, Supervision, Validation, Writing - review and editing.
\textbf{David A. Wolk:} Conceptualization, Data curation, Funding acquisition, Investigation, Methodology, Project administration, Resources, Software, Supervision, Validation, Visualization, Roles/Writing - original draft, Writing - review and editing.
\textbf{Paul A. Yushkevich:} Conceptualization, Data curation, Formal analysis, Funding acquisition, Investigation, Methodology, Project administration, Resources, Software, Supervision, Validation, Writing - review and editing.

\section*{******* \textcolor{red}{\textbf{SUPPLEMENTARY}} *******}
\section*{Architectural details of the nine neural networks}

%\label{supp_sec:nine_nn_arch_details}
\textbf{3D Unet-like:} We implement a custom in-house 3D Unet-like architecture \cite{khandelwal2020domain}, where the input and the output have dimensions \emph{in}xHxWxD, and \emph{out}xHxWxD. The first dimension represents the number of channels, and H, W and D denote the height, width and depth respectively. There are five encoder blocks (with the last block acting as the bottleneck layer), and four decoder blocks, with skip connections. We have used \emph{groupnorm} as the normalization method. Note that the first encoder block does not consist of the MaxPool operation, and the first decoder block does not consist of the first set of conv3D, groupnorm, and ReLU units. The \textit{\textbf{encoder}} block has the following \textit{output} dimensions at each of the encoder block: E1: 16xHxWxD, E2: 32x$\frac{H}{2}$x$\frac{W}{2}$x$\frac{D}{2}$, E3: 64x$\frac{H}{4}$x$\frac{W}{4}$x$\frac{D}{4}$, E4: 128x$\frac{H}{8}$x$\frac{W}{8}$x$\frac{D}{8}$, E5: 256x$\frac{H}{16}$x$\frac{W}{16}$x$\frac{D}{16}$. The \textit{\textbf{decoder}} block has the following \textit{output} dimensions at each of the decoder block: D1: 256x$\frac{H}{8}$x$\frac{W}{8}$x$\frac{D}{8}$, D2: 128x$\frac{H}{4}$x$\frac{W}{4}$x$\frac{D}{4}$, D3: 64x$\frac{H}{2}$x$\frac{W}{2}$x$\frac{D}{2}$, D4: 32xHxWxD.

\textbf{VNet:} We implement the following variant of VNet \cite{milletari2016v}. The input and the output volumes have dimensions \emph{in}xHxWxD (here: \emph{in} is 1), and \emph{out}xHxWxD respectively, where \emph{out} represents the number of output classes. We have four different modules: an InputTransition, DownTransition, UpTransition, and OutputTransition. The InputTransition comprises of a 3D convolution block with input channels \emph{in} and produces \emph{ch} output channels by using a kernel size of 5 and padding of 2, followed by BatchNorm. Next, the input is added to the output of the convolutional block to obtain a residual function and then followed by PReLU, to produce an output with dimensions, \emph{ch}xHxWxD. This output is then fed to the first DownTransition module. The DownTransition block consists of 3D convolution block, with kernel size of 2 and stride of 2, which takes in the input \emph{ch}xHxWxD and downsamples the feature map by a factor of 2 and doubles the number of channels, thus giving an output of (\emph{2$\times$ch})x$\frac{H}{2}$x$\frac{W}{2}$x$\frac{D}{2}$, which is then followed by BatchNorm and PReLU nonlinearity and an optional dropout. A residual function is then created by using another convolutional block by using a kernel size of 5 and padding of 2, followed by PReLU nonlinearity. The second DownTransition block produces an output of (\emph{4$\times$ch})x$\frac{H}{4}$x$\frac{W}{4}$x$\frac{D}{4}$. The third and fourth DownTransition blocks has the dropout set to True and produces an output of (\emph{8$\times$ch})x$\frac{H}{8}$x$\frac{W}{8}$x$\frac{D}{8}$ and (\emph{16$\times$ch})x$\frac{H}{16}$x$\frac{W}{16}$x$\frac{D}{16}$ respectively. The UpTransition block is similar to the DownTransition with the difference that every block upsamples the input using 3D transpose convolutions, and has skip connections like UNet from the corresponding DownTransition block. The first UpTransition block upsamples from  (\emph{16$\times$ch})x$\frac{H}{16}$x$\frac{W}{16}$x$\frac{D}{16}$ to (\emph{16$\times$ch})x$\frac{H}{8}$x$\frac{W}{8}$x$\frac{D}{8}$ with dropout set to true. The next UpTransition block upsamples to (\emph{8$\times$ch})x$\frac{H}{4}$x$\frac{W}{4}$x$\frac{D}{4}$ with dropout set to true. The next two UpTransition blocks produces an output of (\emph{4$\times$ch})x$\frac{H}{2}$x$\frac{W}{2}$x$\frac{D}{2}$ and (\emph{2$\times$ch})xHxWxD respectively. Finally, the last block OutputTransition produces the desired output of \emph{out}xHxWxD using two 3D convolutional layers.

\textbf{VoxResNet:} We implement the following variant of VoxResNet \cite{chen2018voxresnet}. In our implementation, we define a building block called VoxRes, which creates a residual function, comprising two sets of BatchNorm, ReLU and 3D convolution with kernel size of 3, stride of 1 and padding of 1. In the VoxResNet architecture, there are four encoder and four decoder blocks. The first encoder block consists of a 3D convolution with kernel size 3 and padding of 1, BatchNorm, ReLU and 3D convolution, thus changing the input of \emph{in}xHxWxD to \emph{ch}xHxWxD, where \emph{ch}=32. The next three encoder block successively downsamples the input from \emph{ch}xHxWxD to (\emph{2$\times$ch})x$\frac{H}{2}$x$\frac{W}{2}$x$\frac{D}{2}$, (\emph{2$\times$ch})x$\frac{H}{4}$x$\frac{W}{4}$x$\frac{D}{4}$ and (\emph{2$\times$ch})x$\frac{H}{8}$x$\frac{W}{8}$x$\frac{D}{8}$ respectively by stacking BatchNorm, ReLU, 3D convolution, with kernel size of 3, stride of (1, 2, 2) and padding o0f 1, and VoxResNet module. The four decoder blocks takes in the output of the corresponding encoder block. The decoder block transforms the output of the first encoder from \emph{ch}xHxWxD to \emph{ch}xHxWxD using transposed convolution and then from \emph{ch}xHxWxD  to \emph{out}xHxWxD  using convolution where \emph{out} represents the number of output classes. The second decoder block takes in output from the corresponding encoder block to produce an output from (\emph{2$\times$ch})x$\frac{H}{2}$x$\frac{W}{2}$x$\frac{D}{2}$ to \emph{out}xHxWxD using transposed convolution with 64 filters and a 3D convolution with kernel of (1, 2, 2) and stride of (1, 2, 2). Similarly, the third and fourth decoder block takes in output from the corresponding encoder block to produce an output from (\emph{2$\times$ch})x$\frac{H}{4}$x$\frac{W}{4}$x$\frac{D}{4}$ to \emph{out}xHxWxD using transposed convolution with 64 filters with kernel of (1, 4, 4) and stride of (1, 4, 4) and a 3D convolution; and from (\emph{2$\times$ch})x$\frac{H}{8}$x$\frac{W}{8}$x$\frac{D}{8}$ to \emph{out}xHxWxD using transposed convolution with 64 filters with kernel of (1, 8, 8) and stride of (1, 8, 8) and a 3D convolution. The outputs from the four decoder blocks are then added element-wise to produce the final output of size \emph{out}xHxWxD.

\textbf{Attention U-Net:} The building block of Attention U-Net \cite{oktay2018attention} is the AttentionGate module which comprises of a gating mechanism to suppress irrelevant background. The AttentionGate takes in two volumes $G$ and $X$, of size \emph{$ch_{g}$}x$H_{g}$x$W_{g}$x$D_{g}$ and \emph{$ch_{x}$}x$H_{x}$x$W_{x}$x$D_{x}$ respectively, and produce outputs of size \emph{$ch_{inter}$}x$H_{g}$x$W_{g}$x$D_{g}$ using 3D convolutions with kernel size and padding of 1. The two outputs are then summed up element-wise and passed through ReLU activation, followed by another convolution, with kernel size and padding of 1, and a sigmoid activation function. Finally, this output is upsampled and then multiplied by the input volume $X$ to \emph{$ch_{x}$}x$H_{x}$x$W_{x}$x$D_{x}$ to produce the final volume, $\hat{X}$, with dimensions \emph{$ch_{x}$}x$H_{x}$x$W_{x}$x$D_{x}$.

The Attention U-Net architecture mimics a standard U-Net with an AttentionGate mechanism taking in input from the decoder's output for upsampling and the corresponding encoder. The four encoder blocks take in the volume of \emph{in}xHxWxD, where $\emph{in}$ is 1, and successively reduce the resolution to (\emph{64$\times$ch})x$\frac{H}{2}$x$\frac{W}{2}$x$\frac{D}{2}$, (\emph{128$\times$ch})x$\frac{H}{4}$x$\frac{W}{4}$x$\frac{D}{4}$,
(\emph{256$\times$ch})x$\frac{H}{8}$x$\frac{W}{8}$x$\frac{D}{8}$, and (\emph{512$\times$ch})x$\frac{H}{16}$x$\frac{W}{16}$x$\frac{D}{16}$ by using convolution block comprising of a stack of two convolutions with kernel size of (2,2,2) and two ReLU layers.

Now, the first AttentionGate takes in as inputs, the outputs of the third and the fourth encoder blocks, and produces an output of dimension (\emph{256$\times$ch})x$\frac{H}{8}$x$\frac{W}{8}$x$\frac{D}{8}$, which is then concatenated with the upsampled version of the output of the fourth encoder block (\emph{256$\times$ch})x$\frac{H}{8}$x$\frac{W}{8}$x$\frac{D}{8}$, and then this output is again upsampled to (\emph{128$\times$ch})x$\frac{H}{8}$x$\frac{W}{8}$x$\frac{D}{8}$. This procedure is repeated two more times to get output dimensions of (\emph{64$\times$ch})x$\frac{H}{4}$x$\frac{W}{4}$x$\frac{D}{4}$ and (\emph{64$\times$ch})x$\frac{H}{2}$x$\frac{W}{2}$x$\frac{D}{2}$ respectively. A final upsampling block then produces the desired output of $\emph{out}$xHxWxD.

\textbf{AnatomyNet (Vanilla):} AnatomyNet \cite{zhu2019anatomynet} is a variant of U-Net which takes advantage of the Squeeze-and-Excitation residual feature (SE) blocks proposed in \cite{rickmann2019project, roy2018recalibrating} to segment small anatomical structures, which are often missed by networks such as U-Net. We implement four variants of the AnatomyNet, with the first variant AnatomyNet (Vanilla) without any SE blocks. The first block takes an input with size \emph{in}xHxWxD and generates a feature map of size \emph{ch}xHxWxD, here $in$ is 1 and $ch$ is 28, using 3D convolution with kernel size of 3, stride of 2, and padding of 1; followed by a BatchNorm and LeakyReLU activation layer.

The next three blocks take in the input \emph{ch}xHxWxD, and successively increase the feature channels and downsamples by 2 from (28)xHxWxD to (34)x$\frac{H}{2}$x$\frac{W}{2}$x$\frac{D}{2}$. Each of these three blocks consists of two repetitions of a ResidualBasicBlock. The ResidualBasicBlock comprises of a 3D convolution with kernel size of 3, stride of 2, and padding of 1, followed by BatchNorm, LeakyReLU, 3D convolution with kernel size of 3, stride of 2, and padding of 1, BatchNorm, SEBlock and a downsampling 3D convolution with kernel size of 3, stride of 1. The SEBlock consists of one of the three variants of the Squeeze-and-Excitation blocks: Spatial Excitation AnatomyNet (SE), Channel Excitation AnatomyNet (CE) or the Channel-spatial Excitation AnatomyNet (CE + SE). But, here we do not include the SEBlock in the ResidualBasicBlock in the AnatomyNet (Vanilla) variant.

The next block in the network architecture is the ResidualUPBasicBlock, which concatenates two inputs of size (34)x$\frac{H}{2}$x$\frac{W}{2}$x$\frac{D}{2}$ and (32)x$\frac{H}{2}$x$\frac{W}{2}$x$\frac{D}{2}$ to produce (66)x$\frac{H}{2}$x$\frac{W}{2}$x$\frac{D}{2}$, followed by a 3D convolution with kernel size of 3, stride of 2, and padding of 1, followed by BatchNorm, LeakyReLU, 3D convolution with kernel size of 3, stride of 2, and padding of 1, BatchNorm, SEBlock and a downsampling 3D convolution with kernel size of 3 and stride of 1, and LeakyReLU to produce (32)x$\frac{H}{2}$x$\frac{W}{2}$x$\frac{D}{2}$. This is followed by three sets of ResidualBasicBlock, without the downsampling layer, and the ResidualUPBasicBlock to produce an output of size (14)xHxWxD. A final 3D convolution is employed to produce the output of (out)xHxWxD, where $out$ is the number of segmentation classes.

\textbf{Channel Squeeze and Spatial Excitation AnatomyNet (sSE):} The architecture of the Channel Squeeze and Spatial Excitation AnatomyNet (sSE) is similar to the \textit{Vanilla} variant, with the only difference being that the SEBlock consists of Spatial excitation module. The SEBlock squeezes the feature maps along the channel dimension and excites the spatial dimension. This block helps in segmentation of fine-grained structures as proposed in \cite{roy2018recalibrating}. This block consists of a 3D convolution with kernel size of 1 and stride of 1, followed by a Sigmoid activation layer which takes an input with $ch$ channels, $ch$x$H$x$W$x$D$,  and reduces it to $1$x$H$x$W$x$D$. The channel-squeezed output is then multiplied element-wise with the input feature map to produce the spatial-excited feature map of dimension $ch$x$H$x$W$x$D$.

\textbf{Spatial Squeeze and Channel Excitation AnatomyNet (cSE):} The architecture of the Spatial Squeeze and Channel Excitation AnatomyNet (cSE) is similar to the \textit{Vanilla} variant, with the only difference being that the SEBlock consists of a Channel excitation module. This SEBlock consists of the following layers: the spatial channels of dimension $ch$x$H$x$W$x$D$, which is spatially squeezed to $ch$x$1$x$1$x$1$ using adaptive pooling. Two layers of fully connected layer and ReLU activation function transforms this feature map to a vector of length $ch$. This feature map is then element-wise multiplied with the original input feature map to produce a channel-excited map of $ch$x$H$x$W$x$D$.

\textbf{Spatial and Channel Squeeze and Excitation AnatomyNet (scSE):} The architecture of the Spatial excitation AnatomyNet (SE) is similar to the \textit{Vanilla} variant, with the only difference being that the SEBlock consists of Channel-spatial excitation module. The input tensor of size $ch$x$H$x$W$x$D$ is passed through both $cSE$ and $sSE$. An element-wise max is taken between these feature maps to produce the final output of size $ch$x$H$x$W$x$D$.

\textbf{nnU-Net:} We use the default network in the nnU-Net framework as described in Isensee et al \cite{isensee2021nnu}.

\setcounter{figure}{0}
\begin{figure}[H]
\centering
    \includegraphics[width=\textwidth,height=\textheight,keepaspectratio]{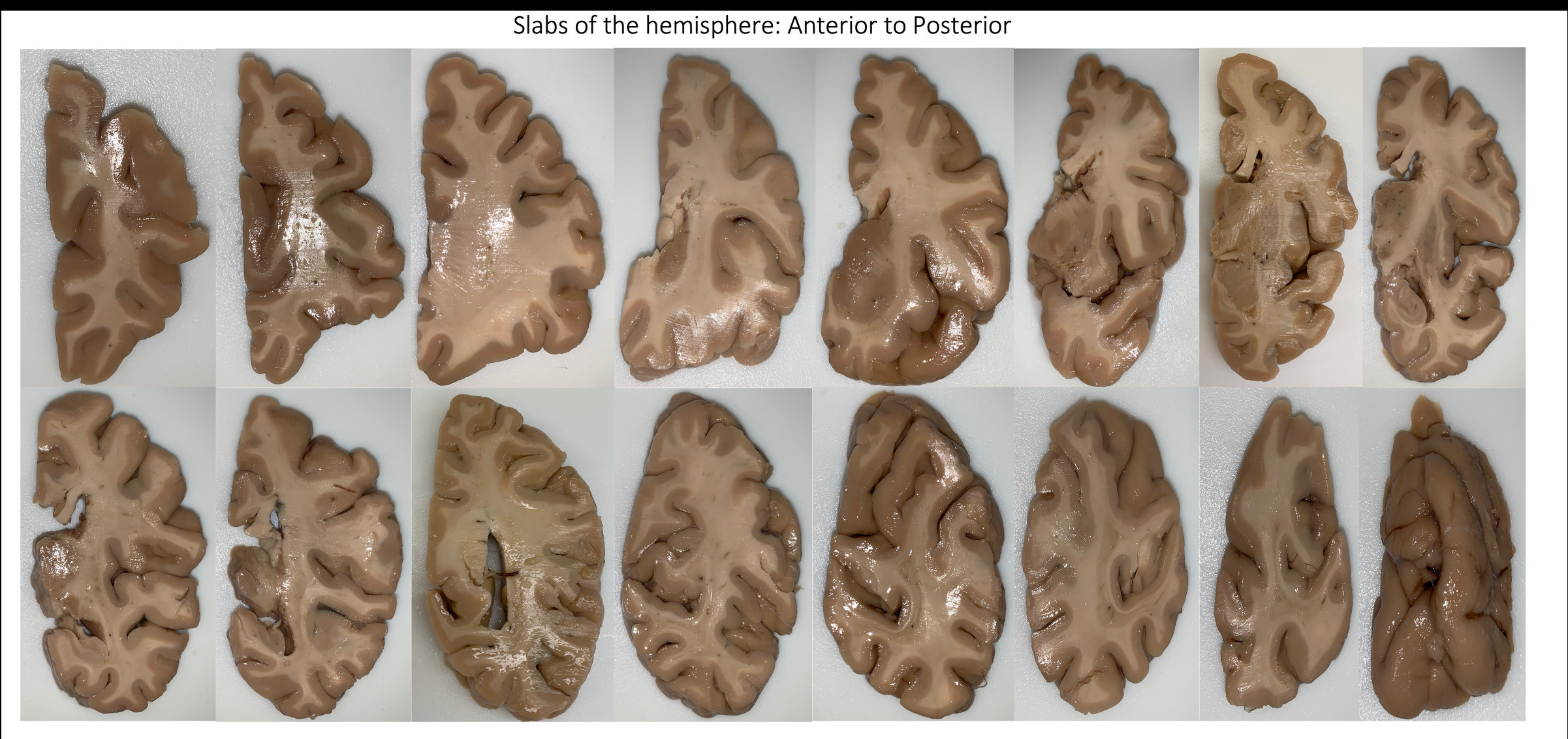}
        \caption{Postmortem tissue blockface photograph of a patient with Parkinson's disease (not demented) and Lewy body disease, deceased at the age of 79. Shown are the slabbed blockface images from anterior to posterior.}
        \label{figure_autopsy_109949R}
\end{figure}

\begin{figure}[H]
\centering
    \includegraphics[width=\textwidth,height=\textheight,keepaspectratio]{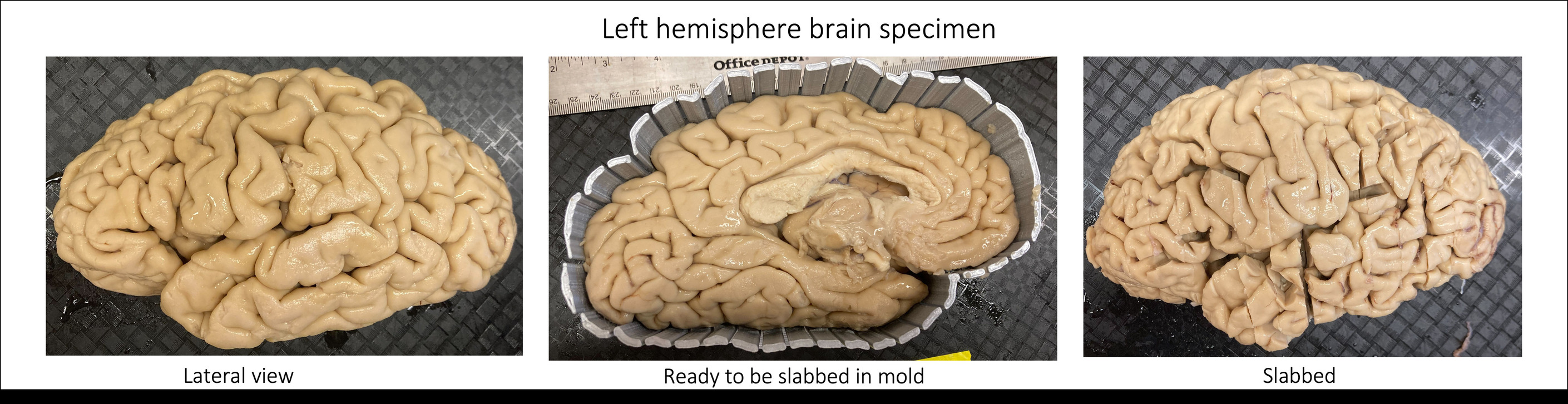}
        \caption{Post-mortem tissue blockface photograph of a patient with FTLD primary-progressive-aphasia (PNFA) and globular glial tauopathy (GGT) disease, deceased at the age of 74. Shown are the lateral and medial views of the left hemisphere. The tissue is then placed in a mold and is subsequently slabbed.}
        \label{figure_autopsy_121200L}
\end{figure}

\begin{figure}
\centering
\includegraphics[width=\textwidth,height=\textheight,keepaspectratio]{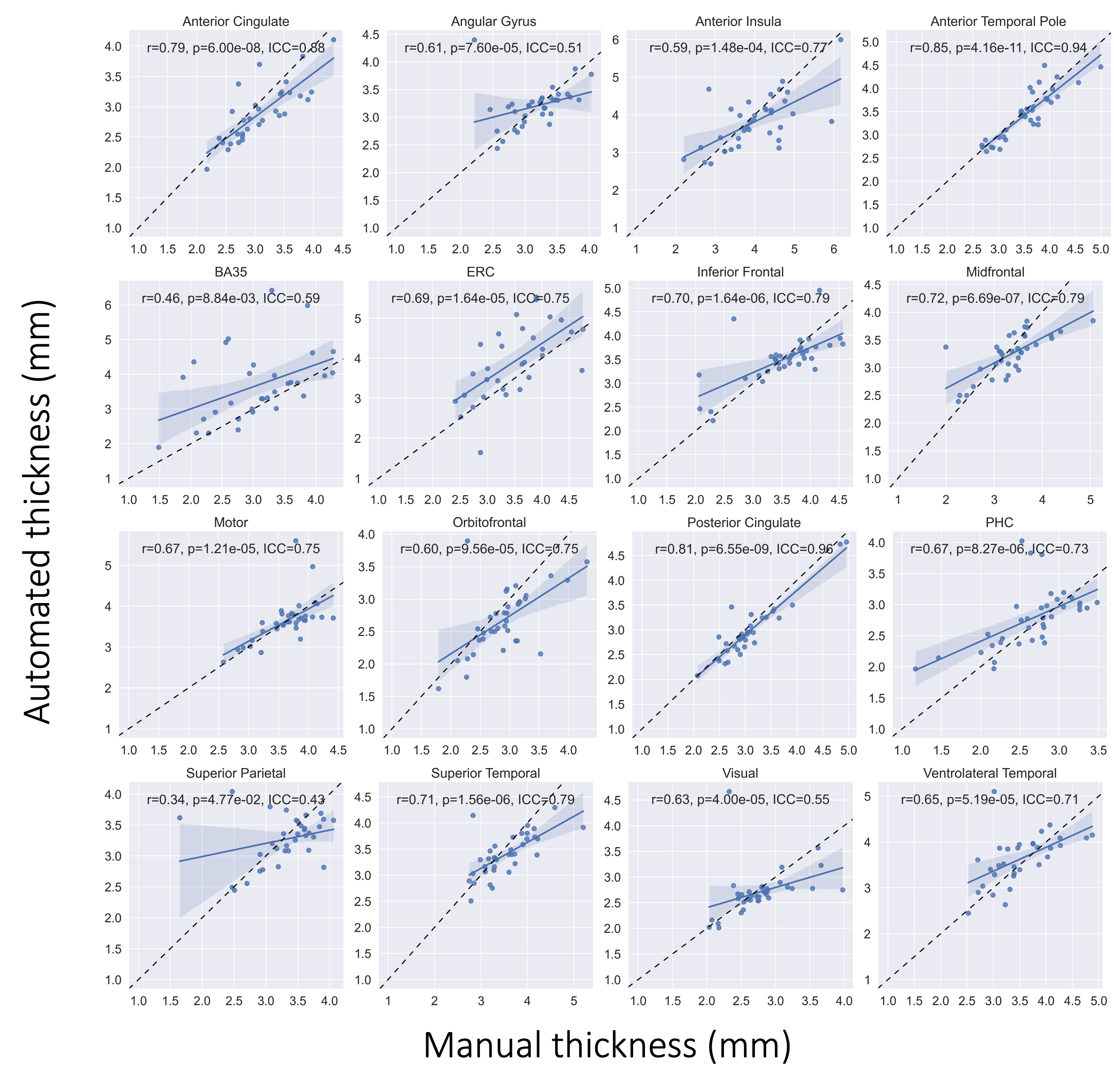}
        \caption{\textcolor{blue}{Regional cortical thickness measurements. Figure shows correlation plots between cortical thickness measured by the automated nnU-Net-CRUISE (y-axis) and reference standard manual segmentations (x-axis). Within each plot, we tabulate Spearman’s correlation coefficient (r), \textit{p} value, and the Average fixed raters Intra-class correlation coefficient (ICC) scores. The dashed line indicates $y=x$.}}
        \label{fig:corr_manual_automated_thickness}
\end{figure}

\setcounter{table}{0}
%\section{Tables}
\begin{adjustbox}{angle=0, caption={\textcolor{blue}{Brief demographic summary of the subjects used for training and evaluation of the deep learning models in the current study. Abbreviations: AD: Alzheimer's Disease, ALS: amyotrophic lateral sclerosis, CVD: cerebrovascular disease, LATE: limbic-predominant age-related TDP-tau 43 encephalopathy, LBD: Lewy body disease, CTE: chronic traumatic encephalopathy, FTLD-TDP: frontotemporal lobar degeneration with TDP inclusions, GGT: globular glial p-tauopathy, CVD: cerebrovascular disease, CBD: corticobasal degeneration, PART: Primary age-related p-tauopathy, PSP: progressive supranuclear palsy, p-tau-Misc: p-tauopathy unclassifiable.}},float=table}
\scalebox{0.50}
{
\label{table_demographics}
\begin{tabular}{|c|c|c|c|c|c|c|c|c|c|l|}
\hline
\textbf{Subject} & \textbf{AD Spectrum} & \textbf{Primary} & \textbf{Secondary}     & \textbf{Age} & \textbf{Race} & \textbf{Sex} & \textbf{\begin{tabular}[c]{@{}c@{}}Cortex\\ cross-val\end{tabular}} & \textbf{\begin{tabular}[c]{@{}c@{}}Subcortical\\ cross-val\end{tabular}} & \textbf{\begin{tabular}[c]{@{}c@{}}WMH\\ cross-val\end{tabular}} & \textbf{WM} \\ \hline
2                        & Yes                  & AD               & LATE+AD                & 87           & White         & M            &                                                                     &                                                                          & Y                                                                &             \\ \hline

4                        & Yes                  & AD               & LATE+AD                & 89           & White         & F            &                                                                     & Y                                                                        &                                                                  & Y           \\ \hline

9                        & Yes                  & AD               & LBD+LATE+AD            & 63           & White         & F            &                                                                     &                                                                          & Y                                                                &             \\ \hline

13                   & Yes                  & AD               & AD                     & 73           & White         & M            & Y                                                                   & Y                                                                        & Y                                                                &             \\ \hline

17                     & Yes                  & LBD              & LBD+AD+CTE             & 71           & White         & M            & Y                                                                   & Y                                                                        & Y                                                                &             \\ \hline

20                     & Yes                  & LBD              & LBD+AD                 & 73           & Black         & M            & Y                                                                   & Y                                                                        & Y                                                                &             \\ \hline

26                      & No                   & FTLD-TDP         & FTLD-TDP43             & 65           & White         & M            & Y                                                                   & Y                                                                        & Y                                                                &             \\ \hline

29                    & No                   & FTLD-TDP         & FTLD+AD                & 81           & White         & M            & Y                                                                   & Y                                                                        &                                                                  &             \\ \hline

31                     & No                   & Tau-Misc         & FTLD                   & 50           & White         & M            &                                                                     &                                                                          & Y                                                                &             \\ \hline

35                     & No                   & Tau-4R           & CBD                    & 77           & White         & F            & Y                                                                   & Y                                                                        &                                                                  & Y           \\ \hline

37                   & No                   & Tau-4R           & CBD+AD                 & 76           & White         & M            &                                                                     &                                                                          & Y                                                                &             \\ \hline
38                     & Yes                  & AD               & LATE+AD                & 83           & White         & F            &                                                                     &                                                                          & Y                                                                &             \\ \hline
\end{tabular}
}
\end{adjustbox}

\begin{adjustbox}{angle=0, caption={The locations from where the neuropathology ratings were obtained from, either the exact (main regions) or the closest (exploratory regions) to the cortical brain regions.},float=table}
\scalebox{0.80}
{
\label{table_neuropath}
\centering
\begin{tabular}{|c|c|c|}
\hline
\textbf{Brain region}                                                                                              & \textbf{Matching pathology region} & \textbf{Exact} \\ \hline
Visual cortex                          & Occipital cortex          & Y \\ \hline
Middle frontal gyrus                   & Middle frontal gyrus      & Y \\ \hline
Orbitofrontal gyrus                    & Orbital frontal cortex    & Y \\ \hline
Anterior cingulate                     & Cingulate gyrus           & Y \\ \hline
Posterior cingulate                    & Cingulate gyrus           & Y \\ \hline
Motor cortex                           & Motor cortex              & Y \\ \hline
Angular gyrus                          & Angular gyrus             & Y \\ \hline
Superior parietal cortex               & Angular gyrus             & N  \\ \hline
Superior temporal cortex               & Superior/ middle temporal & Y \\ \hline
Anterior temporal pole                 & Amygdala                  & N  \\ \hline
Anterior insula                        & Middle frontal gyrus      & N  \\ \hline
\begin{tabular}[c]{@{}c@{}}Ventrolateral part of anterior temporal cortex\\ (Inferior temporal gyrus)\end{tabular} & Entorhinal cortex                  & N             \\ \hline
Inferior frontal cortex (Broca's area) & Middle frontal gyrus      & N  \\ \hline
Entorhinal cortex                      & Entorhinal cortex         & Y \\ \hline
Brodmann area 35                       & Entorhinal cortex         & Y \\ \hline
Parahippocampal cortex                 & CA1/ Subiculum            & N  \\ \hline
\end{tabular}
}
\end{adjustbox}

%%%%%%%%%%%%
\begin{adjustbox}{angle=0, caption={\textcolor{blue}{Associations between morphometric measures and summary measures of tau pathology and neuronal loss in the medial temporal lobe, a region linked to early neurodegeneration in Alzheimer’s disease. Shown is the one-sided Spearman's correlation (controlling for age, sex and PMI) between the cortical thickness measures derived from the topologically corrected nnU-Net-CRUISE gray matter segmentation with corresponding MTL ratings of p-tau pathology and neuronal loss density. Each cell is color coded with darker shades indicating more negative correlations. CI indicates 95\% confidence interval. The asterisk indicates that the test survived Bonferroni multiple testing correction. \underline{Legend} *:0.01 $<$ \emph{p} $\leq$ 0.05; **:0.001 $<$ \emph{p} $\leq$ 0.01; ***:0.0001 $<$ \emph{p} $\leq$ 0.001;  ****:0.00001 $<$ \emph{p} $\leq$ 0.0001.}},float=table}
\label{table:path_all}
\scalebox{0.60}
{
\centering
\label{tab:my-table}
\begin{tabular}{|c|cc|cc|}
\hline
\textbf{Pathology ratings} & \multicolumn{2}{c|}{\cellcolor[HTML]{FD6864}\textbf{p-tau (MTL)}}                       & \multicolumn{2}{c|}{\cellcolor[HTML]{F8A102}\textbf{Neuronal Loss (MTL)}}               \\ \hline
\textbf{ROI}               & \multicolumn{1}{c|}{\textbf{rho}} & \textbf{CI}       & \multicolumn{1}{c|}{\textbf{rho}} & \textbf{CI}       \\ \hline
\textbf{Visual}               & \multicolumn{1}{c|}{\gradientNeuro{-0.1}}          & {[}-1.0, 0.09{]}  & \multicolumn{1}{c|}{\gradientNeuro{-0.105}}        & {[}-1.0, 0.08{]}  \\ \hline
\textbf{Motor cortex}               & \multicolumn{1}{c|}{-}                     & -                & \multicolumn{1}{c|}{-}                     & -                \\ \hline
\textbf{Posterior cingulate}              & \multicolumn{1}{c|}{\gradientNeuro{-0.337}}         & {[}-1.0, -0.13{]} & \multicolumn{1}{c|}{\gradientNeuro{-0.396}***}       & {[}-1.0, -0.2{]}  \\ \hline
\textbf{Midfrontal}                & \multicolumn{1}{c|}{\gradientNeuro{-0.202}}         & {[}-1.0, -0.01{]} & \multicolumn{1}{c|}{\gradientNeuro{-0.177}}        & {[}-1.0, 0.01{]}  \\ \hline
\textbf{Anterior cingulate}              & \multicolumn{1}{c|}{\gradientNeuro{-0.215}}        & {[}-1.0, -0.01{]} & \multicolumn{1}{c|}{\gradientNeuro{0.012}}         & {[}-1.0, 0.22{]}  \\ \hline
\textbf{Orbitofrontal}               & \multicolumn{1}{c|}{-}                     & -                & \multicolumn{1}{c|}{-}                     & -                \\ \hline
\textbf{Superior temporal pole}             & \multicolumn{1}{c|}{\gradientNeuro{-0.115}}        & {[}-1.0, 0.08{]}  & \multicolumn{1}{c|}{\gradientNeuro{-0.153}}        & {[}-1.0, 0.04{]}  \\ \hline
\textbf{Inferior frontal}                & \multicolumn{1}{c|}{\gradientNeuro{-0.027}}         & {[}-1.0, 0.16{]}  & \multicolumn{1}{c|}{\gradientNeuro{-0.033}}        & {[}-1.0, 0.15{]}  \\ \hline
\textbf{Anterior insula}             & \multicolumn{1}{c|}{\gradientNeuro{-0.228}}        & {[}-1.0, -0.04{]} & \multicolumn{1}{c|}{\gradientNeuro{-0.213}}        & {[}-1.0, -0.03{]} \\ \hline
\textbf{Anterior temporal pole}            & \multicolumn{1}{c|}{\gradientNeuro{-0.226}}        & {[}-1.0, -0.04{]} & \multicolumn{1}{c|}{\gradientNeuro{-0.204}}        & {[}-1.0, -0.02{]} \\ \hline
\textbf{Ventrolateral temporal cortex}               & \multicolumn{1}{c|}{\gradientNeuro{-0.14}}         & {[}-1.0, 0.09{]}  & \multicolumn{1}{c|}{\gradientNeuro{-0.049}}        & {[}-1.0, 0.18{]}  \\ \hline
\textbf{Superior parietal}                & \multicolumn{1}{c|}{\gradientNeuro{-0.194}}         & {[}-1.0, 0.0{]}   & \multicolumn{1}{c|}{\gradientNeuro{-0.063}}        & {[}-1.0, 0.14{]}  \\ \hline
\textbf{Angular gyrus}               & \multicolumn{1}{c|}{\gradientNeuro{-0.285}}        & {[}-1.0, -0.1{]}  & \multicolumn{1}{c|}{\gradientNeuro{-0.135}}        & {[}-1.0, 0.05{]}  \\ \hline
\textbf{Entorhinal cortex} & \multicolumn{1}{c|}{\gradientNeuro{-0.53}****} & {[}-1.0, -0.38{]} & \multicolumn{1}{c|}{\gradientNeuro{-0.47}****} & {[}-1.0, -0.31{]} \\ \hline
\textbf{Brodmann Area 35}              & \multicolumn{1}{c|}{\gradientNeuro{-0.18}}         & {[}-1.0, 0.01{]}  & \multicolumn{1}{c|}{\gradientNeuro{-0.321}**}       & {[}-1.0, -0.14{]} \\ \hline
\textbf{Parahippocampal cortex}               & \multicolumn{1}{c|}{\gradientNeuro{-0.19}}         & {[}-1.0, -0.0{]}  & \multicolumn{1}{c|}{\gradientNeuro{-0.195}}        & {[}-1.0, -0.01{]} \\ \hline
    \end{tabular}
}
\end{adjustbox}

%%%%%%%%%%
\begin{adjustbox}{angle=90, caption={\textcolor{blue}{Morphometry associations with underlying Neuropathology. Shown is the one-sided Spearman's correlation (controlling for age, sex and PMI) between regional cortical thickness measures derived from topologically corrected nnU-Net-CRUISE gray matter segmentation with corresponding regional and the medial temporal lobe (MTL) ratings of p-tau pathology, neuronal loss density, and global amyloid-$\beta$ ratings, CERAD score, and Braak staging. The analyses is repeated for thickness measurements based-on manual segmentations. Each cell is color coded with darker shades indicating more negative correlations. The asterisk indicates that the test survived Bonferroni multiple testing correction \cite{bonferroni1935calcolo}.}},float=table}
\label{table:crusie-manual}
\scalebox{0.45}
{
\centering
\begin{tabular}{|l|ll|ll|ll|cl|ll|ll|ll|}
\hline
\textbf{Pathology ratings} &
  \multicolumn{2}{l|}{\cellcolor[HTML]{FFCCC9}\textbf{Abeta}} &
  \multicolumn{2}{l|}{\cellcolor[HTML]{FFCE93}\textbf{Braak}} &
  \multicolumn{2}{l|}{\cellcolor[HTML]{FFFC9E}\textbf{CERAD}} &
  \multicolumn{2}{c|}{\cellcolor[HTML]{FD6864}\textbf{p-tau}} &
  \multicolumn{2}{l|}{\cellcolor[HTML]{F8A102}\textbf{Neuronal loss}} &
  \multicolumn{2}{l|}{\cellcolor[HTML]{FD6864}\textbf{p-tau (MTL)}} &
  \multicolumn{2}{l|}{\cellcolor[HTML]{F8A102}\textbf{Neuronal loss   (MTL)}} \\ \hline
\textbf{ROI} &
  \multicolumn{1}{l|}{\textbf{nnU-Net-CRUISE}} &
  \textbf{Manual} &
  \multicolumn{1}{l|}{\textbf{nnU-Net-CRUISE}} &
  \textbf{Manual} &
  \multicolumn{1}{l|}{\textbf{nnU-Net-CRUISE}} &
  \textbf{Manual} &
  \multicolumn{1}{c|}{\textbf{nnU-Net-CRUISE}} &
  \textbf{Manual} &
  \multicolumn{1}{l|}{\textbf{nnU-Net-CRUISE}} &
  \textbf{Manual} &
  \multicolumn{1}{l|}{\textbf{nnU-Net-CRUISE}} &
  \textbf{Manual} &
  \multicolumn{1}{l|}{\textbf{nnU-Net-CRUISE}} &
  \textbf{Manual} \\ \hline
\textbf{VIS} &
  \multicolumn{1}{l|}{\gradientNeuro{0.049}} &
  \gradientNeuro{0.054} &
  \multicolumn{1}{l|}{\gradientNeuro{-0.111}} &
  \gradientNeuro{-0.164} &
  \multicolumn{1}{l|}{\gradientNeuro{0.033}} &
  \gradientNeuro{-0.104} &
  \multicolumn{1}{c|}{\gradientNeuro{-0.055}} &
  \gradientNeuro{-0.194} &
  \multicolumn{1}{l|}{\gradientNeuro{0.04}} &
  \gradientNeuro{-0.026} &
  \multicolumn{1}{l|}{\gradientNeuro{-0.085}} &
  \gradientNeuro{-0.049} &
  \multicolumn{1}{l|}{\gradientNeuro{-0.259}} &
  \gradientNeuro{-0.243} \\ \hline
\textbf{MOT} &
  \multicolumn{1}{l|}{\gradientNeuro{-0.138}} &
  \gradientNeuro{0.095} &
  \multicolumn{1}{l|}{\gradientNeuro{-0.038}} &
  \gradientNeuro{0.137} &
  \multicolumn{1}{l|}{\gradientNeuro{0.063}} &
  \gradientNeuro{-0.09} &
  \multicolumn{1}{c|}{-} &
  - &
  \multicolumn{1}{l|}{-} &
  - &
  \multicolumn{1}{l|}{-} &
  - &
  \multicolumn{1}{l|}{-} &
  - \\ \hline
\textbf{PCIN} &
  \multicolumn{1}{l|}{\gradientNeuro{-0.443}} &
  \gradientNeuro{-0.508} &
  \multicolumn{1}{l|}{\gradientNeuro{-0.449}} &
  \gradientNeuro{-0.571} &
  \multicolumn{1}{l|}{\gradientNeuro{-0.465}} &
  \gradientNeuro{-0.544} &
  \multicolumn{1}{c|}{\gradientNeuro{-0.439}} &
  \gradientNeuro{-0.442} &
  \multicolumn{1}{l|}{\gradientNeuro{-0.499}} &
  \gradientNeuro{-0.549} &
  \multicolumn{1}{l|}{\gradientNeuro{-0.489}} &
  \gradientNeuro{-0.536} &
  \multicolumn{1}{l|}{\gradientNeuro{-0.473}} &
  \gradientNeuro{-0.541} \\ \hline
\textbf{MF} &
  \multicolumn{1}{l|}{\gradientNeuro{-0.508}} &
  \gradientNeuro{-0.297} &
  \multicolumn{1}{l|}{\gradientNeuro{-0.364}} &
  \gradientNeuro{-0.33} &
  \multicolumn{1}{l|}{\gradientNeuro{-0.143}} &
  \gradientNeuro{-0.142} &
  \multicolumn{1}{c|}{\gradientNeuro{-0.479}} &
  \gradientNeuro{-0.363} &
  \multicolumn{1}{l|}{\gradientNeuro{-0.463}} &
  \gradientNeuro{-0.528} &
  \multicolumn{1}{l|}{\gradientNeuro{-0.323}} &
  \gradientNeuro{-0.192} &
  \multicolumn{1}{l|}{\gradientNeuro{-0.093}} &
  \gradientNeuro{-0.058} \\ \hline
\textbf{ACIN} &
  \multicolumn{1}{l|}{\gradientNeuro{0.037}} &
  \gradientNeuro{-0.185} &
  \multicolumn{1}{l|}{\gradientNeuro{0.017}} &
  \gradientNeuro{-0.087} &
  \multicolumn{1}{l|}{\gradientNeuro{-0.1}} &
  \gradientNeuro{-0.08} &
  \multicolumn{1}{c|}{\gradientNeuro{0.078}} &
  \gradientNeuro{0.125} &
  \multicolumn{1}{l|}{\gradientNeuro{-0.149}} &
  \gradientNeuro{-0.057} &
  \multicolumn{1}{l|}{\gradientNeuro{0.313}} &
  \gradientNeuro{0.105} &
  \multicolumn{1}{l|}{\gradientNeuro{0.382}} &
  \gradientNeuro{0.302} \\ \hline
\textbf{ORF} &
  \multicolumn{1}{l|}{\gradientNeuro{0.103}} &
  \gradientNeuro{0.151} &
  \multicolumn{1}{l|}{\gradientNeuro{0.238}} &
  \gradientNeuro{0.183} &
  \multicolumn{1}{l|}{\gradientNeuro{0.219}} &
  \gradientNeuro{-0.115} &
  \multicolumn{1}{c|}{-} &
  - &
  \multicolumn{1}{l|}{-} &
  - &
  \multicolumn{1}{l|}{-} &
  - &
  \multicolumn{1}{l|}{-} &
  - \\ \hline
\textbf{STEMP} &
  \multicolumn{1}{l|}{\gradientNeuro{-0.028}} &
  \gradientNeuro{0} &
  \multicolumn{1}{l|}{\gradientNeuro{0.076}} &
  \gradientNeuro{-0.014} &
  \multicolumn{1}{l|}{\gradientNeuro{0.092}} &
  \gradientNeuro{0.021} &
  \multicolumn{1}{c|}{\gradientNeuro{0.01}} &
  \gradientNeuro{-0.045} &
  \multicolumn{1}{l|}{\gradientNeuro{0.196}} &
  \gradientNeuro{0.146} &
  \multicolumn{1}{l|}{\gradientNeuro{0.098}} &
  \gradientNeuro{0.096} &
  \multicolumn{1}{l|}{\gradientNeuro{0.212}} &
  \gradientNeuro{0.152} \\ \hline
\textbf{IF} &
  \multicolumn{1}{l|}{\gradientNeuro{0.209}} &
  \gradientNeuro{0.275} &
  \multicolumn{1}{l|}{\gradientNeuro{0.208}} &
  \gradientNeuro{0.296} &
  \multicolumn{1}{l|}{\gradientNeuro{0.055}} &
  \gradientNeuro{0.27} &
  \multicolumn{1}{c|}{\gradientNeuro{-0.026}} &
  \gradientNeuro{0.073} &
  \multicolumn{1}{l|}{\gradientNeuro{0.016}} &
  \gradientNeuro{0.024} &
  \multicolumn{1}{l|}{\gradientNeuro{0.323}} &
  \gradientNeuro{0.374} &
  \multicolumn{1}{l|}{\gradientNeuro{0.327}} &
  \gradientNeuro{0.422} \\ \hline
\textbf{ANTIN} &
  \multicolumn{1}{l|}{\gradientNeuro{-0.366}} &
  \gradientNeuro{-0.381} &
  \multicolumn{1}{l|}{\gradientNeuro{-0.166}} &
  \gradientNeuro{-0.39} &
  \multicolumn{1}{l|}{\gradientNeuro{-0.181}} &
  \gradientNeuro{-0.457} &
  \multicolumn{1}{c|}{\gradientNeuro{-0.247}} &
  \gradientNeuro{-0.434} &
  \multicolumn{1}{l|}{\gradientNeuro{-0.242}} &
  \gradientNeuro{-0.455} &
  \multicolumn{1}{l|}{\gradientNeuro{-0.28}} &
  \gradientNeuro{-0.398} &
  \multicolumn{1}{l|}{\gradientNeuro{-0.178}} &
  \gradientNeuro{-0.268} \\ \hline
\textbf{ATEMPP} &
  \multicolumn{1}{l|}{\gradientNeuro{-0.568}} &
  \gradientNeuro{-0.516} &
  \multicolumn{1}{l|}{\gradientNeuro{-0.649}*} &
  \gradientNeuro{-0.515} &
  \multicolumn{1}{l|}{\gradientNeuro{-0.435}} &
  \gradientNeuro{-0.321} &
  \multicolumn{1}{c|}{\gradientNeuro{-0.615}*} &
  \gradientNeuro{-0.546} &
  \multicolumn{1}{l|}{\gradientNeuro{-0.461}} &
 \gradientNeuro{-0.432} &
  \multicolumn{1}{l|}{\gradientNeuro{-0.512}} &
  \gradientNeuro{-0.55} &
  \multicolumn{1}{l|}{\gradientNeuro{-0.363}} &
  \gradientNeuro{-0.246} \\ \hline
\textbf{VLT} &
  \multicolumn{1}{l|}{\gradientNeuro{0.029}} &
  \gradientNeuro{-0.264} &
  \multicolumn{1}{l|}{\gradientNeuro{0.21}} &
  \gradientNeuro{-0.179} &
  \multicolumn{1}{l|}{\gradientNeuro{0.097}} &
  \gradientNeuro{-0.183} &
  \multicolumn{1}{c|}{\gradientNeuro{0.047}} &
  \gradientNeuro{-0.061} &
  \multicolumn{1}{l|}{\gradientNeuro{0.071}} &
  \gradientNeuro{-0.26} &
  \multicolumn{1}{l|}{\gradientNeuro{0.179}} &
  \gradientNeuro{-0.107} &
  \multicolumn{1}{l|}{\gradientNeuro{0.094}} &
  \gradientNeuro{-0.15} \\ \hline
\textbf{SP} &
  \multicolumn{1}{l|}{\gradientNeuro{-0.207}} &
  \gradientNeuro{-0.115} &
  \multicolumn{1}{l|}{\gradientNeuro{-0.13}} &
  \gradientNeuro{-0.123} &
  \multicolumn{1}{l|}{\gradientNeuro{-0.195}} &
  \gradientNeuro{-0.074} &
  \multicolumn{1}{c|}{\gradientNeuro{-0.275}} &
  \gradientNeuro{-0.482} &
  \multicolumn{1}{l|}{\gradientNeuro{-0.153}} &
  \gradientNeuro{-0.232} &
  \multicolumn{1}{l|}{\gradientNeuro{-0.208}} &
 \gradientNeuro{-0.128} &
  \multicolumn{1}{l|}{\gradientNeuro{-0.148}} &
  \gradientNeuro{-0.146} \\ \hline
\textbf{ANG} &
  \multicolumn{1}{l|}{\gradientNeuro{-0.33}} &
  \gradientNeuro{-0.506} &
  \multicolumn{1}{l|}{\gradientNeuro{-0.42}} &
  \gradientNeuro{-0.519} &
  \multicolumn{1}{l|}{\gradientNeuro{-0.205}} &
  \gradientNeuro{-0.348} &
  \multicolumn{1}{c|}{\gradientNeuro{-0.485}} &
  \gradientNeuro{-0.573}* &
  \multicolumn{1}{l|}{\gradientNeuro{-0.46}} &
  \gradientNeuro{-0.614}* &
  \multicolumn{1}{l|}{\gradientNeuro{-0.31}} &
  \gradientNeuro{-0.406} &
  \multicolumn{1}{l|}{\gradientNeuro{-0.378}} &
  \gradientNeuro{-0.368}\\ \hline
\textbf{ERC} &
  \multicolumn{1}{l|}{\gradientNeuro{-0.256}} &
  \gradientNeuro{-0.416} &
  \multicolumn{1}{l|}{\gradientNeuro{-0.247}} &
  \gradientNeuro{-0.403} &
  \multicolumn{1}{l|}{\gradientNeuro{0.051}} &
  \gradientNeuro{-0.289} &
  \multicolumn{1}{c|}{\gradientNeuro{-0.179}} &
  \gradientNeuro{-0.231} &
  \multicolumn{1}{l|}{\gradientNeuro{-0.37}} &
  \gradientNeuro{-0.395} &
  \multicolumn{1}{l|}{\gradientNeuro{-0.281}} &
  \gradientNeuro{-0.453} &
  \multicolumn{1}{l|}{\gradientNeuro{-0.344}} &
  \gradientNeuro{-0.407} \\ \hline
\textbf{BA35} &
  \multicolumn{1}{l|}{\gradientNeuro{-0.092}} &
  \gradientNeuro{-0.29} &
  \multicolumn{1}{l|}{\gradientNeuro{-0.059}} &
  \gradientNeuro{-0.424} &
  \multicolumn{1}{l|}{\gradientNeuro{-0.312}} &
  \gradientNeuro{-0.303} &
  \multicolumn{1}{c|}{\gradientNeuro{-0.254}} &
  \gradientNeuro{-0.388} &
  \multicolumn{1}{l|}{\gradientNeuro{-0.324}} &
  \gradientNeuro{-0.365} &
  \multicolumn{1}{l|}{\gradientNeuro{-0.3}} &
  \gradientNeuro{-0.485} &
  \multicolumn{1}{l|}{\gradientNeuro{-0.359}} &
  \gradientNeuro{-0.373} \\ \hline
\textbf{PHC} &
  \multicolumn{1}{l|}{\gradientNeuro{-0.11}} &
  \gradientNeuro{-0.374} &
  \multicolumn{1}{l|}{\gradientNeuro{-0.121}} &
  \gradientNeuro{-0.543} &
  \multicolumn{1}{l|}{\gradientNeuro{0.009}} &
  \gradientNeuro{-0.211} &
  \multicolumn{1}{c|}{\gradientNeuro{-0.076}} &
  \gradientNeuro{-0.188} &
  \multicolumn{1}{l|}{\gradientNeuro{-0.1}} &
  \gradientNeuro{-0.228} &
  \multicolumn{1}{l|}{\gradientNeuro{-0.125}} &
  \gradientNeuro{-0.336} &
  \multicolumn{1}{l|}{\gradientNeuro{-0.085}} &
  \gradientNeuro{-0.304} \\ \hline
\end{tabular}
}
\end{adjustbox}

%%%%%%%%%%
\begin{figure}[H]
\centering
\includegraphics[width=\textwidth,height=\textheight,keepaspectratio]{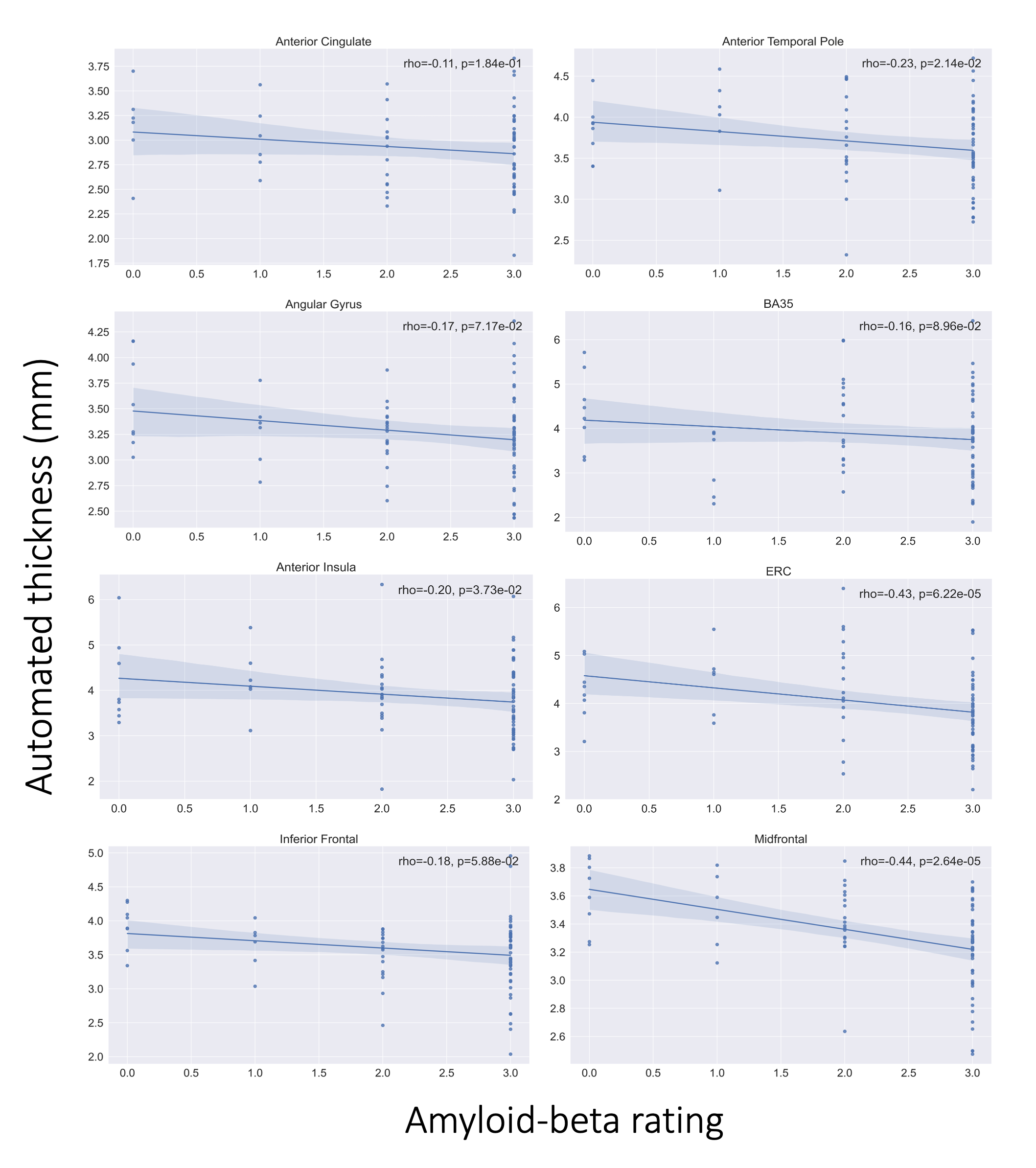}
        \caption{\textcolor{blue}{Spearman's correlation between cortical thickness measures  derived from topologically corrected nnU-Net-CRUISE gray matter segmentation and global amyloid-$\beta$ rating with p-value.}}
        \addtocounter{figure}{-1} 
\end{figure}

\begin{figure}[H]
\centering
\includegraphics[width=\textwidth,height=\textheight,keepaspectratio]{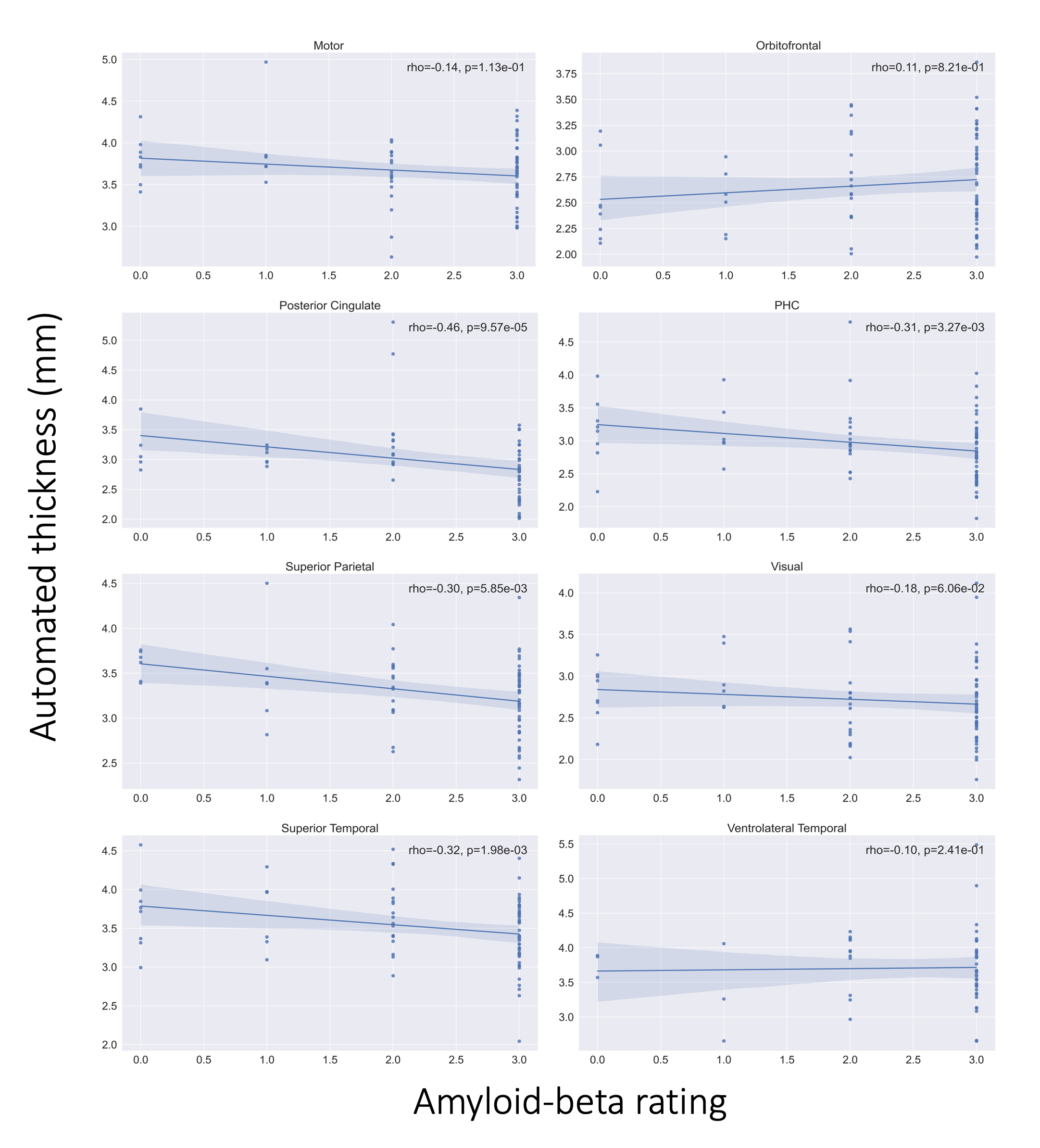}
        \caption{\textcolor{blue}{\textbf{\textit{Continued.}} Spearman's correlation between cortical thickness measures  derived from topologically corrected nnU-Net-CRUISE gray matter segmentation and global amyloid-$\beta$ rating with p-value.}}
\end{figure}

%%%%%%%%%%
\begin{figure}[H]
\centering
\includegraphics[width=\textwidth,height=\textheight,keepaspectratio]{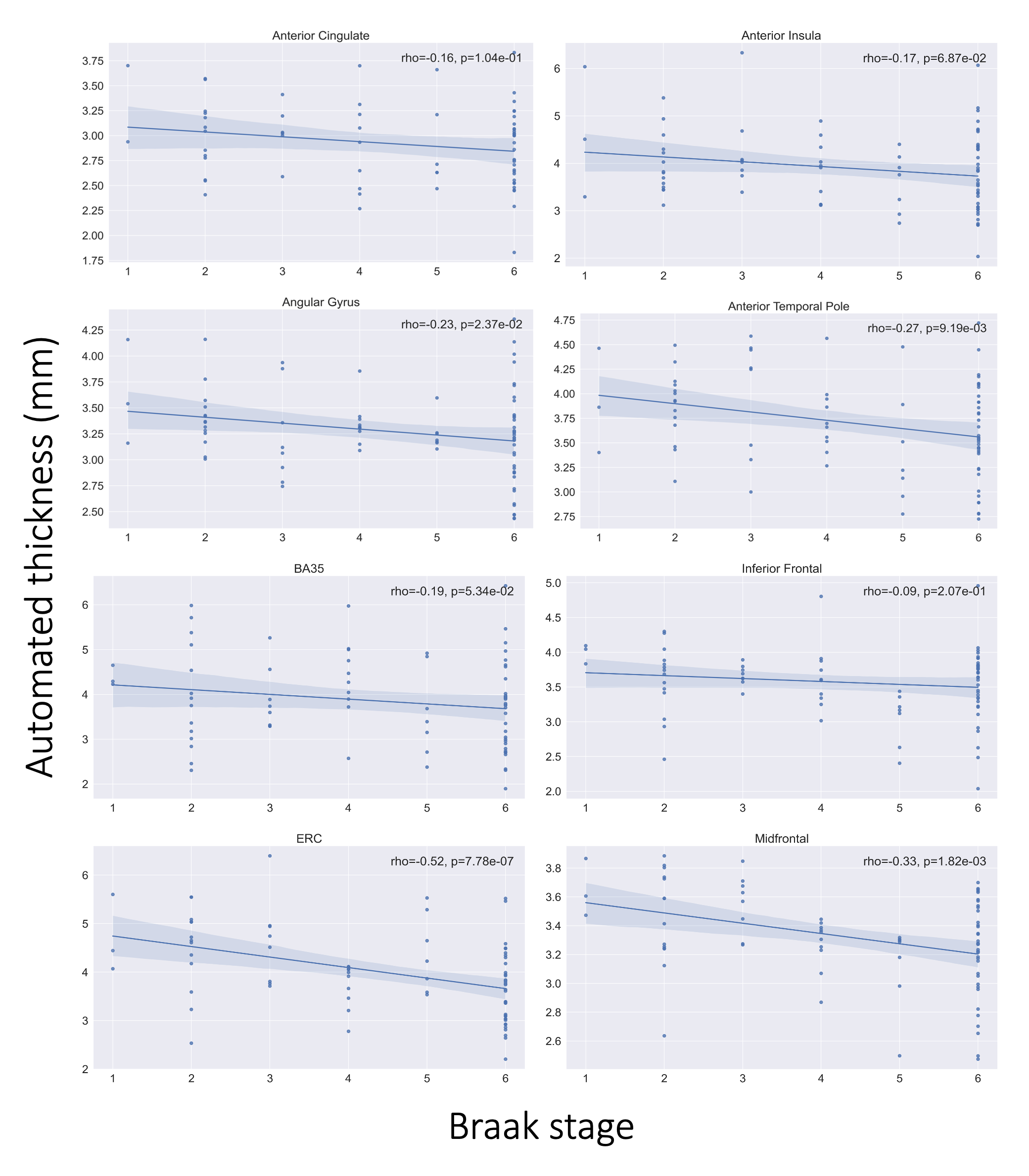}
        \caption{\textcolor{blue}{Spearman's correlation between cortical thickness measures  derived from topologically corrected nnU-Net-CRUISE  gray matter segmentation and Braak stage with p-value.}}
        \addtocounter{figure}{-1} 
\end{figure}

\begin{figure}[H]
\centering
\includegraphics[width=\textwidth,height=\textheight,keepaspectratio]{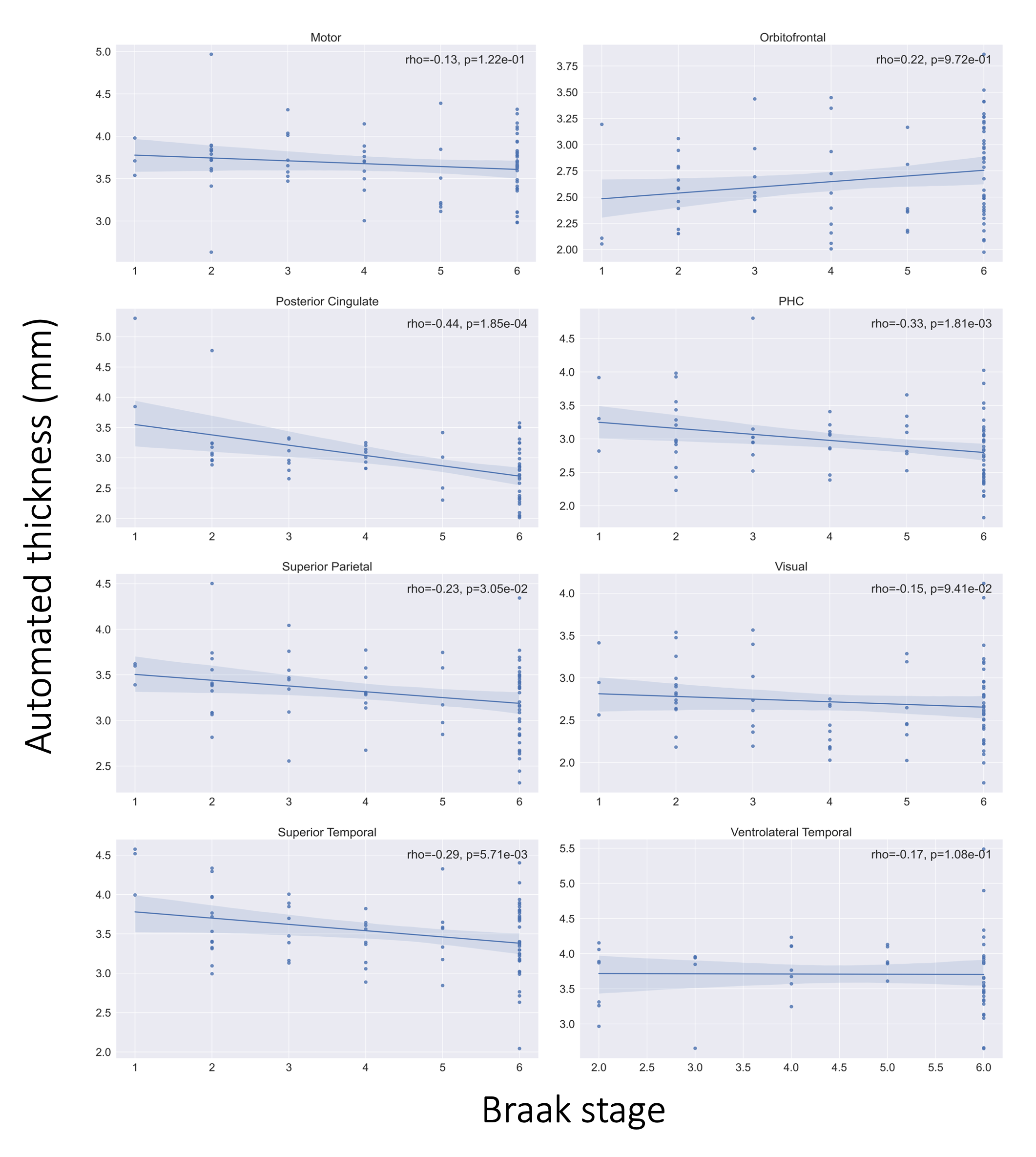}
        \caption{\textcolor{blue}{\textbf{\textit{Continued.}} Spearman's correlation between cortical thickness measures  derived from topologically corrected nnU-Net-CRUISE  gray matter segmentation and Braak stage with p-value.}}
\end{figure}

%%%%%%%%%%
\begin{figure}[H]
\centering
\includegraphics[width=\textwidth,height=\textheight,keepaspectratio]{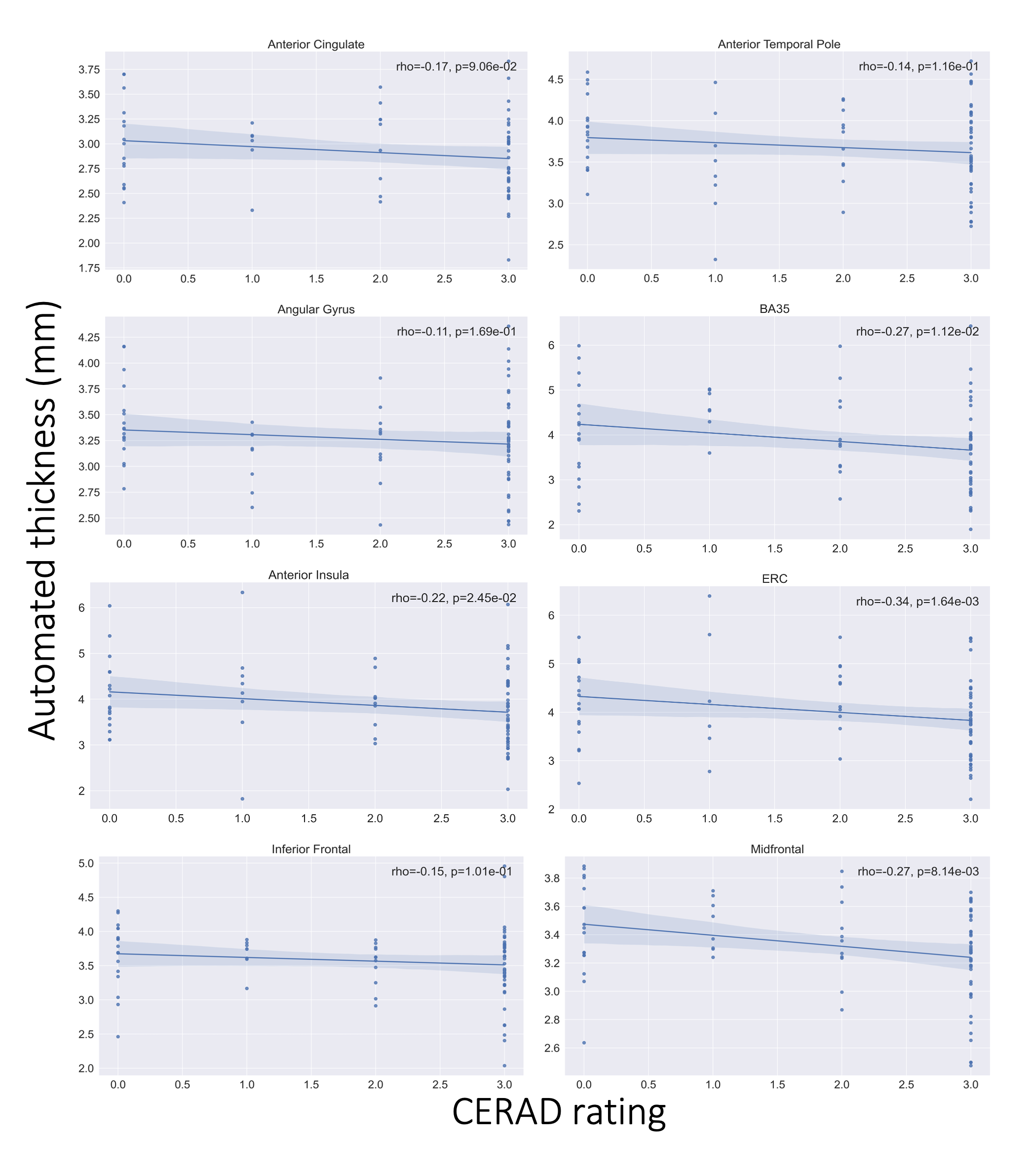}
        \caption{\textcolor{blue}{Spearman's correlation between cortical thickness measures  derived from topologically corrected nnU-Net-CRUISE gray matter segmentation and global CERAD rating with p-value.}}
        \addtocounter{figure}{-1} 
\end{figure}

\begin{figure}[H]
\centering
\includegraphics[width=\textwidth,height=\textheight,keepaspectratio]{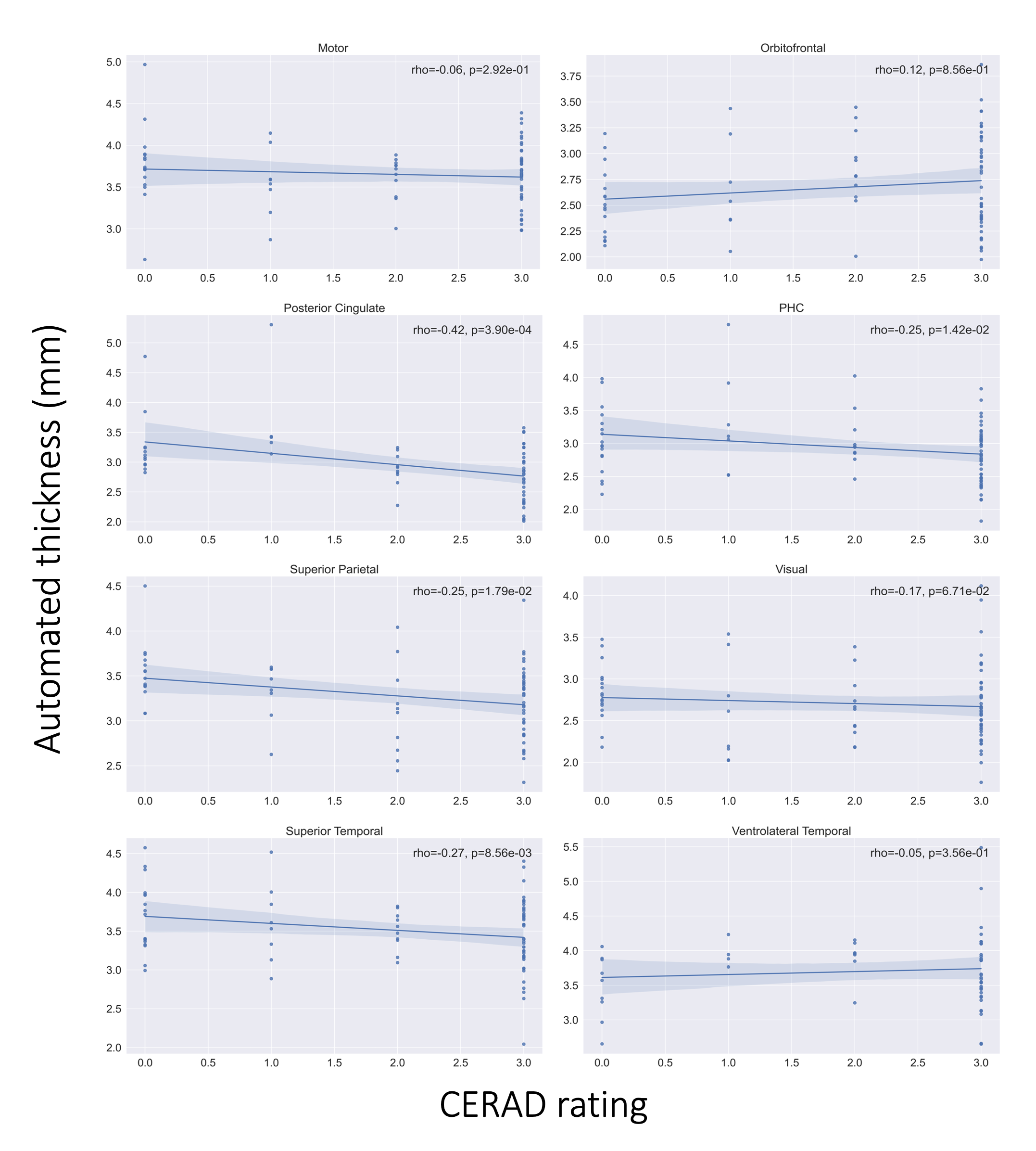}
        \caption{\textcolor{blue}{\textbf{\textit{Continued.}} Spearman's correlation between cortical thickness measures  derived from topologically corrected nnU-Net-CRUISE gray matter segmentation and global CERAD rating with p-value.}}
\end{figure}

%%%%%%%%%%
\begin{figure}[H]
\centering
\includegraphics[width=\textwidth,height=\textheight,keepaspectratio]{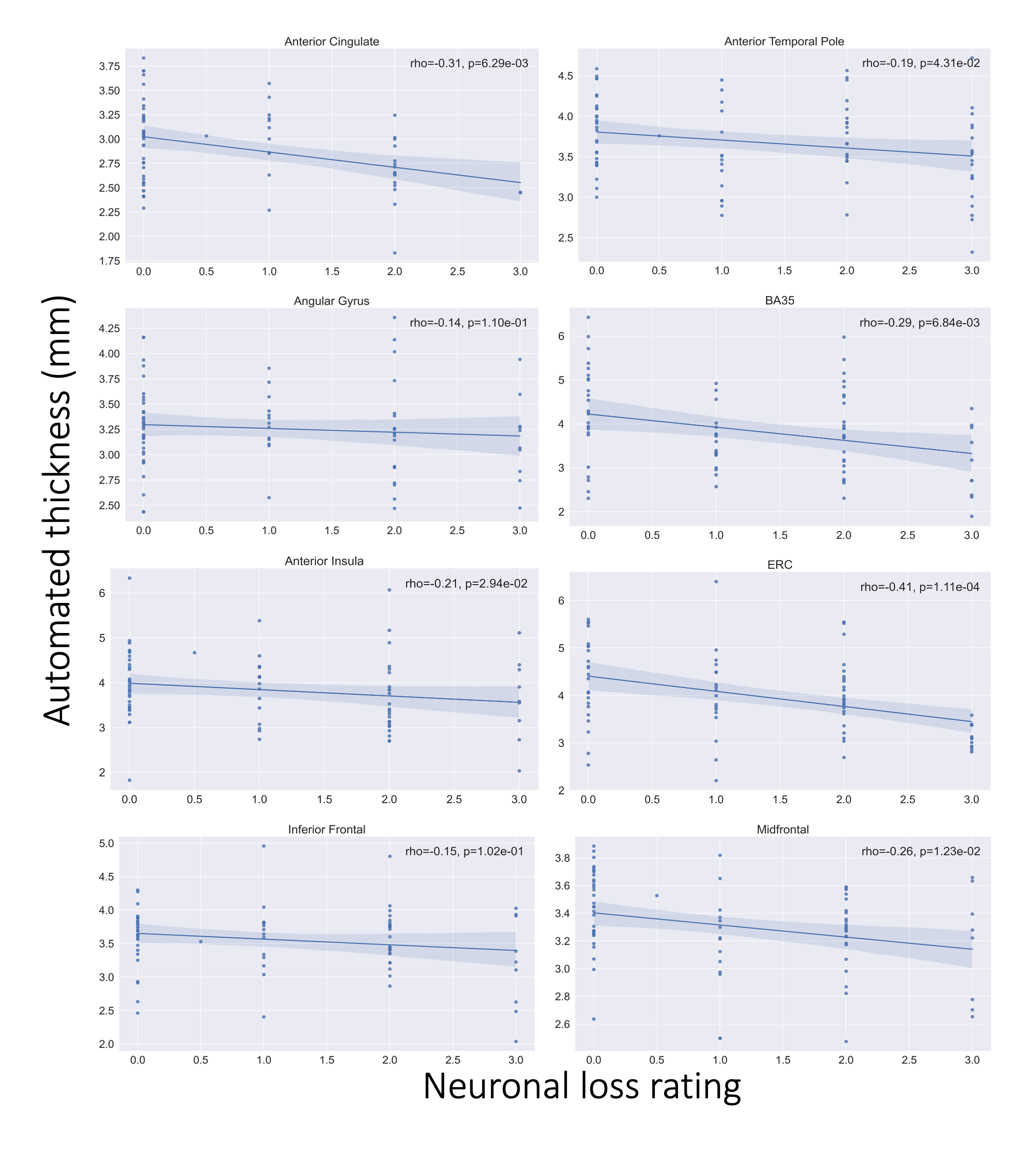}
        \caption{\textcolor{blue}{Spearman's correlation between cortical thickness measures  derived from topologically corrected nnU-Net-CRUISE gray matter segmentation and regional neuronal loss rating with p-value.}}
        \addtocounter{figure}{-1} 
\end{figure}

\begin{figure}[H]
\centering
\includegraphics[width=\textwidth,height=\textheight,keepaspectratio]{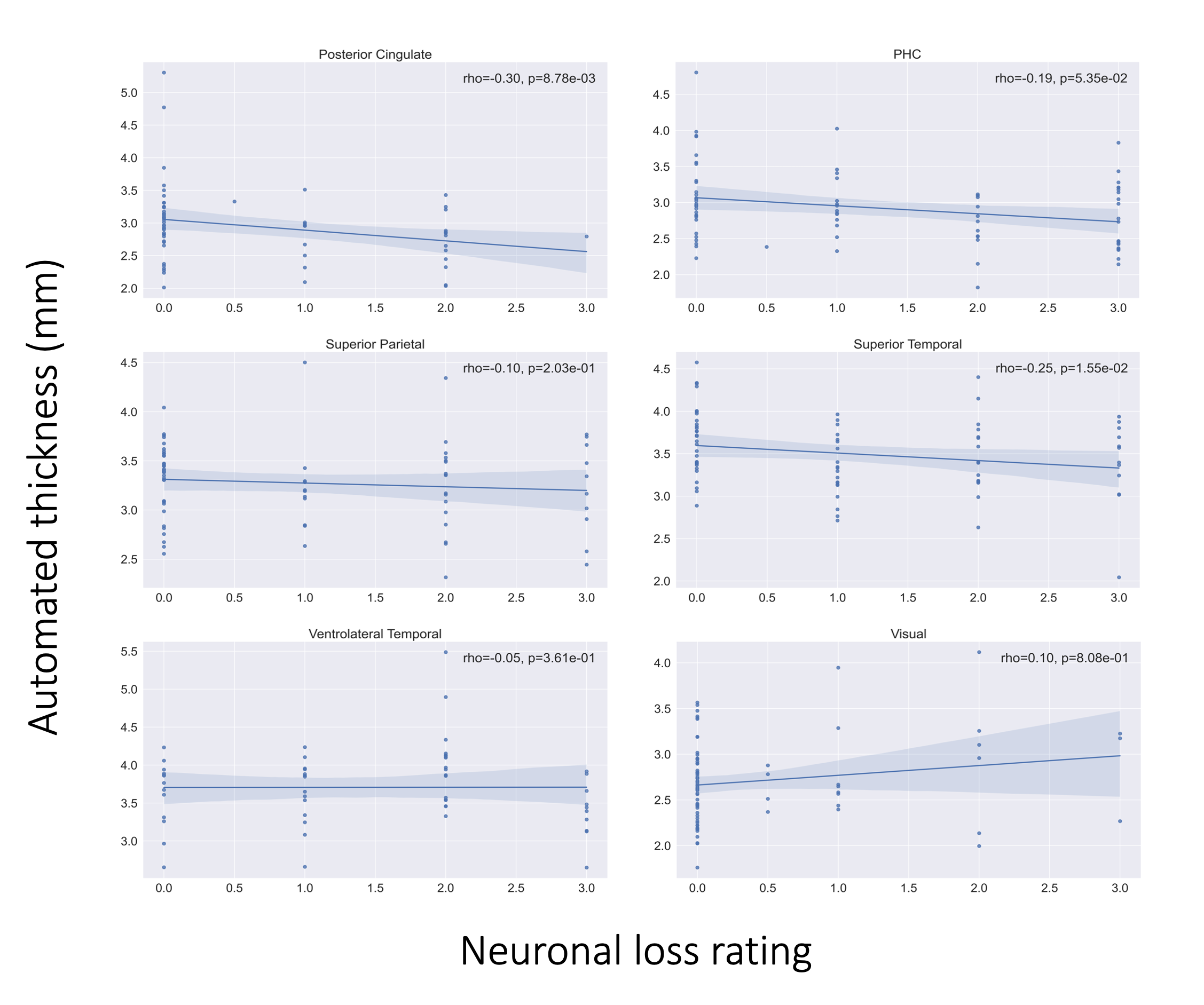}
        \caption{\textcolor{blue}{\textbf{\textit{Continued.}} Spearman's correlation between cortical thickness measures  derived from topologically corrected nnU-Net-CRUISE gray matter segmentation and regional neuronal loss rating with p-value.}}
\end{figure}

%%%%%%%%%%
\begin{figure}[H]
\centering
\includegraphics[width=\textwidth,height=\textheight,keepaspectratio]{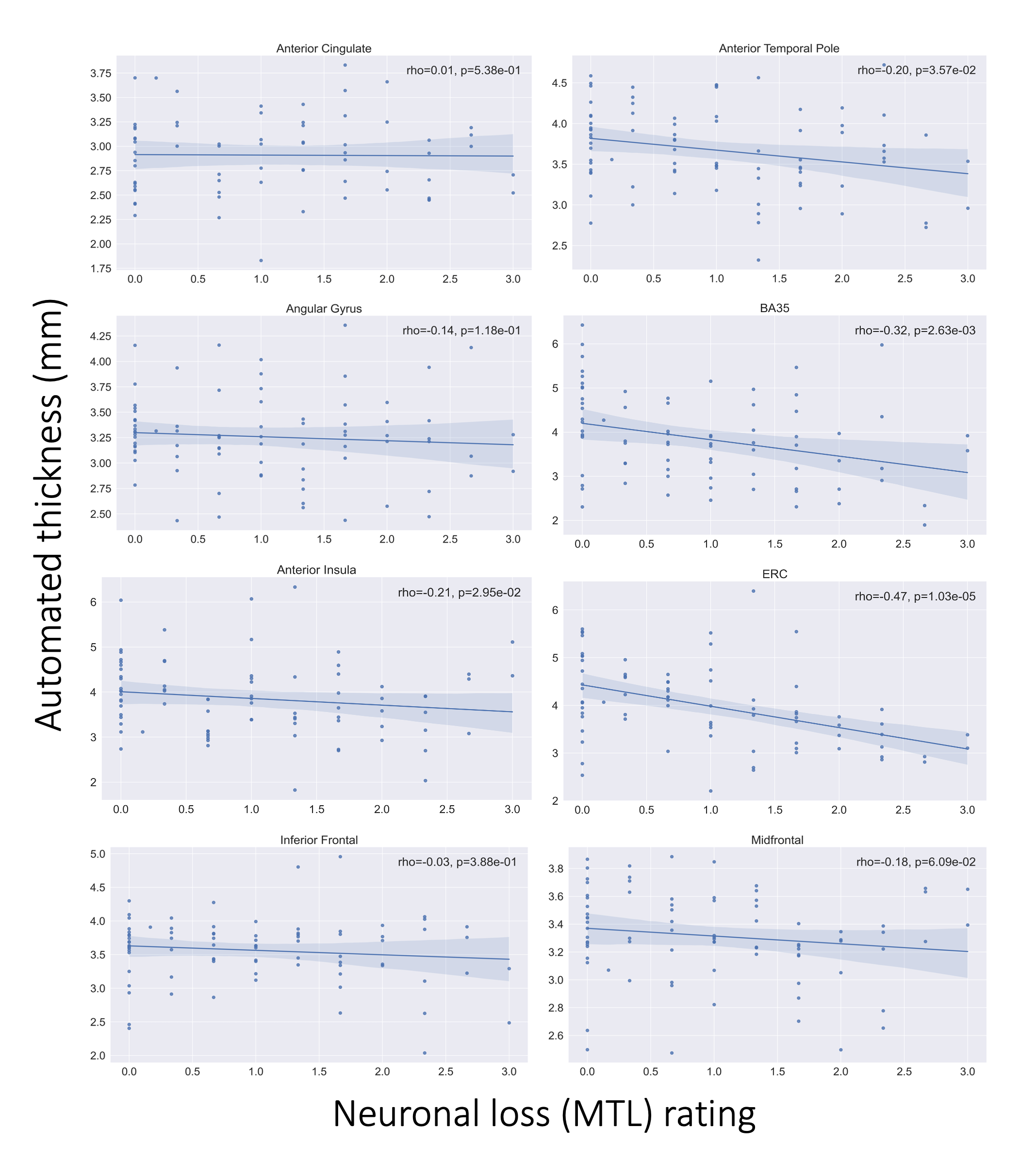}
        \caption{\textcolor{blue}{Spearman's correlation between cortical thickness measures  derived from topologically corrected nnU-Net-CRUISE gray matter segmentation and medial temporal lobe (MTL) neuronal loss rating with p-value.}}
        \addtocounter{figure}{-1} 
\end{figure}

\begin{figure}[H]
\centering
\includegraphics[width=\textwidth,height=\textheight,keepaspectratio]{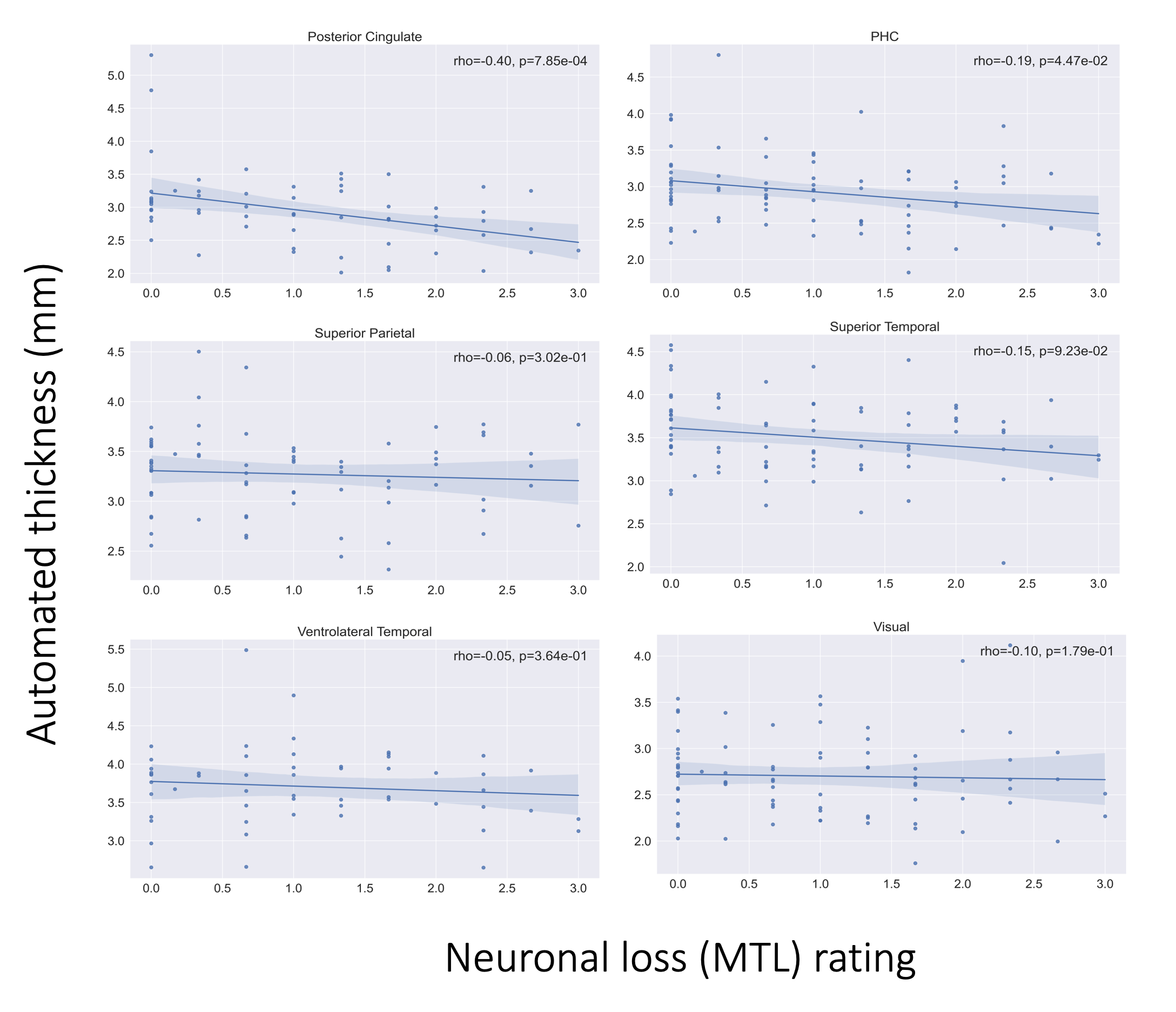}
        \caption{\textcolor{blue}{\textbf{\textit{Continued.}} Spearman's correlation between cortical thickness measures  derived from topologically corrected nnU-Net-CRUISE gray matter segmentation and the medial temporal lobe (MTL) neuronal loss rating with p-value.}}
\end{figure}

%%%%%%%%%%
\begin{figure}[H]
\centering
\includegraphics[width=\textwidth,height=\textheight,keepaspectratio]{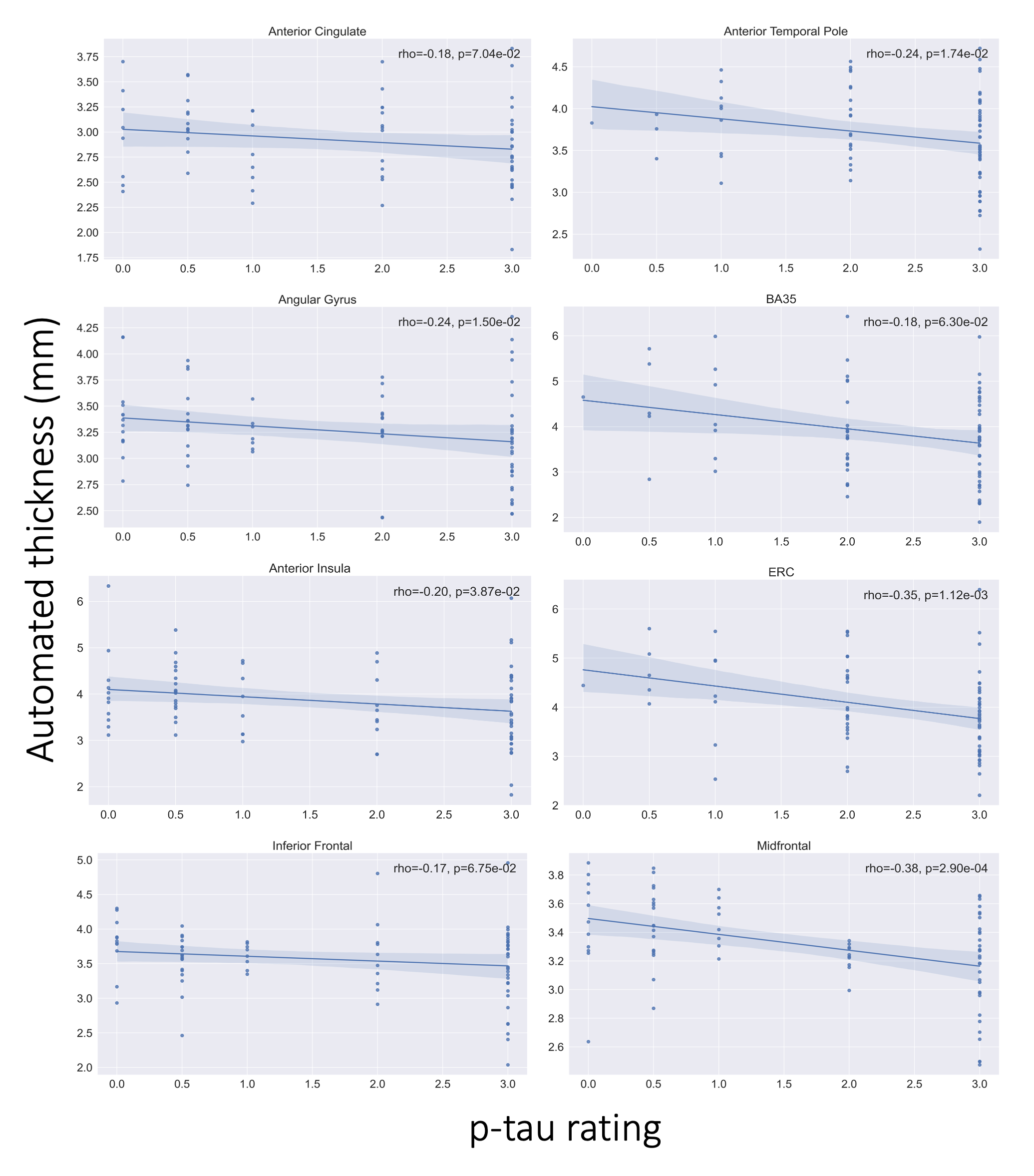}
        \caption{\textcolor{blue}{Spearman's correlation between cortical thickness measures  derived from topologically corrected nnU-Net-CRUISE gray matter segmentation and regional p-tau rating with p-value.}}
        \addtocounter{figure}{-1} 
\end{figure}

\begin{figure}[H]
\centering
\includegraphics[width=\textwidth,height=\textheight,keepaspectratio]{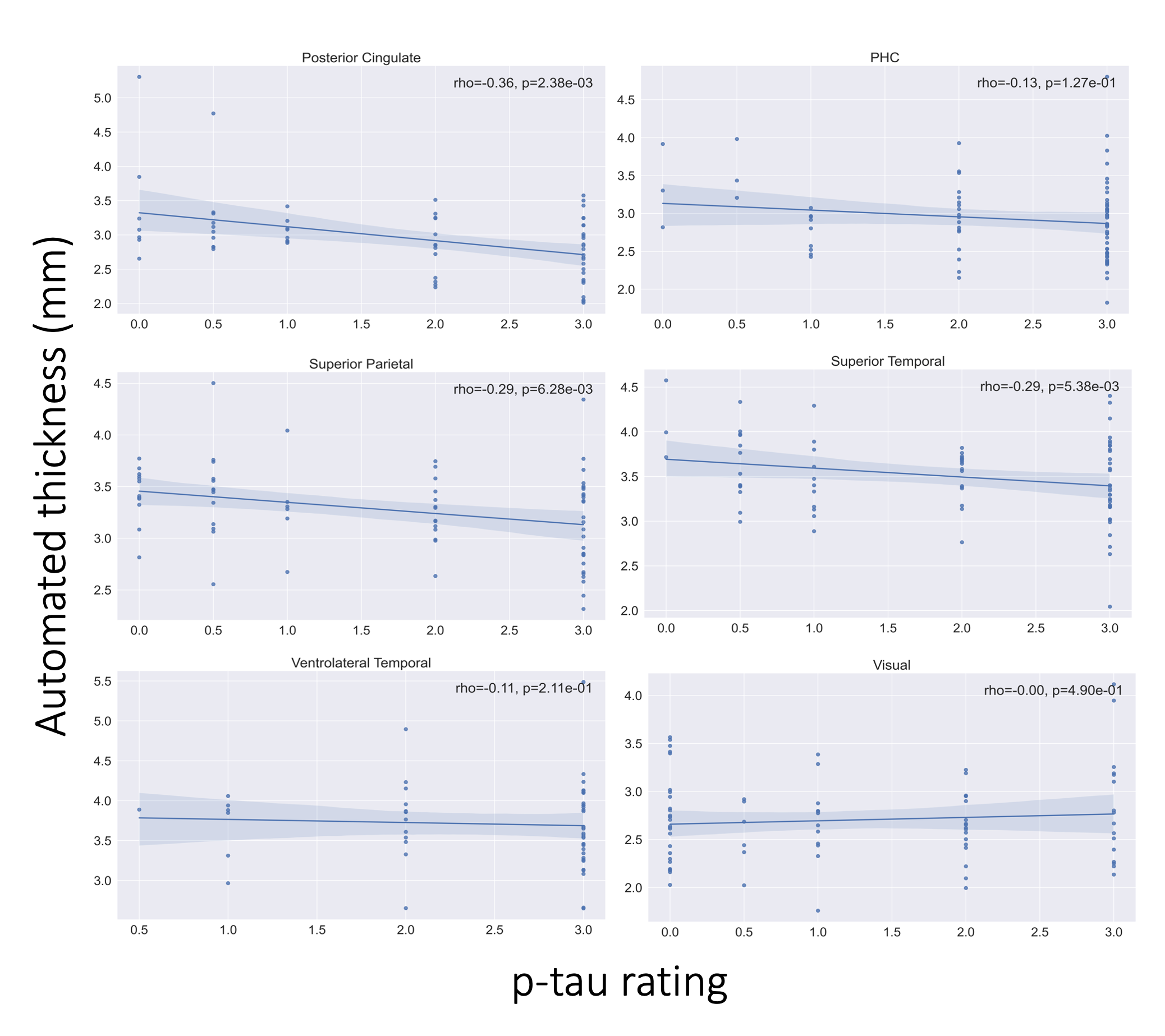}
        \caption{\textcolor{blue}{\textbf{\textit{Continued.}} Spearman's correlation between cortical thickness measures  derived from topologically corrected nnU-Net-CRUISE gray matter segmentation and regional p-tau rating with p-value.}}
\end{figure}

%%%%%%%%%%
\begin{figure}[H]
\centering
\includegraphics[width=\textwidth,height=\textheight,keepaspectratio]{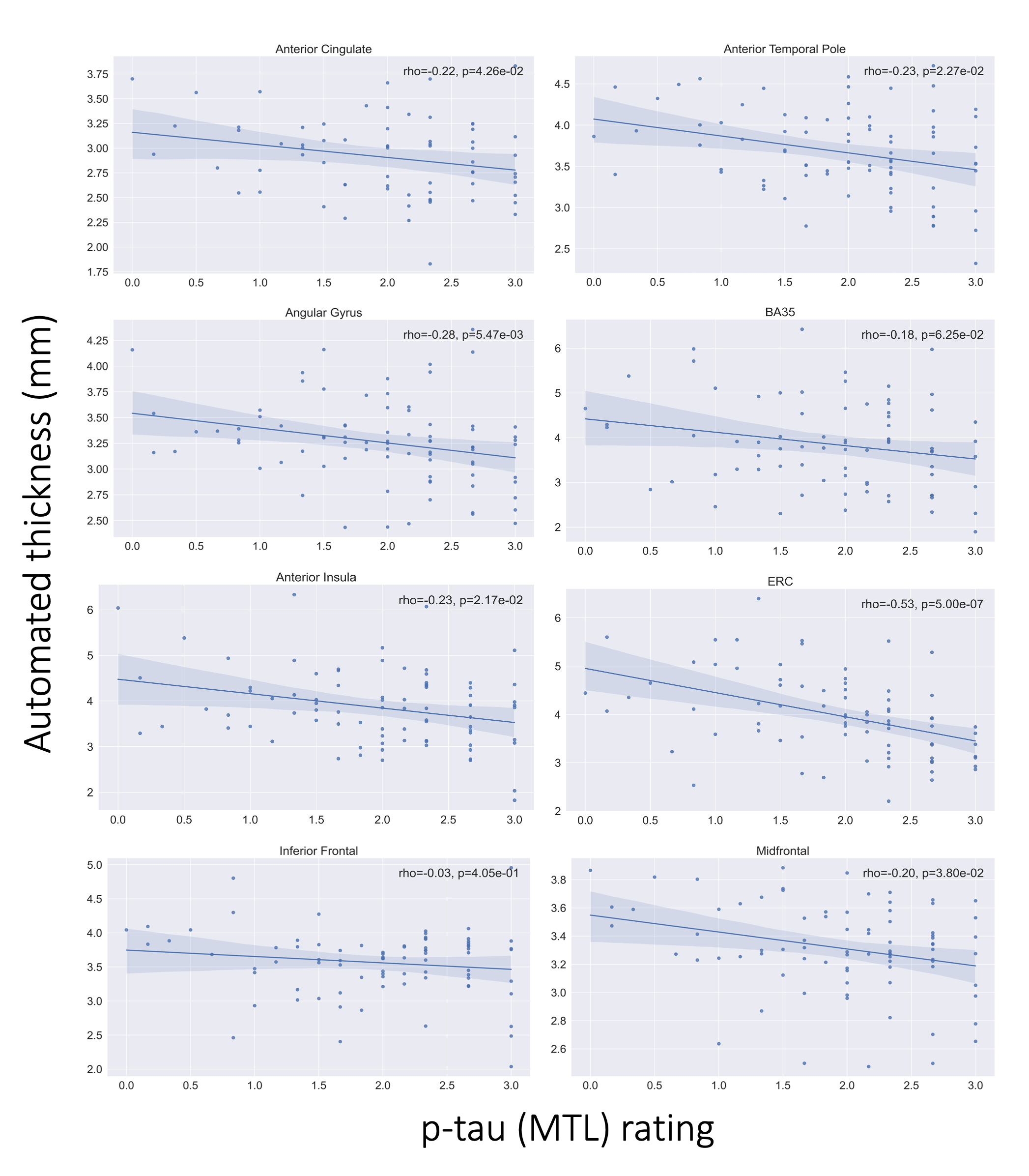}
        \caption{\textcolor{blue}{Spearman's correlation between cortical thickness measures  derived from topologically corrected nnU-Net-CRUISE gray matter segmentation and medial temporal lobe (MTL) p-tau rating with p-value.}}
        \addtocounter{figure}{-1} 
\end{figure}

\begin{figure}[H]
\centering
\includegraphics[width=\textwidth,height=\textheight,keepaspectratio]{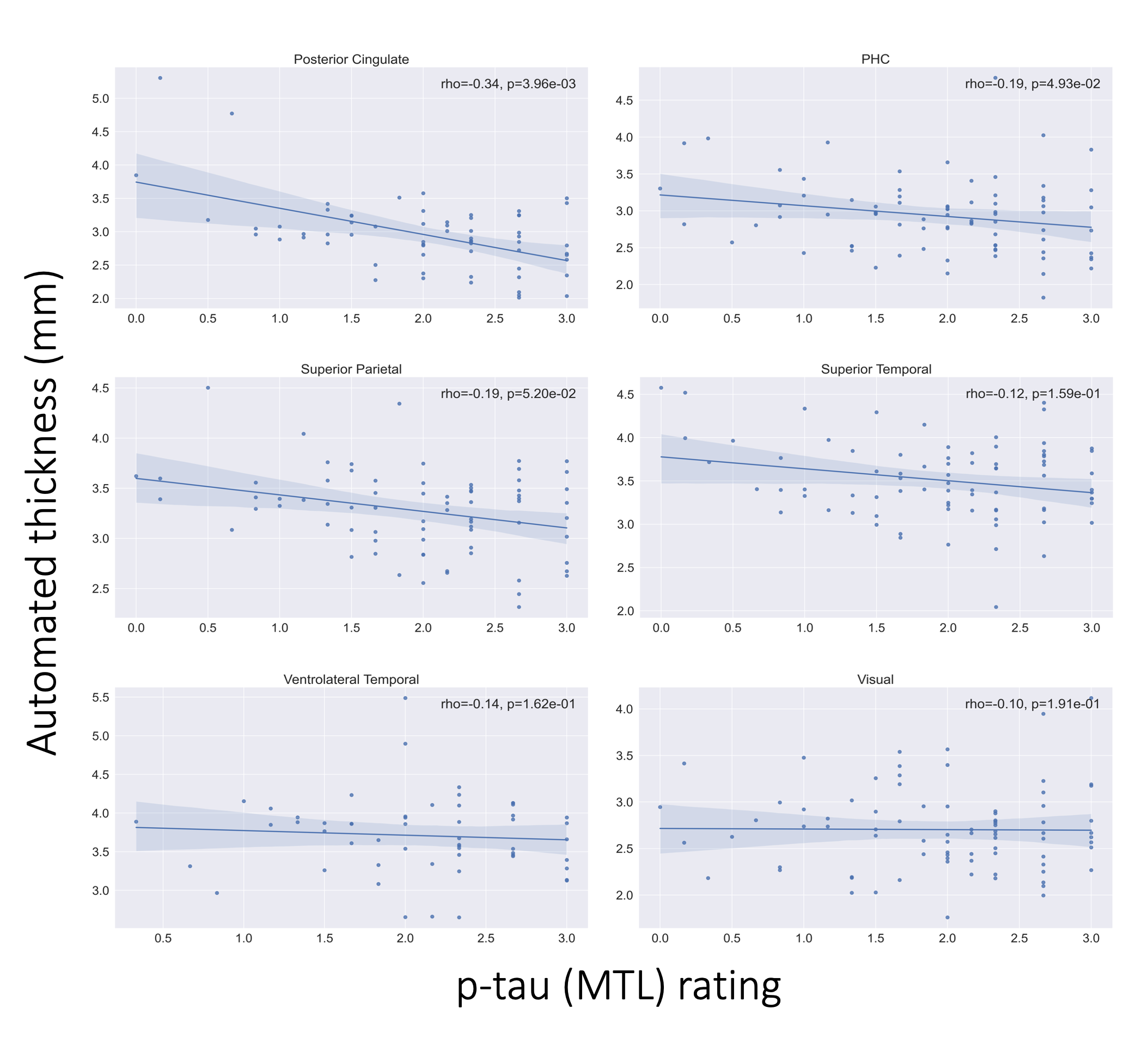}
        \caption{\textcolor{blue}{\textbf{\textit{Continued.}} Spearman's correlation between cortical thickness measures  derived from topologically corrected nnU-Net-CRUISE gray matter segmentation and medial temporal lobe (MTL) p-tau rating with p-value.}}
\end{figure}

%%%%%%%%%%
\begin{figure}[H]
\centering
\includegraphics[width=\textwidth,height=\textheight,keepaspectratio]{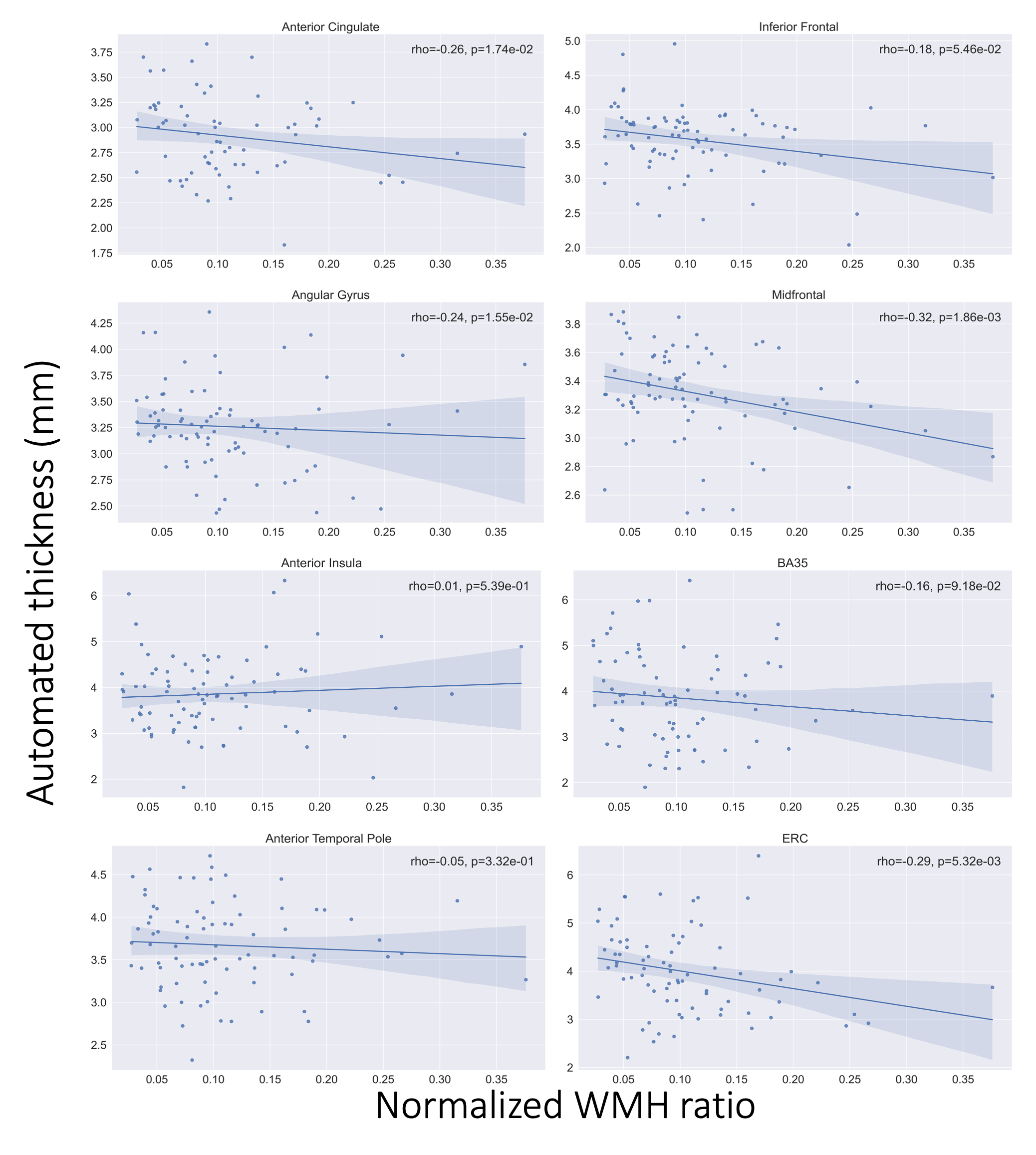}
        \caption{\textcolor{blue}{Spearman's correlation between cortical thickness measures derived from topologically corrected nnU-Net-CRUISE gray matter segmentation and the normalized white matter hyperintensities volume.}}
        \addtocounter{figure}{-1} 
\end{figure}

\begin{figure}[H]
\centering
\includegraphics[width=\textwidth,height=\textheight,keepaspectratio]{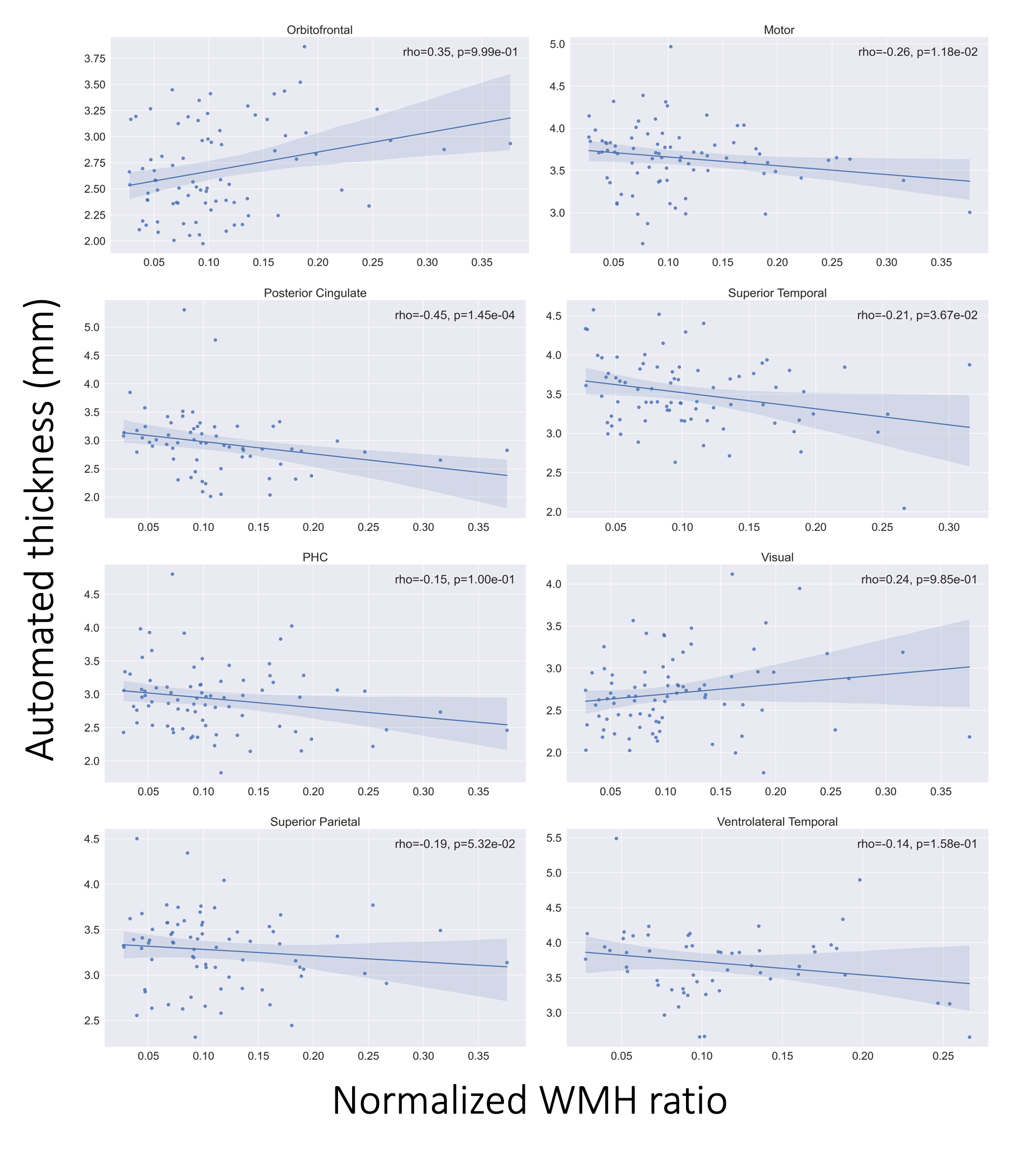}
        \caption{\textcolor{blue}{\textbf{\textit{Continued.}} Spearman's correlation between cortical thickness measures derived from topologically corrected nnU-Net-CRUISE gray matter segmentation and the normalized white matter hyperintensities volume.}}
\end{figure}

%%%%%%%%%%
\begin{figure}[H]
\centering
\includegraphics[width=\textwidth,height=\textheight,keepaspectratio]{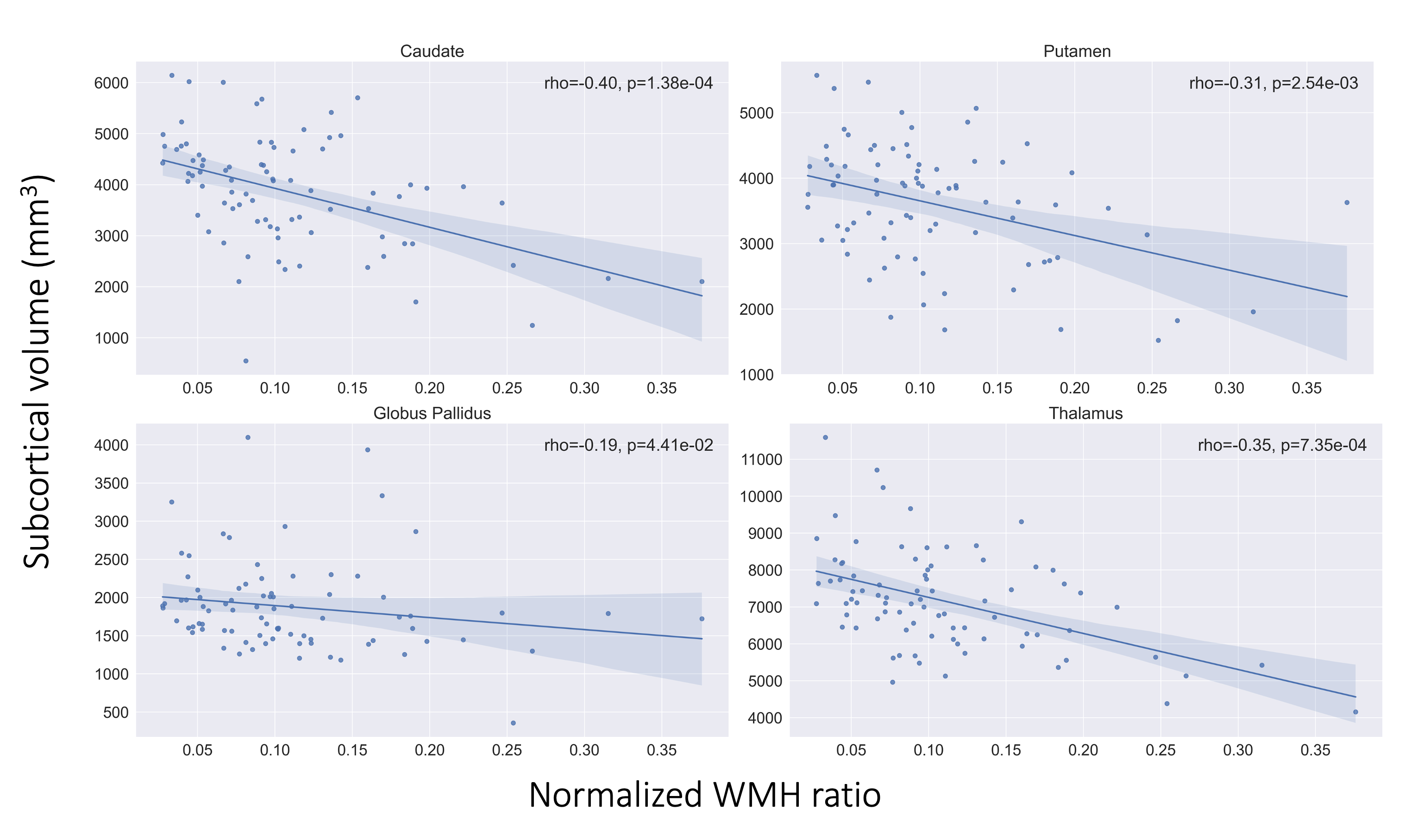}
        \caption{\textcolor{blue}{Spearman's correlation between subcortical structure volume and the normalized white matter hyperintensities volume.}}
\end{figure}

\bibliography{mybibfile}
\end{document}